\documentclass[a4paper,fleqn]{cas-sc}
\usepackage{blindtext}
\usepackage{tabularx}
\usepackage{hyperref}
\usepackage{multirow}
\usepackage[sort,numbers]{natbib}
\usepackage{hyperref}
\usepackage{threeparttablex}
\usepackage{adjustbox}
\usepackage{caption}
\usepackage{rotating}
\usepackage{svg}
\usepackage{array}
\usepackage{pifont}
\usepackage{float}
\usepackage{caption}
\usepackage{comment}
\usepackage{amsmath}

\usepackage{subcaption}
\usepackage{soul,xcolor}
\usepackage[section]{placeins}
\setstcolor{red}
\usepackage{lscape}
\usepackage{longtable}
\usepackage{etoolbox}
\AtBeginEnvironment{longtable}{\sffamily\small}
\usepackage{amssymb}% http://ctan.org/pkg/amssymb
\usepackage{pifont}% http://ctan.org/pkg/pifont
\newcolumntype{P}[1]{>{\centering\arraybackslash}p{#1}}
\def\tsc#1{\csdef{#1}{\textsc{\lowercase{#1}}\xspace}}
\tsc{WGM}
\tsc{QE}
\begin{document}
\let\WriteBookmarks\relax
\def\floatpagepagefraction{1}
\def\textpagefraction{.001}

\shorttitle{}    

\shortauthors{Yuhao Nie \textit{et al.}}  

\title [mode = title]{Open-Source Ground-based Sky Image Datasets for Very Short-term Solar Forecasting, Cloud Analysis and Modeling: A Comprehensive Survey}

% OR
%\title [mode = title]{Open-Source Sky Image Datasets for Solar Nowcasting and Cloud Analysis with Deep Learning: A Comprehensive Survey}

%\tnotemark[<tnote number>] 

%\tnotetext[<tnote number>]{<tnote text>} 


\author[1]{Yuhao Nie}[orcid=0000-0003-0345-457X]
\cormark[1]

%\fnmark[]

\ead{ynie@stanford.edu}


%\credit{Conceptualization of this study, Methodology, Software}

\affiliation[1]{organization={Department of Energy Resources Engineering},
            addressline={Stanford University}, 
            city={Stanford},
            %citysep={}, % Uncomment if no comma needed between city and postcode
            state={California},
            postcode={94305},
            country={United States}}

\author[2]{Xiatong Li}[]

\affiliation[2]{organization={Department of Civil and Environmental Engineering},
            addressline={Stanford University}, 
            city={Stanford},
%          citysep={}, % Uncomment if no comma needed between city and postcode
            postcode={94305}, 
            state={California},
            country={United States}}
            
\author[3, 4]{Quentin Paletta}[orcid=0000-0001-5254-9933]

\affiliation[3]{organization={Department of Engineering},
            addressline={University of Cambridge}, 
            city={Cambridge},
%          citysep={}, % Uncomment if no comma needed between city and postcode
            postcode={CB2 1PZ}, 
            country={United Kingdom}}

\affiliation[4]{organization={Engie Lab CRIGEN},
%            addressline={Engie}, 
            city={Stains},
%          citysep={}, % Uncomment if no comma needed between city and postcode
            postcode={93240}, 
            country={France}}

\author[5, 6]{Max Aragon}[]

\affiliation[5]{organization={Department of Geoinformatics},
            addressline={Universität Salzburg}, 
            city={Salzburg},
%          citysep={}, % Uncomment if no comma needed between city and postcode
            postcode={5020}, 
            %state={},
            country={Austria}}

\affiliation[6]{organization={Department of Computer Science},
            addressline={Université Bretagne Sud}, 
            city={Vannes},
%          citysep={}, % Uncomment if no comma needed between city and postcode
            postcode={56000}, 
            %state={},
            country={France}}

\author[1]{Andea Scott}[]
            
\author[1]{Adam Brandt}[orcid=0000-0002-2528-1473]
% Footnote of the second author

%\cormark[1]

% Email id of the second author
%\ead{abrandt@stanford.edu}



\cortext[1]{Corresponding author}



\begin{abstract}
Sky-image-based solar forecasting using deep learning has been recognized as a promising approach in reducing the uncertainty in solar power generation. However, one of the biggest challenges is the lack of massive and diversified sky image samples. In this study, we present a comprehensive survey of open-source ground-based sky image datasets for very short-term solar forecasting (i.e., forecasting horizon less than 30 minutes), as well as related research areas which can potentially help improve solar forecasting methods, including cloud segmentation, cloud classification and cloud motion prediction. We first identify 72 open-source sky image datasets that satisfy the needs of machine/deep learning. Then a database of information about various aspects of the identified datasets is constructed. To evaluate each surveyed datasets, we further develop a multi-criteria ranking system based on 8 dimensions of the datasets which could have important impacts on usage of the data. Finally, we provide insights on the usage of these datasets for different applications. We hope this paper can provide an overview for researchers who are looking for datasets for very short-term solar forecasting and related areas.
\end{abstract}


\begin{highlights}
\item 72 open-source ground-based sky image datasets covering diverse climate zones are identified globally.
\item A database consisting of extensive information on various aspects of the datasets is constructed.
\item A multi-criteria ranking system is developed to evaluate a dataset for different applications.
\item Insights and means of access are provided to the potential users of the datasets.
\end{highlights}

% Keywords

\maketitle

% Main text
\section{Introduction}\label{sec:intro}
Solar energy has been recognized as one of the crucial components for transition to a next-generation sustainable energy system \cite{gielen2019role}. Though with massive potential \cite{kabir2018solar}, large-scale deployment of solar power, primarily photovoltaics (PVs), is hindered by the intermittency of solar energy. The variability in solar power generation is mostly associated with local weather conditions, especially cloud motion \cite{barbieri2017very, Sun2019}. For example, up to 80\% drop in power output can occur to a rooftop solar PV system in less than a minute due to a cloud passage event \cite{Sun2019dissertation}. Under high penetration of renewable resources in future energy systems, rapid loss of megawatts or gigawatts of power from centralized PV plants can cause challenges for electricity grids. Thus, the development of accurate and reliable forecasting methods is urgently needed for handling uncertainty in solar power generation.

Depending on the time horizon it targets, solar forecasting\footnote{For simplicity, we refer to the forecast of solar irradiance and PV power output combined as solar forecasting in this study.} can be classified into the following categories, although as yet there is no common agreement on the classification criterion \cite{ahmed2020review}: (1) Very short-term forecasting, covering forecast horizon from a few seconds to 30 minutes \cite{ren2015ensemble}, is beneficial to activities such as electricity marketing or pricing, real-time dispatch of other generators and energy storage control \cite{das2018forecasting}; (2) Short-term forecasting, spanning from 30 min to 6 hours \cite{ren2015ensemble}, is useful for renewable energy integrated power systems operation and management \cite{ahmed2020review}; (3) Medium-term forecasting, covering horizons from 6 to 24 hours \cite{ren2015ensemble}, is essential for power system electro-mechanical machinery maintenance scheduling \cite{das2018forecasting}; (4) Long-term forecasting, for predicting 24 hours in advance or more \cite{ren2015ensemble}, is suitable for long-term  power generation, transmission and distribution scheduling and in power market bidding and clearing \cite{behera2018solar}. For the selection of solar forecasting methods, temporal and spatial resolution are the critical factors for consideration \cite{barbieri2017very}. Ground-based sky imagers are suitable for very short-term forecasting at a single location or nearby locations, given its high temporal (from seconds to minutes) and spatial resolution (<1x kms) \cite{van2018review}. Satellite and numerical weather prediction (NWP) both have coarse temporal (from minutes to 10x hours for satellite, from minutes up to 1000x hours for NWP) and spatial resolution (1x$\sim$100x kms) \cite{van2018review}. Thus, satellite fits better for short- to medium-term forecasting, while NWP is more useful for medium- to long-term forecasting. Both of those methods are more useful at larger scales. Although all of these methods may become important to the operation of power systems, in this review, we focus on ground-based sky images for very-short forecasting.

Sky-image-based solar forecasting has become more popular since 2011 \cite{chow2011intra}. Early works tend to first extract features from ground-based sky images, such as red-to-blue ratio (RBR), cloud coverage and cloud motion vectors, and then use these features for building physical deterministic models \cite{chow2011intra,marquez2013intra,quesada-ruizCloudtrackingMethodologyIntrahour2014a} or training machine learning models such as artificial neural networks \cite{fu2013predicting,chuHybridIntrahourDNI2013a,Chu2015realtime,Chu2015reforcast,pedroAdaptiveImageFeatures2019}. In addition, several all-sky cameras can be used in stereo-vision mode to model the cloud cover in three dimensions to provide local irradiance maps \cite{peng3DCloudDetection2015, blancShorttermForecastingHigh2017a, kuhnValidationAllskyImager2018a, Kuhn_2019, Blum_2021}. In recent 5 years, with the development of computer vision techniques, efforts have been shifted to build end-to-end deep learning models to predict irradiance or PV power output based on a historical sky image sequence, which generally achieve state-of-the-art performance despite some limitations in their anticipation ability \cite{palettaBenchmarkingDeepLearning2021}. These deep learning models are mainly based on convolutional neural networks (CNNs), either solely using CNNs \cite{Sun2019,Venugopal2019,Feng2020, palettaConvolutionalNeuralNetworks2020, Feng2022} or hybridizing CNNs with recurrent neural networks (RNNs), such as LSTM \cite{Zhang2018,palettaBenchmarkingDeepLearning2021,Paletta2021eclipse}. 

Asides from direct forecast of irradiance or PV power output, cloud analysis or modeling using ground-based sky images has also attracted wide attention as a parallel avenue of research. Clouds are one of the most important factors that affect surface irradiance and PV power generation. Research on clouds using sky images can potentially contribute to the development of more accurate and robust solar forecasting models \cite{barbieri2017very}. Existing efforts include but not limited to cloud segmentation, cloud classification and cloud motion prediction. 

Cloud segmentation is one of the first steps for sky/cloud image analysis \cite{SWIMSEG2017}, from which cloud pixels are identified and cloud coverage, cloud shadow projection and various cloud features can be derived for further research. Traditional methods tend to distinguish cloud and clear sky pixels by applying a threshold, either fixed or adaptive, on features extracted from the red and blue channels of sky images. This is based on the fact of different scattering behaviors of red and blue bands of the solar beam when encountering air molecules (Rayleigh scattering) and cloud particles (Mie scattering). Such features include red-blue difference \cite{heinle2010automatic,liu2014automatic}, red-blue ratio \cite{long2006retrieving,ghonima2012method}, normalized red-blue ratio \cite{li2011hybrid,chauvin2015cloud,Nie2020}, saturation \cite{souza2006simple} and Euclidean Geometric Distance \cite{neto2010use}. Besides manually adjusting the thresholds of these color features, learning-based methods have been used in recent years. Deep learning models, featured by CNN-based architectures such as U-Net, have seen increasing popularity \cite{dev2019cloudsegnet,xie2020segcloud,zhang2021ground,fabel2022applying} and shown the highest capability according to a benchmark study by \citet{hasenbalg2020benchmarking}. 

Identification of cloud categories is also important. Different clouds can cause different extents of attenuation of solar irradiance, e.g., low-level clouds like cumulus block sunlight more significantly than high-level clouds like cirrus \cite{barbieri2017very}. Different classification criteria are observed in existing studies, which are largely based on: (1) the main cloud genera recommended by the World Meteorological Organization \cite{zhuo2014cloud,CCSN2018,ye2019supervised} (e.g., Cirrus, Cumulus, Stratus, Nimbus), (2) the visual characteristics of clouds \cite{SWIMCAT2015,li2016pixels,luo2018ground} (e.g., patterned clouds, thick dark clouds, thin white clouds, veil clouds), or (3) the height of clouds \cite{fabel2022applying} (e.g., high-level clouds, medium-level clouds, low-level clouds). Most studies are learning-based \cite{fabel2022applying}, and a common workflow is feature extraction followed by classification. Different classifiers are used, including machine learning models, such as k-nearest neighbors, support vector machines \cite{heinle2010automatic,zhuo2014cloud,ye2017deepcloud} and in recent years end-to-end deep learning models such as CNNs for cloud classification have emerged as a promising approach \cite{ye2017deepcloud,CCSN2018,fabel2022applying}. 

A better cloud motion prediction can potentially help improve the anticipation skill of the current solar forecasting models \cite{paletta2022omnivision}. Image-based motion estimation has been studied broadly. Traditional methods based on patch matching methods, e.g., particle image velocimetry \cite{willert1991digital} and optical flow \cite{beauchemin1995computation}, to identify wind vectors and linearly extrapolating cloud motion, are usually less satisfactory as clouds are deformable and highly volatile \cite{Huang2013,dev2016short}. More recently, efforts have shifted to using deep learning models to predict cloud motion by generating future cloud images with end-to-end data-driven training \cite{andrianakos2019sky,LeGuen2020_solar}, which shows better performance in capturing the cloud dynamics although the generated images look blurry and generally do not consider the stochasticity of cloud motion.

Deep learning methods have recently shown promising performance in solar forecasting and related fields mentioned above. However, one of the major challenges for deep-learning-based models is the lack of high-quality data. This is especially problematic because deep learning models are often data hungry. Data collection is costly due to expensive recording devices and costs for human operators for regular maintenance, upkeep, and replacement of these devices. Making deep learning models generalize well often requires massive and diversified training data. In recent years, the increasing release of sky image datasets have provided great opportunities for researchers, while the exposure of these datasets might be limited.

Although there are numerous efforts \cite{inman2013solar,diagne2013review,antonanzas2016review,voyant2017machine,barbieri2017very,sobri2018solar,das2018forecasting,yang2018history,kumar2020solar,li2020review,guermoui2020comprehensive,ahmed2020review,wang2020taxonomy,alkhayat2021review,sharma2021review,kumari2021deep,chu2021intra,lin2022recent} on reviewing different aspects of solar forecasting, including methodology, performance evaluation metrics, major challenges, technological infrastructures, etc., to our best knowledge, there are currently no reviews specifically focused on open-source sky image datasets. To fill this gap, this study presents a comprehensive survey and evaluation of open-source sky image datasets for solar forecasting, cloud analysis (i.e. cloud segmentation, classification) and modeling (i.e. cloud motion prediction). Although we were stringent in searching for such datasets, there is no guarantee that we obtained all such datasets nor that cover all aspects of the datasets correctly. We encourage the researchers to use this study as a resource for locating datasets which they can access for additional details. It should be noted that in this study, our main focus is on the datasets that are suitable for use in machine/deep learning methods for solar forecasting and related research areas, including cloud classification, cloud segmentation and cloud motion prediction using sky images, while these datasets could also be used with other traditional methods or even in completely different applications.

The rest of this paper is organized as follows: in Section \ref{sec:method}, we describe the search processes to identify the open-source sky image datasets from different sources on the web, the information we collect from these datasets as well as the multi-criteria ranking system we developed to evaluate the datasets. Section \ref{sec:results} gives results, first presenting an overview of all datasets, following by a detailed description of features of the datasets, including data inclusion and specifications, spatial and temporal coverage, and a usage analysis of these datasets by the scientific community. Next, we show a comprehensive evaluation of the datasets using the multi-criteria ranking system. Finally, we summarize the findings and provides insights regarding the choices of datasets in Section \ref{sec:conclusion}. The detailed data specifications of each dataset and the references of studies using the datasets are provided in Appendix \ref{sec:appendixA} and \ref{sec:appendixB}, respectively.

% Use the style of numbering in square brackets.
% If nothing is used, default style will be taken.
%\begin{enumerate}[1)]
%\item 
%\item 
%\item 
%\end{enumerate}  

%\begin{itemize}
%\item 
%\item 
%\item 
%\end{itemize}  

%\begin{description}
%\item[]
%\item[] 
%\item[] 
%\end{description}  

% Uncomment and use as the case may be
%\begin{theorem} 
%\end{theorem}

% Uncomment and use as the case may be
%\begin{lemma} 
%\end{lemma}


\section{Method}\label{sec:method}

\subsection{Dataset search}
%How do we find the datasets? \\
%1. Potential sources for dataset search \\
%2. What databases/search engine did we use to do the search? %\\
%3. What keywords did we use to search? \\
%4. Dataset screening criteria \\

For datasets search, we consider literature sources including peer-reviewed journal articles, conference proceedings and pre-prints, popular open research data repositories including Mendeley Data, DRYAD and Zenodo, as well as the open-source platform GitHub. Additionally, sky image datasets from research campaigns of Atmospheric Radiation Measurement (ARM) program managed by the US Department of Energy (DOE) are included. ARM is a multi-platform scientific user facility equipped with instruments collecting ground-based measurements of atmospheric data at various locations around the world. GitHub is mainly used for code sharing and documentation and might not be often used for data storage especially when the dataset size is large, and we thus consider it more as a location for dataset tracking rather than a deposit.

Different search engines are used for the different types of sources. For literature search, Google Scholar is used as it often provides comprehensive coverage of resources in various scientific disciplines \cite{harzing2013preliminary}. \citet{yang2018history} conducted a text mining analysis on Google scholar data to establish the technological infrastructure and identify recent key innovations for solar forecasting. For research data repositories, search engines are provided on their websites. It should be noted that Mendeley Data is a comprehensive search platform, and it provides the options to include datasets from other platforms like DRYAD and Zenodo in the search results. However, the search outcome might include different versions of the same dataset, therefore, instead of using the search returned by Mendeley Data for DRYAD and Zenodo, we conduct search separately on these two platforms. For GitHub, search are conducted using Github Search. ARM Data Center \footnote{\url{https://adc.arm.gov/discovery/\#/}} provides an easy access to various ground-based measurement data.

Search key words were defined and employed for searching datasets. Table \ref{tab:search_string} summarizes the search strings used in different search engines. Given the syntax and the scope of the search engines, Google Scholar, Mendeley Data and Zenodo share the same search strings, while the other three apply individual search strings. Also, instead of repeatedly defining similar words, an asterisk (*) is added to a word as a wildcard for variant versions (e.g., sky image* can retrieve sky image, sky images, sky imagery, sky imager, etc.) if it is a valid syntax in that search engine.

\begin{table}[!htp]
\caption{The initial searching strings applied in different bases}
\label{tab:search_string}
\begin{tabular}{lll}
\toprule
\multicolumn{1}{c}{\textbf{\#}} & \multicolumn{1}{l}{\textbf{Base}} & \multicolumn{1}{l}{\textbf{Initial Search String}} \\ \midrule
1 & ARM Data Center & "Sky imager" \\ \midrule 
2 & Github Search & ("Sky image*" OR "Sky patch" OR "Fish-eye camera") AND (Dataset* OR Database*) \\ \midrule
3 & Google Scholar & \begin{tabular}[c]{@{}l@{}}\textbf{Solar forecasting:}\\ (“Sky image*” OR “Sky patch*” OR “Fish-eye camera”)\\ AND (“Solar forecast*” OR “Irradiance forecast*”\\ OR “Irradiance predict*” OR “PV power forecast*” OR Nowcast* OR “PV power predict*”)\end{tabular} \\
4 & Mendeley Data & \begin{tabular}[c]{@{}l@{}}AND (Dataset* OR Database*)\\ \\ \textbf{Cloud analysis and modeling:} \\ (“Sky image*” OR “Sky patch*” OR “Fish-eye camera”)\end{tabular} \\
5 & Zenodo & \begin{tabular}[c]{@{}l@{}}AND (“Cloud segmentation” OR “Cloud detection” OR “Cloud classification” \\ OR “Cloud categorization” OR “Cloud motion predict*”  OR “Cloud movement predict*” \\ OR “Cloud forecast*”)\\ AND (Dataset* OR Database*)\end{tabular} \\ \midrule
6 & DRYAD & "Sky image", "Sky patch" \\ 
\bottomrule
\end{tabular}
\end{table}

For search engines other than Google Scholar, we can directly obtain information about the datasets based on the aforementioned search keywords. If references or links to the relevant publications are provided, we further track them for more details about the datasets. For Google Scholar, a reading inspection is first conducted based on the initial results. In general, three types of papers are found at this stage, i.e. dataset papers, research papers and review papers, and different strategies are used to track the datasets. In most cases, dataset papers describe the data collection, data specifications and use cases, etc. In addition, some dataset papers develop tools or software packages to help researchers access publicly available datasets, e.g., \cite{yang2018solardata,Feng2019Opensolar}. For this type, we trace back the datasets that the tools or software provide access to, based on the provided references. Research papers often include a section or subsection describing the dataset used in the study, e.g., a ``Data'' section exclusively describing the dataset, or a paragraph in ``Experiments'', ``Methods and Materials'' or ``Case study'' sections introducing the data used for the experiments and analysis. For review papers, a review of the datasets that are publicly available and commonly used by the solar forecasting community, is sometimes provided. We thus trace back the references of the datasets mentioned in the review for more details.

After initial results are obtained, the following 4 criteria are used to exclude datasets as out of scope for our study:
\begin{enumerate}
    \item The dataset must be open-source, i.e. everyone can download it from online data repositories, or noted in the paper that it can be requested from the authors, or it can be accessed after submitting an online form request;
    \item The dataset must contain ground-based sky images of different kinds, e.g., either whole sky images or sky patches, either in the visible or infrared spectrum, either in RGB or grayscale, either normal exposure or high dynamic range (HDR). The datasets that only contain irradiance or PV power measurements are not considered in this study, although these datasets can be used for solar forecasting solely based on the time series data, e.g., the Baseline Surface Radiation Network (BSRN) \cite{Driemel_2018}, NREL Solar Power Data for Integration Studies (SPDIS) dataset \cite{osti2016}, NIST Campus Photovoltaic Arrays and Weather Station Data Sets \cite{boyd2017nist}, National Solar Radiation Data Base \cite{sengupta2018national} and SOLETE dataset \cite{VazquezPombo2022}.
    \item It ought to contain ``labels'' that are suitable for a machine learning or deep learning setup. Such labels can be solar irradiance measurements, PV generation measurements, cloud categories, or segmentation maps (labelled manually or generated by algorithms), which can be used in a supervised learning fashion. It should be noted that it might not need to contain a label only for the case of unsupervised/self-supervised tasks, like cloud image/video prediction, i.e. predicting future sky image frames based on context frames.
    \item As additional information, some datasets contain meteorological data, like temperature, wind speed, wind direction, etc., or other data types such as sky video, satellite imagery, NWP, or PV panel parameters, etc.

\end{enumerate}

Table \ref{tab:dataset_results} shows the resulting amount of data obtained before and after applying our criteria separately. After exclusion we obtained 72 datasets, most of which mainly come from ARM and Google Scholar. It should be noted that the results returned from different search engines might have overlaps. For example, a dataset paper can be published in a journal, and meanwhile the authors could deposit the data in one of the repositories and create a GitHub page for documentation and code sharing.

%\textcolor{red}{A table here for after screening results?}
%\begin{table}[!htp]
%        \flushleft
%\caption{Datasets identified from each %platform after applying the screening %criteria}
%\label{tab:dataset_criteria_results}
%\begin{tabular}{p{0.06\textwidth}p{0.2\textwid%th}P{0.3\textwidth}}
%\hline
%\textbf{\#} & \textbf{Dataset search platform} %  & \textbf{Result after screening criteria} %\\ \hline
%1           & ARM Data Center & 26            %                           \\
%2           & Github Search   & 3             %                           \\
%3           & Google Scholar  & 46            %                           \\
%4           & Mendeley Data   & 2             %                           \\
%5           & Zenodo          & 6             %                           \\
%6           & DRYAD           & 1             %                           \\ \hline
%\end{tabular}
%\end{table}

\begin{table}[htb!]
\begin{threeparttable}
        \flushleft
\caption{Initial search results and results after applying the screening criteria for different search engines}
\label{tab:dataset_results}
\begin{tabular}{p{0.1\textwidth}p{0.2\textwidth}p{0.25\textwidth}p{0.35\textwidth}}
\hline
\textbf{\#} & \textbf{Search engine}   & \textbf{Initial search results} & \textbf{Results after applying screening criteria} \\ \hline
1           & ARM Data Center & 26                      & 26                                       \\
2           & Github Search   & 3                       & 3                                        \\
3           & Google Scholar  & 1364                    & 50                                       \\
4           & Mendeley Data   & 2403                    & 4                                        \\
5           & Zenodo          & 883                     & 7                                        \\
6           & DRYAD           & 4                       & 1                                        \\ \hline
\end{tabular}
\begin{tablenotes}[flushleft]
    \item Note that the results returned from different search engines might have overlaps, i.e. search results from different search engines point to the same dataset.
\end{tablenotes}
\end{threeparttable}
\end{table}

\subsection{Information collection}
\label{subsec:info_collect}
%What information are we trying to collect from the dataset?\\
%1. Bibliographical information \\
%2. Data inclusion \\
%3. Collection period \\
%4. Geographic location \\
%5. Access to the dataset \\
%6. Track studies cite and use the datasets \\

\begin{figure}[htb!]
	\centering		\includegraphics[width=.65\textwidth]{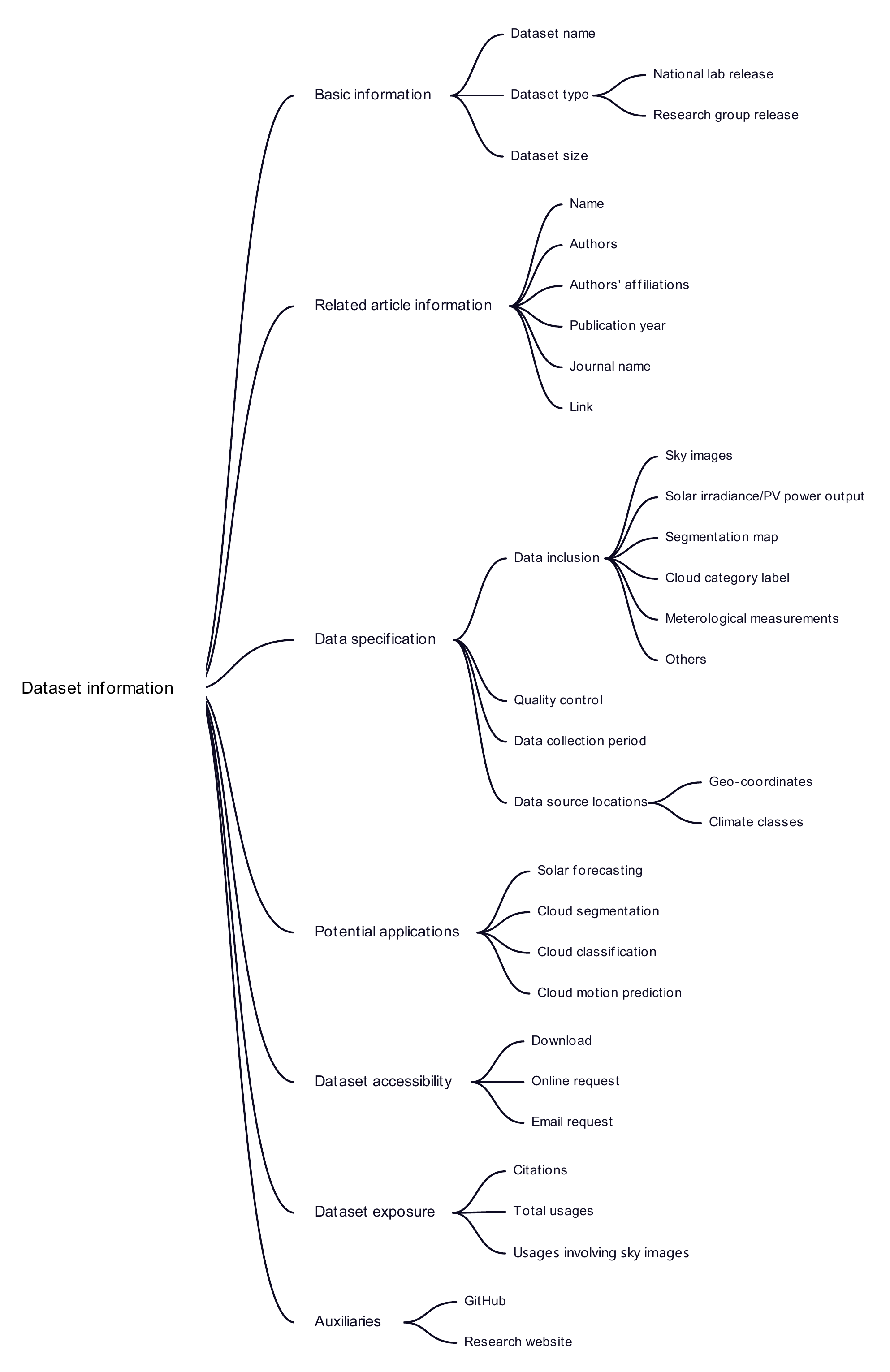}
	  \caption{Dataset information collection in different dimensions.}\label{fig:dataset_info_collection}
\end{figure}

After identifying the valid datasets, we collect information about these datasets and organize them as columns in an Excel spreadsheet. An illustration of the information being collected is shown in Figure \ref{fig:dataset_info_collection}. In general, 7 dimensions are considered, including (1) basics of the dataset; (2) related article bibliographic information; (3) data specifications; (4) potential applications; (5) dataset accessibility; (6) dataset usage and (7) auxiliaries. 

\begin{figure}[htb!]
	\centering
		\includegraphics[width=.85\textwidth]{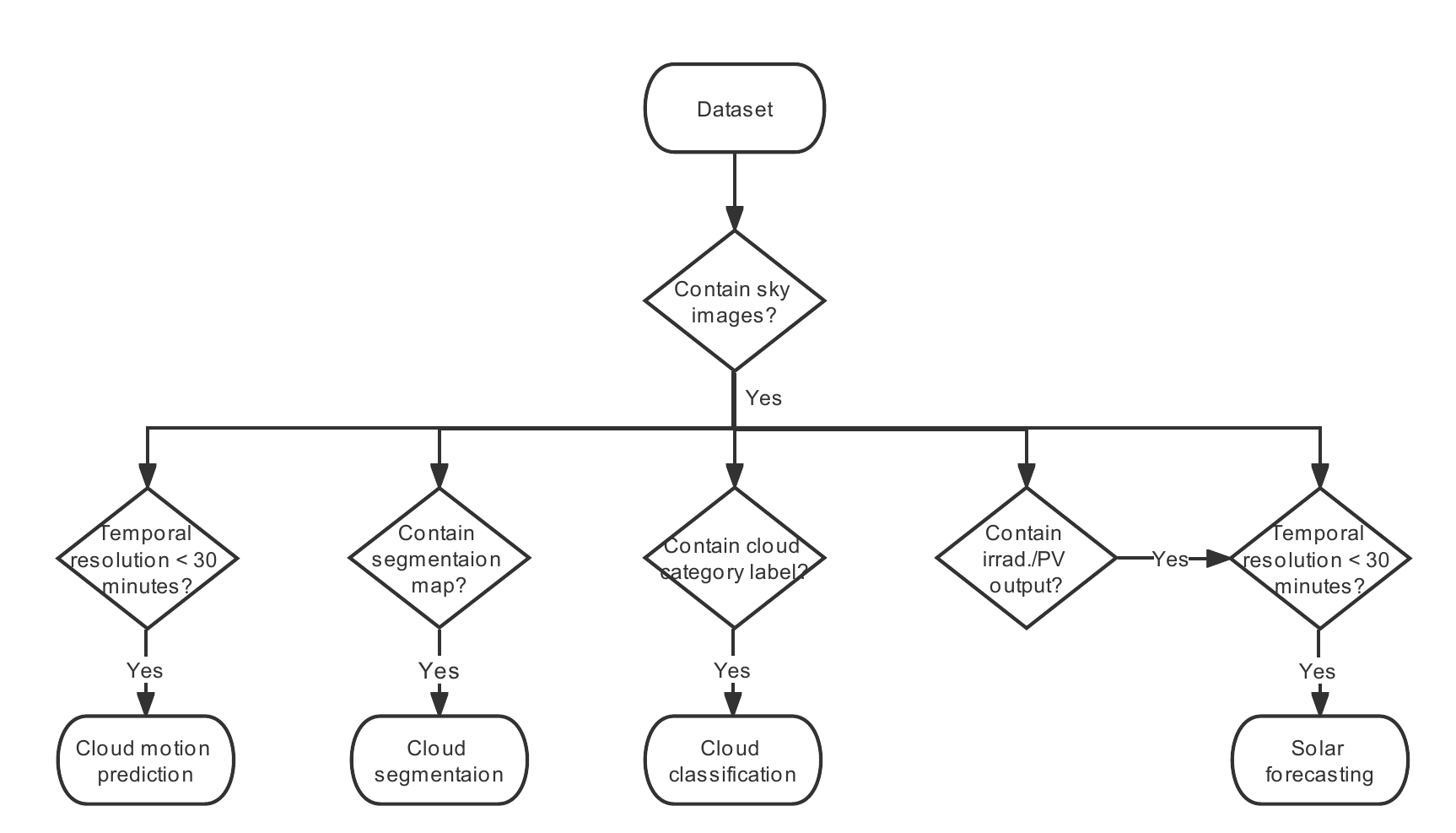}
	  \caption{Flow chart for inferring the dataset potential applications.}\label{fig:dataset_potential_apps}
\end{figure}

For datasets which do not provide any information on the dataset size, we estimate it based on the average file size (calculated based on a sample of days of data) and the data collection period. For datasets in which the data collection is still continuing, we present in this study the dataset size until July 1st, 2022. For potential applications of the datasets, we infer them based on the data inclusion and the data temporal resolution. Figure \ref{fig:dataset_potential_apps} shows how we infer the potential applications of the dataset and the inference is based on the premise that all the datasets identified contain sky images. Cloud segmentation and cloud classification can be easily inferred, as in general, if a dataset contains segmentation maps, it can be used for cloud segmentation, and similarly, if a dataset contains cloud category labels, it can be used for cloud classification. More steps are needed for checking if a dataset is suitable for solar forecasting or cloud motion prediction. Here we focus on very short-term prediction, namely, forecasting horizon less than 30 minutes. Very short-term prediction requires high temporal resolution data, so the datasets need to have temporal resolutions of at least less than 30 minutes. For dataset exposure, we consider three metrics: citations, total usage, and usage involving sky images. We track the citations of the datasets based on the citation counts of the related dataset articles returned by Google scholar, again based on the counts until July 1st, 2022. We only consider the papers written in English, so the non-English citations are not considered, and we also removed the repeated citations returned by Google Scholar as it sometimes returns different versions of the same paper that cite the datasets. Regarding the usage of the dataset, we conducted a reading inspection for each paper that cites the dataset to determine whether these papers actually use the dataset in their studies or just cite the related work. We also track the data that are used by these studies as well as the research topics or methods of these citing studies, to verify whether these studies actually use sky images and irradiance measurements to train deep learning models or only use the irradiance values for building time series models? We are mainly interested in the former type of studies in this survey.

\subsection{Dataset evaluation}
\label{subsec:dataset_evaluation_criteria}
Data are critical to the development of deep learning models. Not only the quantity or amount of data matter, but also the quality of the data. We develop a multi-criteria ranking system to evaluate the identified ground-based sky datasets. For a comprehensive evaluation, 8 dimensions of the datasets which could potentially have important impacts are considered. Table \ref{tab:dataset_eval_criteria} shows the ranking system for evaluating the datasets. The ranking system provides a semi-subjective guideline, i.e. the selection of the criteria are subjective, while the results are based on objective statistical information on the datasets. For each dimension, we set the highest rank to 10, and the lowest rank to 2, with other mid-levels in between. If the information is not available for certain dimensions, they are marked as N/A. 

It should be noted that the criteria are partly application dependent as each usage might favor slightly different data characteristics. Solar forecasting and cloud motion prediction are evaluated using the same criteria as both of these two tasks require the input of an image sequence with a rather high temporal resolution, while the other two tasks, e.g. cloud segmentation and cloud classification, are built upon one-on-one correlation between images and labels and do not require the input of an image sequence. To this end, we only evaluate the temporal resolution of datasets for solar forecasting and cloud motion prediction, and not for cloud segmentation and classification. 

Also, the datasets used for cloud segmentation and classification generally contain very specific data types, i.e. images plus segmentation maps or images plus cloud category labels, while other data types like meteorological measurements are often missing. Therefore, the comprehensiveness criteria is not applied to datasets for cloud segmentation and classification. Since each segmentation map or cloud category label is generated by human experts, it is reasonable to assume that the data are quality checked for cloud segmentation and classification, and thus we do not apply this criteria for these two applications. The number of samples is usually reported for cloud segmentation and classification datasets, but not so frequently for solar forecasting and cloud motion prediction datasets. So we evaluate this criteria for cloud segmentation and classification, but not for solar forecasting and cloud motion prediction. One can get an estimate of the number of samples based on temporal coverage and temporal resolution of the data.

\begin{enumerate}
    \item Comprehensiveness, essentially the broadness of data types included in the dataset. In this study, we potentially have the following data types: sky images, solar irradiance/PV power generation, segmentation map, cloud category, meteorological measurements, and other data such as sky video, satellite images, NWP and extracted features from imagery or time series data. A comprehensive dataset is expected to provide more information and suit diverse research purposes.
    \item Quality control, which is important as measurements with errors (e.g., abnormal negative irradiance measurements) or irrelevant information (e.g., birds, water drop) can be detrimental to the model training. 
    \item Temporal coverage. A measure of the data collection period. A multi-year effort in data collection can cover the seasonal and annual variations, which are more representative of the local climate patterns.
    \item Spatial coverage, i.e., collecting data from a single location or multiple locations. Different locations might have different climate conditions. Diversified samples for model training can help improve the model generalization.
    \item Temporal resolution. The sample interval, how frequent the data is collected, e.g., minutely, hourly, or daily. High temporal resolution data is more suitable for the case of very-short-term solar forecasting, as low temporal resolution data might miss key ramp events, which happens within a short time frame, but cause significant changes of irradiance or PV power output.
    \item Image resolution, essentially the image pixels resolution. A higher resolution sky image potentially provides more information about the sky. Although it is not necessarily better using high resolution images for model training as the computation cost increases significantly, it provides more flexibility in research. Users can customize their research by figuring out how to utilize the high resolution information without increasing computation burdens too much. For example, using a crop of the key areas of the high resolution images \cite{palettaSPINSimplifyingPolar2021}, downsampling, or fusing high resolution images with low resolution for solar forecasting model training.
    \item Number of samples. We give a higher rank for more data to encourage enough samples for model training without considering other aspects such as the diversity or representativeness of data which are difficult to ascertain and somewhat context or application dependent.
    \item Dataset usage, which are measured by three metrics in this study, namely, citations (C), total usage (U), and usage involving sky images (USI). Citation and total usage reflect the broad impact of a dataset, while USI measures how often a dataset is used in image-based modeling. It should be noted that all these metrics are based on the statistics returned by Google Scholar search of the dataset related publication. Although we present all three metrics, we mainly focus on the metric USI for evaluating the fitness of a certain dataset to be used in solar forecasting and cloud analysis/modeling.
\end{enumerate}

\begin{table}[htb!]
\begin{threeparttable}
\caption{Multi-criteria ranking system for dataset evaluation.}
\label{tab:dataset_eval_criteria}
\begin{tabular}{p{0.7\linewidth}p{0.1\linewidth}}
\toprule
Common criteria for all four potential applications: SF, CS, CC and CMP & \\  \midrule
\hspace{2mm}Temporal coverage & Rank\\ \hline
\hspace{4mm}Have more than 3 years of data & 10 \\
\hspace{4mm}Have 2 to 3 years of data & 8 \\
\hspace{4mm}Have 1 to 2 years of data & 6 \\
\hspace{4mm}Have 6 months to 1 year of data& 4 \\
\hspace{4mm}Have less than 6 months of data & 2 \\
\midrule
\hspace{2mm}Spatial coverage & Rank \\ \hline
\hspace{4mm}Collect data from 5 or more different sites & 10 \\
\hspace{4mm}Collect data from 4 different sites & 8 \\
\hspace{4mm}Collect data from 3 different sites & 6 \\
\hspace{4mm}Collect data from 2 different sites & 4 \\
\hspace{4mm}Collect data from 1 site & 2 \\
\midrule
\hspace{2mm}Image resolution & Rank \\ \hline
\hspace{4mm}Pixel resolution$\geq$1024$\times$1024 & 10 \\
\hspace{4mm}512$\times$512$\geq$Pixel resolution$<$1024$\times$1024 & 8 \\
\hspace{4mm}256$\times$256$\geq$Pixel resolution$<$512$\times$512 & 6 \\
\hspace{4mm}128$\times$128$\geq$Pixel resolution$<$256$\times$256 & 4 \\
\hspace{4mm}Pixel resolution$<$128$\times$128 & 2 \\
\midrule
\hspace{2mm}Dataset usage &  Rank \\ \hline
\hspace{4mm}$USI>15$ & 10 \\
\hspace{4mm}$10<USI\leq15$ & 8 \\
\hspace{4mm}$5<USI\leq10$ & 6 \\
\hspace{4mm}$1<USI\leq5$ & 4 \\
\hspace{4mm}$USI\leq1$ & 2 \\
\midrule
Unique criteria for individual applications & \\ \hline
\hspace{2mm}Data comprehensiveness (SF, CMP) & Rank\\ \hline
\hspace{4mm}Contains sky images, labels$^*$, meteorological measurements and other data$^{**}$ & 10 \\
\hspace{4mm}Contains sky images, labels$^*$, meteorological measurements or other data$^{**}$ & 6 \\
\hspace{4mm}Contains sky images and labels$^*$ & 2 \\
\midrule
\hspace{2mm}Data quality control (SF, CMP) & Rank \\ \hline
\hspace{4mm}Data is quality controlled & 10 \\
\hspace{4mm}No information released regarding the data quality & 2 \\
\midrule
\hspace{2mm}Temporal resolution (SF, CMP) & Rank\\ \hline
\hspace{4mm}Sample frequency$\leq$1 minute & 10 \\
\hspace{4mm}1$<$Sample frequency$\leq$3 minutes & 8 \\
\hspace{4mm}3$<$Sample frequency$\leq$5 minutes & 6 \\
\hspace{4mm}5$<$Sample frequency$\leq$10 minutes & 4 \\
\hspace{4mm}10$<$Sample frequency$\leq$30 minutes & 2 \\
\midrule
\hspace{2mm}Number of samples (CS, CC) & Rank\\ \hline
\hspace{4mm}Number of samples$>$10000  & 10 \\
\hspace{4mm}5000$<$Number of samples$\leq$10000  & 8 \\
\hspace{4mm}1000$<$Number of samples$\leq$5000  & 6 \\
\hspace{4mm}500$<$Number of samples$\leq$1000  & 4 \\
\hspace{4mm}Number of samples$\leq$500  & 2 \\
%\midrule
%\hspace{2mm}Segmentation map quality (CS) & %Rank\\ \hline
%\hspace{4mm}Segmentation map labeled by human %expert & 10 \\
%\hspace{4mm}Mix of human labeling and algorithm %generation & 5 \\
%\hspace{4mm}Segmentation map generated by %algorithms & 1 \\
\bottomrule
\end{tabular}
\begin{tablenotes}[flushleft]
\item $^*$ Labels can be either solar irradiance or PV power generation. $^{**}$ Other data can be cloud fraction values derived from sky images, satellite imagery, NWP, or PV panel parameters, etc.
\end{tablenotes}
\end{threeparttable}
\end{table}

\section{Results and discussion}
\label{sec:results}
\subsection{Datasets overview}
After the initial screening, we have identified a total of 72 open-source ground-based sky image datasets, along with various sensor measurements and data labels such as segmentation maps and cloud categories. The complete list of the datasets can be found in Table \ref{tab:dataset_basic_info}, with some basic information provided, including the dataset type based on its releasing party, the general data types included in the dataset, whether or not the dataset is quality controlled, the temporal resolution of image samples and the potential applications of the dataset inferred from the data inclusion and the temporal resolution of data.

\begin{ThreePartTable}
\begin{TableNotes}[flushleft]
\item Type --- O: datasets released by national labs or scientific organizations; R: datasets released by research groups from universities. Data inclusion --- TSI: total sky image; ASI: all sky image; Irrad.: solar irradiance measurements; MM: meteorological measurements, e.g., temperature, pressure, humidity, wind speed, wind direction; SPI: sky patch image; CCL-n: n-level cloud category/cover labels; SM-n(A): n-level segmentation map generated by algorithms; SM-n(H): n-level segmentation map labeled by human; HDR: high dynamic range; SI: satellite imagery; NWP: numerical weather prediction; FE: feature extractions from irradiance and imagery data; PVP: PV power generation measurements; PParam.: PV panel parameter measurements include voltage, current and temperature; ASV: All sky video. QC --- quality control. %) --- N/A: no information released regarding data quality control. %Accessibility --- D: downloadable; OFR: online form request; ER: email request. 
Image temp. res. --- temporal resolution of image samples. Application --- SF: image-based solar forecasting; CS: cloud segmentation; CC: cloud classification; CMP: cloud motion prediction; \ding{108}: the dataset is suitable for a certain application; \ding{109}: the dataset can theoretically be used for a certain application, but might not be preferred (e.g., segmentation maps generated by human is preferred over those generated by algorithms). $^*$ SRRL-BMS has a live-view of ASI updated minutely on its website but not archived. $^{**}$ The Girasol dataset article \cite{Girasol2021} does not mention that it includes segmentation maps or cloud category labels. However, follow-up studies by the authors \cite{terren2021segmentation,terren2021comparative} used 12 segmentation maps labeled manually and 8200 labeled images with 4 different sky conditions. Contact the authors for details. $^{***}$ most of the datasets intended for use in cloud segmentation and classification generally do not provide temporal resolution as such datasets do not require image data in sequence.
\end{TableNotes}

\begin{longtable}{lcc>{\raggedright}p{0.26\linewidth}cP{0.07\linewidth}cccc}
\caption{List of open-source ground-based sky image datasets} \label{tab:dataset_basic_info} \\

\toprule
\multirow{3}{*}{Dataset} & \multirow{3}{*}{Year} & \multirow{3}{*}{Type} & \multirow{3}{*}{Data inclusion} & \multirow{3}{*}{QC} & \multirow{3}{0.05\textwidth}{\centering Image temp. res.} & \multicolumn{4}{c}{\multirow{2}{*}{Potential applications}} \\ 
\noalign{\vskip -1mm}
 &  &  &  &  &  &  &  &  &  \\ \cline{7-10} % Table header row 
\noalign{\vskip 1mm}
 &  &  &  &  &  & SF & CS & CC & CMP \\ % Table header row
\midrule 
\endfirsthead

\caption {List of open-source ground-based sky image datasets (continued)}\\
\toprule
\multirow{3}{*}{Dataset} & \multirow{3}{*}{Year} & \multirow{3}{*}{Type} & \multirow{3}{*}{Data inclusion} & \multirow{3}{*}{QC} & \multirow{3}{0.05\textwidth}{\centering Image temp. res.} & \multicolumn{4}{c}{\multirow{2}{*}{Potential applications}} \\ 
\noalign{\vskip -1mm}
 &  &  &  &  &  &  &  &  &  \\ \cline{7-10} % Table header row 
\noalign{\vskip 1mm}
 &  &  &  &  &  & SF & CS & CC & CMP \\ % Table header row
\midrule 
\endhead

\hline
\multicolumn{10}{r}{{Continued on next page}} \\ \hline
\endfoot

\bottomrule
\insertTableNotes
\endlastfoot

SRRL-BMS\cite{SRRL1981} & 1981 & O & TSI, ASI, Irrad., MM, SM-2(A) & Yes & 10 min$^*$ & \ding{108} & \ding{109} & & \ding{108} \\
SURFRAD \cite{SURFRAD2000} & 2000 & O & TSI, Irrad., MM, SM-2(A) & Yes & 1 hr &  & \ding{109} & &  \\ 
SIRTA \cite{SIRTA2005} & 2005 & O & TSI, ASI, Irrad., PVP, MM, SM-2(A) & Yes & 1-2 min & \ding{108} & \ding{109} & & \ding{108} \\
ARM-MASRAD \cite{ARM_MASRAD2005} & 2005 & O & TSI, Irrad., MM, SM-2(A) & Yes & 30 sec & \ding{108} & \ding{109} & & \ding{108} \\
ARM-RADAGAST \cite{ARM_RADAGAST2008} & 2008 & O & TSI, Irrad., MM, SM-2(A) & Yes & 30 sec & \ding{108} & \ding{109} & & \ding{108} \\
ARM-STORMVEX \cite{ARM_STORMVEX2010} & 2010 & O & TSI, Irrad., MM, SM-2(A) & Yes & 30 sec & \ding{108} & \ding{109} & & \ding{108} \\
HYTA (Binary) \cite{li2011hybrid} & 2011 & R & SPI, SM-2(H) & N/A & N/A$^{***}$ & & \ding{108} &  & \\
ARM-AMIE-GAN \cite{ARM_AMIE_GAN2011} & 2011 & O & TSI, Irrad., MM, SM-2(A) & Yes & 30 sec & \ding{108} & \ding{109} & & \ding{108} \\
ARM-COPS \cite{ARM_COPS2011} & 2011 & O & TSI, Irrad., MM, SM-2(A) & Yes & 30 sec & \ding{108} & \ding{109} & & \ding{108} \\
ARM-HFE \cite{ARM_EAST_AIRC2011} & 2011 & O & TSI, Irrad., MM, SM-2(A) & Yes & 30 sec & \ding{108} & \ding{109} & & \ding{108} \\
ARM-GVAX \cite{ARM_GVAX2013} & 2013 & O & TSI, Irrad., MM, SM-2(A) & Yes & 30 sec & \ding{108} & \ding{109} & & \ding{108} \\
SWIMCAT \cite{SWIMCAT2015} & 2015 & R & SPI, CCL-5 & N/A & N/A$^{***}$ & & & \ding{108} & \\
HYTA (Ternary) \cite{dev2015multi} & 2015 & R & SPI, SM-3(H) & N/A & N/A$^{***}$ & & \ding{108} & & \\
ARM-CAP-MBL \cite{ARM_CAP_MBL2015} & 2015 & O & TSI, Irrad., MM, SM-2(A) & Yes & 30 sec & \ding{108} & \ding{109} & & \ding{108} \\
ARM-TCAP \cite{ARM_TCAP2015} & 2015 & O & TSI, Irrad., MM, SM-2(A) & Yes & 30 sec & \ding{108} & \ding{109} & & \ding{108} \\
TCIS \cite{li2016pixels} & 2016 & R & ASI, CCL-5 & N/A & N/A$^{***}$ &  &  & \ding{108} & \\
NCU \cite{cheng2017cloud} & 2017 & R & ASI, SM-2(H) & N/A & N/A$^{***}$ &  & \ding{108} &  & \\
ARM-BAECC \cite{BAECCA2016} & 2016 & O & TSI, Irrad., MM, SM-2(A) & Yes & 30 sec & \ding{108} & \ding{109} & & \ding{108} \\
ARM-ACAPEX \cite{ARM_ACAPEX2016} & 2016 & O & TSI, Irrad., MM, SM-2(A) & Yes & 30 sec & \ding{108} & \ding{109} & & \ding{108} \\
ARM-MAGIC \cite{ARM_MAGIC2016} & 2016 & O & TSI, Irrad., MM, SM-2(A) & Yes & 30 sec & \ding{108} & \ding{109} & & \ding{108} \\
ARM-GoAmazon \cite{ARM_GoAmazon2016} & 2016 & O & TSI, Irrad., MM, SM-2(A) & Yes & 30 sec & \ding{108} & \ding{109} & & \ding{108} \\
ARM-NSA \cite{ARM_NSA2016} & 2016 & O & TSI, Irrad., MM, SM-2(A) & Yes & 30 sec & \ding{108} & \ding{109} & & \ding{108} \\
ARM-SGP \cite{ARM_SGP2016} & 2016 & O & TSI, Irrad., MM, SM-2(A) & Yes & 30 sec & \ding{108} & \ding{109} & & \ding{108} \\
ARM-TWP \cite{ARM_TWP2016} & 2016 & O & TSI, Irrad., MM, SM-2(A) & Yes & 30 sec & \ding{108} & \ding{109} & & \ding{108} \\
SWIMSEG \cite{SWIMSEG2017} & 2017 & R & SPI, SM-2(H) & N/A & N/A$^{***}$ & & \ding{108} & & \\
SWINSEG \cite{dev2017nighttime} & 2017 & R & SPI (Nighttime), SM-2(H) & N/A & N/A$^{***}$ & & \ding{108} & & \\
SHWIMSEG \cite{SHWIMSEG2018} & 2018 & R & SPI (HDR), SM-2(H) & N/A & N/A$^{***}$ & & \ding{108} &  & \\
Zenithal \cite{luo2018ground} & 2018 & R & SPI (Infrared), CCL-5 & N/A & N/A$^{***}$ & &  & \ding{108} & \\
CCSN \cite{CCSN2018} & 2018 & R & SPI, CCL-11 & N/A & N/A$^{***}$ & & & \ding{108} & \\
ARM-LASIC \cite{ARM_LASIC2018} & 2018 & O & TSI, Irrad., MM, SM-2(A) & N/A & 30 sec & \ding{108} & \ding{109} & & \ding{108} \\
NAO-CAS \cite{shi2019diurnal} & 2019 & R & ASI, SM-2(H) & N/A & N/A$^{***}$ &  & \ding{108} &  & \\
FGCDR \cite{ye2019supervised} & 2019 & R & ASI, SM-8(H) & N/A & N/A$^{***}$ & & \ding{108} & \ding{108} & \\
LES dataset \cite{caldas2019very} & 2019 & R & ASI, Irrad. & Yes & 1 min & \ding{108} & &  & \ding{108} \\
SWINySEG \cite{dev2019cloudsegnet} & 2019 & R & SPI (Day+Nighttime), SM-2(H) & N/A & N/A$^{***}$ & & \ding{108} & & \\
UCSD-Folsom \cite{UCSD2019} & 2019 & R & ASI, Irrad., MM, SI, NWP, FE & Yes & 1 min & \ding{108} & & & \ding{108} \\
UoH \cite{dandiniHaloRatioGroundbased2019} & 2019 & R & ASI, Irrad., MM & N/A & 30 sec & \ding{108} & & & \ding{108}\\
P2OA-RAPACE \cite{lothon2019elifan} & 2019 & O & ASI, Irrad., MM, SM-2(A) & Yes & 5 min & \ding{108} & \ding{109} & & \ding{108} \\
OHP \cite{OHP} & 2019 & O & TSI, Irrad., MM & Yes & 5 min & \ding{108} & & & \ding{108} \\
OPAR \cite{OPAR} & 2019 & O & ASI, Irrad., MM & Yes & N/A & \ding{108} & & & \ding{108} \\
ARM-CACTI \cite{ARM_CACTI2019} & 2019 & O & TSI, Irrad., MM, SM-2(A) & N/A & 30 sec & \ding{108} & \ding{109} & & \ding{108} \\
ARM-HOU \cite{ARM_HOU2019} & 2019 & O & TSI, Irrad., MM, SM-2(A) & N/A & 30 sec & \ding{108} & \ding{109} & & \ding{108} \\
ARM-MARCUS \cite{ARM_MARCUS2019} & 2019 & O & TSI, Irrad., MM, SM-2(A) & Yes & 30 sec & \ding{108} & \ding{109} & & \ding{108} \\
El Arenosillo \cite{trigo2005development,El_Arenosillo} & 2020 & O & ASI, Irrad., MM & Yes & N/A & \ding{108} & & & \ding{108} \\
MGCD \cite{liu2020multi} & 2020 & R & ASI, CCL-7 & N/A & N/A$^{***}$ &  & & \ding{108} & \\
GRSCD \cite{liu2020ground} & 2020 & R & ASI, CCL-7 & N/A & N/A$^{***}$ & & & \ding{108} & \\
WSISEG \cite{xie2020segcloud} & 2020 & R & ASI (HDR), SM-3(H) & Yes & N/A$^{***}$ & & \ding{108} & & \\
WMD \cite{krauz2020assessing} & 2020 & R & ASI, SM-4(H) & N/A & N/A$^{***}$ & & \ding{108} & \ding{108} & \\
CO-PDD \cite{baray2020cezeaux} & 2020 & O & ASI, Irrad., MM, SM-2(A) & N/A & 1-2 min & \ding{108} & \ding{109} & & \ding{108} \\
ARM-AWARE \cite{ARM_AWARE2020} & 2020 & O & TSI, Irrad., MM, SM-2(A) & No & 30 sec & \ding{108} & \ding{109} & & \ding{108} \\
GCD \cite{liu2021ground} & 2021 & R & SPI, CCL-7 & N/A & N/A$^{***}$ & & & \ding{108} & \\
BASS \cite{c2021feasibility} & 2021 & R & ASI (HDR), Irrad. & N/A & 1 min & \ding{108} & &  & \ding{108} \\
NIMS-KMA \cite{kim2021twenty} & 2021 & O & ASI (Day+Nighttime), CCL-10 & N/A & N/A$^{***}$ & & & \ding{108} & \\
Girasol \cite{Girasol2021} & 2021 & R & ASI (HDR Grayscale), SPI (Infrared), Irrad., MM & Yes & 15 sec & \ding{108} & \ding{108}$^{**}$ & \ding{108}$^{**}$ & \ding{108} \\
SkyCam \cite{SkyCam2021} & 2021 & R & ASI (HDR), Irrad. & N/A & 10 sec & \ding{108} & & & \ding{108} \\
SIPM \cite{SIPM2021} & 2021 & R & ASI, PVP, PParam. & N/A & 1 sec & \ding{108} & & & \ding{108} \\
SIPPMIF \cite{UOW2021} & 2021 & R & ASI, PVP & N/A & 10 sec & \ding{108} & & & \ding{108} \\
Waggle \cite{park2021prediction} & 2021 & O & SPI, PVP, Irrad., SM-2(A) & N/A & 15 sec & \ding{108} & \ding{109} & & \ding{108} \\
TAN1802 Voyage \cite{TAN1802_Voyage2021} & 2021 & O & ASI (HDR), Irrad., MM, SM-2(A) & Yes & 5 min & \ding{108} & \ding{109} & & \ding{108} \\
ARM-SAIL/GUC \cite{ARM_GUC2021} & 2021 & O & TSI, Irrad., MM, SM-2(A) & No & 30 sec & \ding{108} & \ding{109} & & \ding{108} \\
SKIPP'D \cite{nie2022skippd} & 2022 & R & ASI, ASV, PVP & Yes & 1 min & \ding{108} & & & \ding{108} \\
TCDD \cite{zhang2021ground} & 2021 & R & SPI, SM-2(H) & N/A & N/A$^{***}$ &  & \ding{108} & &  \\
PSA Fabel \cite{fabel2022applying} & 2022 & O & ASI, SM-3(H) & N/A & N/A$^{***}$ & & \ding{108} & \ding{108} & \\
NAO-CAS XJ \cite{li2022all} & 2022 & R & ASI, CCL-4 & N/A & N/A$^{***}$ & & & \ding{108}  & \\
TLCDD \cite{zhang2022ground} & 2022 & R & SPI, SM-2(H) & N/A & N/A$^{***}$ & & \ding{108} &  & \\
ACS WSI \cite{ye2022self} & 2022 & R & ASI, SM-2(H) & N/A & N/A$^{***}$ & & \ding{108} &  & \\
ARM-COMBLE \cite{ARM_COMBLE2022} & 2022 & O & TSI, Irrad., MM, SM-2(A) & Yes & 30 sec & \ding{108} & \ding{109} & & \ding{108} \\
ARM-MOSAIC \cite{ARM_MOSAIC2022} & 2022 & O & TSI, Irrad., MM, SM-2(A) & Yes & 30 sec & \ding{108} & \ding{109} & & \ding{108} \\
ARM-ACE-ENA \cite{ARM_ACE_ENA2022} & 2022 & O & TSI, Irrad., MM, SM-2(A) & Yes & 30 sec & \ding{108} & \ding{109} & & \ding{108} \\
Orion StarShoot \cite{Orion_Starshoot} & N/A & O & ASI, Irrad., MM & N/A & 1 min & \ding{108} & & & \ding{108} \\
LTR \cite{warsaw} & N/A & R & ASI, Irrad., MM & N/A & N/A & \ding{108} & & & \ding{108} \\
LOA \cite{LOA} & N/A & O & ASI, Irrad., MM & N/A & N/A & \ding{108} &  & & \ding{108} \\
ARM-OLI \cite{ARM_OLI} & N/A & O & TSI, Irrad., MM, SM-2(A) & Yes & 30 sec & \ding{108} & \ding{109} & & \ding{108} \\
\end{longtable}
\end{ThreePartTable}

The datasets are mainly released by two types of parties. The first category consists of national labs and government scientific organizations (denoted by O in the Table \ref{tab:dataset_basic_info} column Type), which often have multi-site and multi-year efforts in collecting a wide range of sensor observations for the purpose of studying the Earth's atmosphere and climate. This type of datasets generally go through stringent quality control processes. 39 out of the 72 identified datasets belong to this type, to name a few, the Solar Radiation Research Laboratory Baseline Measurement System (SRRL-BMS) \cite{stoffel1981nrel} by the National Renewable Energy Laboratory (NREL) of the US, the National Surface Radiation (SURFRAD) Budget Network by National Oceanic and Atmospheric Administration (NOAA) \cite{augustine2000surfrad} of US, a series of campaigns deployed by the Atmospheric Radiation Measurement (ARM) Program under the hood of Department of Energy of US (totally 26 datasets from ARM program, e.g., \cite{ARM_AMIE_GAN2011,ARM_CAP_MBL2015,ARM_ACAPEX2016,ARM_CACTI2019,ARM_AWARE2020,ARM_ACE_ENA2022}), and the Site Instrumental de Recherche par Télédétection Atmosphérique (SIRTA) \cite{SIRTA2005} by Institut Pierre Simon Laplace in France. The second type of party is research groups from universities (denoted by R in the Table \ref{tab:dataset_basic_info} column Type). To promote the openness and accessibility of research, researchers choose to release the datasets along with their research articles (e.g., \cite{li2011hybrid,SWIMCAT2015,CCSN2018,ye2019supervised,xie2020segcloud,c2021feasibility}) or publish specific dataset articles for their datasets (e.g., \cite{UCSD2019,Girasol2021,SkyCam2021,nie2022skippd}). Most datasets from this type of party are published for a specific task, e.g., solar forecasting or cloud segmentation or cloud classification, so the type of data included is generally less comprehensive compared with the type O datasets. Moreover, the data quality might vary from datasets to datasets. Some datasets note that they have a quality control process (e.g., \cite{UCSD2019,Girasol2021,nie2022skippd}), while others did not released any information on that (denoted by N/A in the Table \ref{tab:dataset_basic_info} column QC). It should be noted that although most of the datasets intended for cloud segmentation and classification do not mention that they have a quality control process, it is reasonable to assume that they have a quality check on the imagery data as they provide cloud segmentation maps or cloud category labels labeled by human experts and abnormal images, such as images with presence of birds, have been removed.The El Arenosillo dataset is referenced twice as it is composed of both night and day observations which were presented separately \cite{trigo2005development, El_Arenosillo}.

To give a glimpse of data types included in each dataset, we categorize the data into the following types as shown in Table \ref{tab:dataset_basic_info}: (1) imagery data, i.e. ground-based sky images of various kinds, that could serve as the input of the models for solar forecasting, cloud segmentation and classification; (2) data that could potentially be used as labels for machine learning or deep learning model development, including solar irradiance (denoted by irrad.), PV power generation (PVP), segmentation map (SM) and cloud category label (CCL); auxiliary data that could potentially serve as additional input for the model, including (3) meteorological measurements (MM) such as air temperature, wind speed and direction, humidity and pressure and (4) other data, including but not limited to sky video, numerical weather prediction (NWP), and secondary data such as features extractions (FE) from sky images or time series measurements.  It should be noted that besides using irrad., PVP, SM and CCL as labels, sky images itself can be used as labels, for example, for cloud motion prediction task, sky image sequence in the past can be used as context to forecast the sky image frame(s) in the future \cite{andrianakos2019sky,LeGuen2020_solar}, which works in a so-called self-supervised learning fashion. We present detailed descriptions of each data type in section \ref{subsec:data_inclusion_specs}.

The temporal resolution of the image samples of each dataset is also listed in the table. The detail specifications of different data types can be found in Table \ref{tab:data_spec_sf_cmp} to \ref{tab:data_spec_cc} in Appendix \ref{sec:appendixA} for different applications. It should be noted that most of the datasets intended for use in cloud segmentation and cloud classification tasks generally do not provide temporal resolution information as such datasets do not require image data in sequence as solar forecasting and cloud motion prediction do. Samples are selected from a pool mostly based on their representativenss of a certain sky condition or cloud class without considering too much on their temporal dependency. 

The potential applications for the datasets are inferred based on the data inclusion as well as the temporal resolution of data. The method we used for making the inference are shown in Figure \ref{fig:dataset_potential_apps} of Section \ref{subsec:info_collect}. 47 out of 72 datasets can potentially be used for solar forecasting as well as cloud motion prediction. 51 datasets can theoretically be used for cloud segmentation, out of which 15 datasets have segmentation maps labeled by human experts, while the rest 36 datasets have algorithm-generated segmentation maps. Segmentation maps labeled by human experts are preferred over those generated by algorithms as they generally have better quality and less noise. Therefore, for analysis of cloud segmentation datasets in the following sections, we will mainly focus on the 15 datasets with human-generated segmentation maps. 13 datasets are suitable for cloud classification and all cloud category labels are generated by human experts. It should be noted that there might be overlaps between datasets for cloud segmentation and classification. Some cloud classification datasets have a pixel-level labeling of the cloud types which can also satisfy the needs of cloud segmentation. A detail analysis of the dataset applications based on the papers that use these datasets is presented in Section \ref{subsec:data_usage_analysis}.

The specific ways of accessing each dataset, i.e. the links to the data repositories for downloading the data or/and the email addresses of the corresponding authors for requesting the datasets, as well as the data policy or license information of each dataset if available are provided in Table \ref{tab:access_to_datasets}. Although most of the datasets listed here are free and open for research use, commercial use is generally prohibited, with special notes in their data policy or enforced licenses such as CC BY-NC 4.0 \footnote{CC BY-NC 4.0: \url{https://creativecommons.org/licenses/by-nc/4.0/}}. However, certain datasets have no restriction on commercial use with a CC0 1.0 license \footnote{CC0 1.0: \url{https://creativecommons.org/publicdomain/zero/1.0/}}, e.g., CCSN\cite{CCSN2018}, Girasol \cite{Girasol2021}. We encourage the users to double check the license of the datasets or contact the creators for more details before using these datasets in publications or for commercial purposes.

\begin{landscape}
\begin{ThreePartTable}
\begin{TableNotes}[flushleft]
\item 
\end{TableNotes}
\begin{longtable}{l>{\raggedright}p{0.52\linewidth}>{\raggedright\arraybackslash}p{0.32\linewidth}}
\caption{Access to and data policy/license of the open-source sky image datasets} \label{tab:access_to_datasets} \\
\toprule
Dataset  & Access & Data policy/License information \\ 
\midrule 
\endfirsthead

\caption{Access to and data policy/license of the open-source sky image datasets (continued)}\\
\toprule
Dataset  & Access & Data policy/License information \\ 
\midrule 
\endhead

\hline
\multicolumn{3}{r}{{Continued on next page}} \\ \hline
\endfoot

\bottomrule
\insertTableNotes
\endlastfoot

SRRL-BMS \cite{SRRL1981}  &  \url{https://midcdmz.nrel.gov/apps/sitehome.pl?site=BMS} & \url{https://data.nrel.gov/submissions/7} \\ \hline
SURFRAD \cite{SURFRAD2000}  & \url{https://gml.noaa.gov/grad/surfrad/index.html} & \url{https://gml.noaa.gov/about/disclaimer.html} \\ \hline
\multirow{4}{*}{SIRTA \cite{SIRTA2005}} &   \url{https://sirta.ipsl.polytechnique.fr/index.html} & \multirow{4}{1.0\linewidth}{\url{https://sirta.ipsl.polytechnique.fr/data_policy.html}}\\
& Actris: \url{https://www.actris.fr/catalogue/} (keyword: sky imager, pyranometer) &  \\
&  Icare: \url{ftp://ftp.icare.univ-lille1.fr} (GROUND-BASED/SIRTA\_Palaiseau) &  \\
& PV data: \url{https://gitlab.in2p3.fr/energy4climate/public/sirta-pv1-data} & \\
\hline
ARM-MASRAD \cite{ARM_MASRAD2005} & \url{https://adc.arm.gov/discovery/#/results/s::sky\%20images\%20MASRAD} & \multirow{3}{1.0\linewidth}{\url{https://www.arm.gov/guidance/datause/generalguidelines}} \\ \cline{1-2}
ARM-RADAGAST \cite{ARM_RADAGAST2008}    & 
\url{https://adc.arm.gov/discovery/#/results/s::sky\%20images\%20RADAGAST} & \\ \cline{1-2}
ARM-STORMVEX \cite{ARM_STORMVEX2010}    & \url{https://adc.arm.gov/discovery/#/results/s::STORMVEX\%20sky\%20images} & \\ \hline
HYTA (Binary) \cite{li2011hybrid}  & \url{https://github.com/Soumyabrata/HYTA} & N/A \\ \hline
ARM-AMIE-GAN \cite{ARM_AMIE_GAN2011}  & \url{https://adc.arm.gov/discovery/#/results/s::sky\%20image\%20amie-gan} & \multirow{4}{1.0\linewidth}{\url{https://www.arm.gov/guidance/datause/generalguidelines}} \\ \cline{1-2}
ARM-COPS \cite{ARM_COPS2011}  &  \url{https://adc.arm.gov/discovery/#/results/iopShortName::amf2007cops} &\\ \cline{1-2}
ARM-HFE \cite{ARM_EAST_AIRC2011} &  \url{https://adc.arm.gov/discovery/#/results/s::tsiskyimage\%20HFE} &\\ \cline{1-2}
ARM-GVAX \cite{ARM_GVAX2013}  &  \url{https://adc.arm.gov/discovery/#/results/s::sky\%20images\%20gvax} & \\ \hline
SWIMCAT \cite{SWIMCAT2015}   & \url{http://vintage.winklerbros.net/swimcat.html} & CC BY-NC 4.0 (Attribution-NonCommercial 4.0) \\ \hline
HYTA (Ternary) \cite{dev2015multi}  & \url{https://github.com/Soumyabrata/HYTA}  & N/A \\ \hline
ARM-CAP-MBL \cite{ARM_CAP_MBL2015}  &  \url{https://adc.arm.gov/discovery/#/results/s::tsiskyimage\%20CAP-MBL} & \multirow{2}{1.0\linewidth}{\url{https://www.arm.gov/guidance/datause/generalguidelines}}\\ 
\cline{1-2}
ARM-TCAP \cite{ARM_TCAP2015} &  \url{https://adc.arm.gov/discovery/#/results/s::sky\%20images\%20TCAP} &\\ \hline
\multirow{3}{*}{TCIS \cite{li2016pixels}}  &  \url{http://icn.bjtu.edu.cn/visint/resources/CloudImages} & \multirow{3}{*}{N/A} \\ 
& Email Qingyong Li (\href{mailto:qingyongli@gmail.com}{qingyongli@gmail.com}), Weitao Lu (\href{mailto:wtlu@cams.cma.gov.cn}{wtlu@cams.cma.gov.cn}) if the link is not accessible &  \\ \hline
\multirow{2}{*}{NCU \cite{cheng2017cloud}}  & \url{https://drive.google.com/openid=0B38yagaBviZYNmxReVBIQkVJYkk} & \multirow{2}{*}{N/A} \\
& Email Hsu-Yung Cheng (\href{mailto:breeze.cheng@gmail.com}{breeze.cheng@gmail.com}) if the link is not accessible &  \\ \hline
ARM-BAECC \cite{BAECCA2016} & \url{https://adc.arm.gov/discovery/#/results/s::BAECC\%20sky\%20images} & \multirow{7}{1.0\linewidth}{\url{https://www.arm.gov/guidance/datause/generalguidelines}} \\  \cline{1-2}
ARM-ACAPEX \cite{ARM_ACAPEX2016}  & \url{https://adc.arm.gov/discovery/#/results/s::ACAPEX\%20sky\%20images} & \\ \cline{1-2}
ARM-MAGIC \cite{ARM_MAGIC2016} & \url{https://adc.arm.gov/discovery/#/results/s::sky\%20images\%20magic} & \\ \cline{1-2}
ARM-GoAmazon \cite{ARM_GoAmazon2016}  & \url{https://adc.arm.gov/discovery/#/results/s::sky\%20images\%20go\%20amazon}& \\ \cline{1-2}
ARM-NSA \cite{ARM_NSA2016}  & \url{https://adc.arm.gov/discovery/#/results/s::sky\%20images\%20NSA} &\\ \cline{1-2}
ARM-SGP \cite{ARM_SGP2016}  &  \url{https://adc.arm.gov/discovery/#/results/s::SGP\%20sky\%20images} &\\ \cline{1-2}
ARM-TWP \cite{ARM_TWP2016} & \url{https://adc.arm.gov/discovery/#/results/s::TWP\%20sky\%20images} & \\ \hline
SWIMSEG \cite{SWIMSEG2017}  & \url{http://vintage.winklerbros.net/swimseg.html}& \multirow{5}{1.0\linewidth}{CC BY-NC 4.0 (Attribution-NonCommercial 4.0)}\\ \cline{1-2}
\multirow{2}{*}{SWINSEG \cite{dev2017nighttime}}  & \url{http://vintage.winklerbros.net/swinseg.html}& \\
  & GitHub: \url{https://github.com/Soumyabrata/nighttime-imaging} & \\ \cline{1-2} %The code is only for academic and research purposes\\ \hline
\multirow{2}{*}{SHWIMSEG \cite{SHWIMSEG2018}}  & \url{http://vintage.winklerbros.net/shwimseg.html} &  \\
& GitHub: \url{https://github.com/Soumyabrata/HDR-cloud-segmentation} & \\ \hline %The code is only for academic and research purposes \\ \hline
Zenithal \cite{luo2018ground}  & Email Qixiang Luo (\href{mailto:qixiang_luo@aliyun.com}{qixiang\_luo@aliyun.com}) & N/A \\ 

\multirow{2}{*}{CCSN \cite{CCSN2018}} & \url{https://doi.org/10.7910/DVN/CADDPD} & \multirow{2}{*}{CC0 1.0 Universal Public Domain Dedication} \\
& GitHub: \url{https://github.com/upuil/CCSN-Database} & \\ \hline
\multirow{2}{*}{ARM-LASIC \cite{ARM_LASIC2018}} &  \multirow{2}{*}{\url{https://adc.arm.gov/discovery/\#/results/iopShortName::amf2016lasic}} & \url{https://www.arm.gov/guidance/datause/generalguidelines} \\ \hline
NAO-CAS \cite{shi2019diurnal} & Email Chaojun Shi (\href{mailto:474756323@qq.com}{474756323@qq.com}), Yatong Zhou (\href{mailto:zyt@hebut.edu.cn}{zyt@hebut.edu.cn}) & N/A  \\ \hline
\multirow{2}{*}{FGCDR \cite{ye2019supervised}} & \url{https://github.com/liangye518/FGCDR} & \multirow{2}{*}{N/A} \\
 & Email Zhiguo Cao (\href{mailto:zgcao@hust.edu.cn}{zgcao@hust.edu.cn}) if data cannot be found &  \\ \hline
\multirow{2}{*}{LES dataset \cite{caldas2019very}} & \url{http://symphony.les.edu.uy/} & \multirow{2}{*}{N/A} \\
& Email M. Caldas (\href{mailto:mcaldas@fisica.edu.uy}{mcaldas@fisica.edu.uy}) if the link is not accessible & \\ \hline
\multirow{2}{*}{SWINySEG \cite{dev2019cloudsegnet}}  & \url{http://vintage.winklerbros.net/swinyseg.html} & \multirow{2}{1.0\linewidth}{CC BY-NC 4.0 (Attribution-NonCommercial 4.0)} \\
& GitHub: \url{https://github.com/Soumyabrata/CloudSegNet} & \\ \hline
UCSD-Folsom \cite{UCSD2019}  & \url{https://doi.org/10.5281/zenodo.2826939} & CC BY-NC 4.0 (Attribution-NonCommercial 4.0) \\ \hline
UoH \cite{dandiniHaloRatioGroundbased2019}  & \url{http://observatory.herts.ac.uk/allsky/imagefind.php?c=7} & N/A \\ \hline
\multirow{4}{*}{P2OA-RAPACE \cite{lothon2019elifan}} & \url{https://p2oa.aeris-data.fr/data/} & \multirow{4}{1.0\linewidth}{\url{https://www7.obs-mip.fr/wp-content-aeris/uploads/sites/71/2021/12/data_policy_P2OA.pdf}}\\    
& Actris: \url{https://www.actris.fr/catalogue/} (keyword: sky imager) & \\
& Icare: \url{ftp://ftp.icare.univ-lille1.fr} (GROUND-BASED/P2OA\_Pic-du-Midi or P2OA\_Lannemez) & \\ \hline
\multirow{3}{*}{OHP \cite{OHP}}  & \url{https://www.actris.fr/catalogue/} (keyword: sky imager) & \multirow{3}{1.0\linewidth}{\url{https://www.actris.eu/sites/default/files/Documents/ACTRIS\%20PPP/Deliverables/ACTRIS\%20PPP_WP2_D2.3_ACTRIS\%20Data\%20policy.pdf}}\\
& Icare: \url{ftp://ftp.icare.univ-lille1.fr} (GROUND-BASED/OHP\_St-Michel) & \\
& & \\ \hline
\multirow{4}{*}{OPAR \cite{OPAR}}  & \url{https://www.actris.fr/catalogue/} (keyword: sky Imager (La Reunion station)) & \multirow{3}{1.0\linewidth}{\url{https://www.actris.eu/sites/default/files/Documents/ACTRIS\%20PPP/Deliverables/ACTRIS\%20PPP_WP2_D2.3_ACTRIS\%20Data\%20policy.pdf}} \\
& Icare: \url{ftp://ftp.icare.univ-lille1.fr} (GROUND-BASED/OPAR\_La-Reunion) &  \\ %\hline
&  &  \\ 
&OSUR: \url{ftp://tramontane.univ-reunion.fr/data-access/} & CC-BY-NC \\ \hline
ARM-CACTI \cite{ARM_CACTI2019}  & \url{https://adc.arm.gov/discovery/#/results/iopShortName::amf2018cacti} & \multirow{3}{1.0\linewidth}{\url{https://www.arm.gov/guidance/datause/generalguidelines}} \\ \cline{1-2}
ARM-HOU \cite{ARM_HOU2019}  & \url{https://adc.arm.gov/discovery/#/results/s::sky\%20images\%20HOU} & \\ \cline{1-2}
ARM-MARCUS \cite{ARM_MARCUS2019}  & \url{https://adc.arm.gov/discovery/#/results/s::sky\%20images\%20MARCUS} & \\ \hline
El Arenosillo \cite{trigo2005development, El_Arenosillo}  & Email Carmen Córdoba-Jabonero (\href{mailto:cordobajc@inta.es}{cordobajc@inta.es}) & N/A \\ \hline
\multirow{3}{*}{MGCD \cite{liu2020multi}} & Download the agreement: \url{https://github.com/shuangliutjnu/Multimodal-Ground-based-Cloud-Database}; Sign and return to Shuang Liu (\href{mailto:shuangliu.tjnu@gmail.com}{shuangliu.tjnu@gmail.com} or \href{mailto:s.liu@tjnu.edu.cn}{s.liu@tjnu.edu.cn}) & \multirow{3}{*}{Details provided in the agreement} \\ \hline
\multirow{4}{*}{GRSCD \cite{liu2020ground}} & Download the agreement: \url{https://zenodo.org/record/3635497#.Y0UKR-zMKrO}; Sign and return to Shuang Liu (\href{mailto:shuangliu.tjnu@gmail.com}{shuangliu.tjnu@gmail.com} or \href{mailto:s.liu@tjnu.edu.cn}{s.liu@tjnu.edu.cn}) & \multirow{4}{*}{Details provided in the agreement}\\
& GitHub: \url{https://github.com/shuangliutjnu/TJNU-Ground-based-Remote-Sensing-Cloud-Database/tree/V1.0.0} & \\ \hline
WSISEG \cite{xie2020segcloud}  & \url{https://github.com/CV-Application/WSISEG-Database} & N/A \\

WMD \cite{krauz2020assessing}  & \url{https://zenodo.org/record/6203375#.YsczU-yZOrM} & \multirow{1}{1.0\linewidth}{CC BY-NC 4.0 (Attribution-NonCommercial 4.0)} \\ \hline
\multirow{3}{*}{CO-PDD \cite{baray2020cezeaux}}  & 
\url{https://www.actris.fr/catalogue/} & \multirow{3}{1.0\linewidth}{\url{https://www.actris.eu/sites/default/files/Documents/ACTRIS\%20PPP/Deliverables/ACTRIS\%20PPP_WP2_D2.3_ACTRIS\%20Data\%20policy.pdf}} \\
& Icare: \url{ftp://ftp.icare.univ-lille1.fr} (GROUND-BASED/COPDD\_Puy-de-Dome) &   \\ \hline
\multirow{2}{*}{ARM-AWARE \cite{ARM_AWARE2020}}  & \multirow{2}{*}{\url{https://adc.arm.gov/discovery/\#/results/iopShortName::amf2015aware}} & \url{https://www.arm.gov/guidance/datause/generalguidelines} \\ \hline
\multirow{3}{*}{GCD \cite{liu2021ground}}  & Download the agreement: \url{https://github.com/shuangliutjnu/TJNU-Ground-based-Cloud-Dataset}; Sign and return to Shuang Liu (\href{mailto:shuangliu.tjnu@gmail.com}{shuangliu.tjnu@gmail.com} or \href{mailto:s.liu@tjnu.edu.cn}{s.liu@tjnu.edu.cn}) & \multirow{3}{*}{Details provided in the agreement} \\ \hline 
BASS \cite{c2021feasibility}  & Email Pedro C. Valdelomar (\href{mailto:pedro.catalan@uv.es}{pedro.catalan@uv.es}) & N/A \\ \hline
NIMS-KMA \cite{kim2021twenty} & Email Bu-Yo Kim (\href{mailto:kimbuyo@korea.kr}{kimbuyo@korea.kr}) & N/A \\ \hline
Girasol \cite{Girasol2021}  & \url{https://doi.org/10.5061/dryad.zcrjdfn9m} & CC0 1.0 Universal Public Domain Dedication\\ \hline
\multirow{2}{*}{SkyCam \cite{SkyCam2021}} & Email Evangelos Ntavelis (\href{mailto:evangelos.ntavelis@csem.ch}{evangelos.ntavelis@csem.ch}) & \multirow{2}{1\linewidth}{CC BY-NC-SA 4.0 (Attribution-NonCommercial-ShareAlike 4.0)} \\
& GitHub: \url{https://github.com/vglsd/SkyCam} & \\ \hline
SIPM \cite{SIPM2021} &  \url{https://data.mendeley.com/datasets/r83r6g5y6t/1} & CC BY-NC 4.0 (Attribution-NonCommercial 4.0) \\ \hline
SIPPMIF \cite{UOW2021} & \url{https://data.mendeley.com/datasets/cb8t8np9z3/2} & CC BY-NC 3.0 (Attribution-NonCommercial 3.0) \\ \hline
\multirow{4}{*}{Waggle \cite{park2021prediction}} & Image: Email Seongha Park (\href{mailto:seongha.park@anl.gov}{seongha.park@anl.gov}) & \multirow{4}{*}{N/A} \\ 
& Irradiance data: \url{https://www.atmos.anl.gov/ANLMET/index.html} &  \\
& PV data: \url{https://dashboard.ioc.anl.gov/viewer.html?proj=Argonne} &  \\ 
& GitHub: \url{https://github.com/waggle-sensor/solar-irradiance-estimation} & \\ \hline
TAN1802 Voyage \cite{TAN1802_Voyage2021}  & \url{https://zenodo.org/record/4060237} & CC BY-NC 4.0 (Attribution-NonCommercial 4.0) \\ \hline
\multirow{2}{*}{ARM-SAIL/GUC \cite{ARM_GUC2021}} & \multirow{2}{*}{\url{https://adc.arm.gov/discovery/\#/results/s::sky\%20images\%20GUC}} & \url{https://www.arm.gov/guidance/datause/generalguidelines} \\ \hline
\multirow{5}{*}{SKIPP'D \cite{nie2022skippd}} & Benchmark dataset 2017-2019: \url{https://purl.stanford.edu/dj417rh1007} &  \multirow{5}{*}{CC BY-NC 4.0 (Attribution-NonCommercial 4.0)} \\
& Raw dataset 2017: \url{https://purl.stanford.edu/sm043zf7254} & \\ 
& Raw dataset 2018: \url{https://purl.stanford.edu/fb002mq9407} & \\ 
& Raw dataset 2019: \url{https://purl.stanford.edu/jj716hx9049} & \\
& \url{https://github.com/yuhao-nie/Stanford-solar-forecasting-dataset} & \\ \hline
\multirow{3}{*}{TCDD \cite{zhang2021ground}} & Download the agreement: \url{https://github.com/shuangliutjnu/TJNU-Cloud-Detection-Database}; Sign and return to Shuang Liu (\href{mailto:shuangliu.tjnu@gmail.com}{shuangliu.tjnu@gmail.com} or \href{mailto:s.liu@tjnu.edu.cn}{s.liu@tjnu.edu.cn}) & \multirow{3}{*}{Details provided in the agreement} \\ \hline 
PSA Fabel \cite{fabel2022applying} & Email Yann Fabel (\href{mailto:yann.fabel@dlr.de}{yann.fabel@dlr.de}), Bijan Nouri (\href{mailto:bijan.nouri@dlr.de}{bijan.nouri@dlr.de}) & N/A \\ \hline
NAO-CAS XJ \cite{li2022all} & Email Xiaotong Li (\href{mailto:865517454@qq.com}{865517454@qq.com}), Bo Qiu (\href{mailto:qiubo@hebut.edu.cn}{qiubo@hebut.edu.cn}) & N/A \\ \hline
\multirow{4}{*}{TLCDD \cite{zhang2022ground}} & Download the agreement: \url{https://zenodo.org/record/6464743#.Y0URDOzMKrM}; Sign and return to Shuang Liu (\href{mailto:shuangliu.tjnu@gmail.com}{shuangliu.tjnu@gmail.com} or \href{mailto:s.liu@tjnu.edu.cn}{s.liu@tjnu.edu.cn}) & \multirow{4}{*}{Details provided in the agreement} \\ 
& GitHub: \url{https://github.com/shuangliutjnu/TJNU-Large-Scale-Cloud-Detection-Database/tree/v1.0.0} & \\

\multirow{2}{*}{ACS WSI \cite{ye2022self}} & Download the agreement: \url{https://github.com/liangye518/ACS_WSI/tree/v1.0.0}; Sign and return to Liang Ye (\href{mailto:yeliang@wust.edu.cn}{yeliang@wust.edu.cn}) & \multirow{3}{*}{Details provided in the agreement} \\ \hline
ARM-COMBLE \cite{ARM_COMBLE2022} & \url{https://adc.arm.gov/discovery/#/results/s::sky\%20images\%20comble} & \multirow{4}{1.0\linewidth}{\url{https://www.arm.gov/guidance/datause/generalguidelines}} \\ \cline{1-2}
ARM-MOSAIC \cite{ARM_MOSAIC2022} & \url{https://adc.arm.gov/discovery/#/results/s::sky\%20images\%20mosaic} & \\ \cline{1-2}
\multirow{2}{*}{ARM-ACE-ENA \cite{ARM_ACE_ENA2022}}  & \url{https://adc.arm.gov/discovery/\#/results/s::sky\%20images\%20Eastern\%20North\%20Atlantic} & \\ \hline
\multirow{2}{*}{Orion StarShoot
\cite{Orion_Starshoot}}  & \url{http://www.ciao.imaa.cnr.it/index.php?option=com_content&view=article&id=186&Itemid=255} & \multirow{2}{1.0\linewidth}{\url{http://www.ciao.imaa.cnr.it/images/ciao/documenti/Access/ciao_data_policy.pdf}}\\ \hline
LTR \cite{warsaw}  & \url{https://www.igf.fuw.edu.pl/en/meteo-station/lab_tr_pasteura_5/} & N/A \\ \hline
\multirow{3}{*}{LOA \cite{LOA}} & \url{https://www.actris.fr/catalogue/#aeris-metadata-platforms} & \multirow{3}{1.0\linewidth}{\url{https://www.actris.eu/sites/default/files/Documents/ACTRIS\%20PPP/Deliverables/ACTRIS\%20PPP_WP2_D2.3_ACTRIS\%20Data\%20policy.pdf}}\\
& Icare: \url{ftp://ftp.icare.univ-lille1.fr} (GROUND-BASED/LOA\_Lille) & \\ 
& & \\ \hline
\multirow{2}{*}{ARM-OLI \cite{ARM_OLI}} & \multirow{2}{1.0\linewidth}{\url{https://adc.arm.gov/discovery/\#/results/s::sky\%20images\%20OLI}} & \url{https://www.arm.gov/guidance/datause/generalguidelines}\\

\end{longtable}
\end{ThreePartTable}
\end{landscape}

\subsection{Data types in sky image datasets}
\label{subsec:data_inclusion_specs}
In this section, we present detailed descriptions of each data type including sky image, irradiance and PV power generation, segmentation map, cloud category label and meteorological measurements. The data specifications for each dataset, such as image pixel resolution, data temporal resolution and characteristics of segmentation maps and cloud category labels can be found in Table \ref{tab:data_spec_sf_cmp}, \ref{tab:data_spec_cs} and \ref{tab:data_spec_cc} in Appendix \ref{sec:appendixA}.

\begin{figure}[htb!]
	\centering
		\includegraphics[width=.8\textwidth]{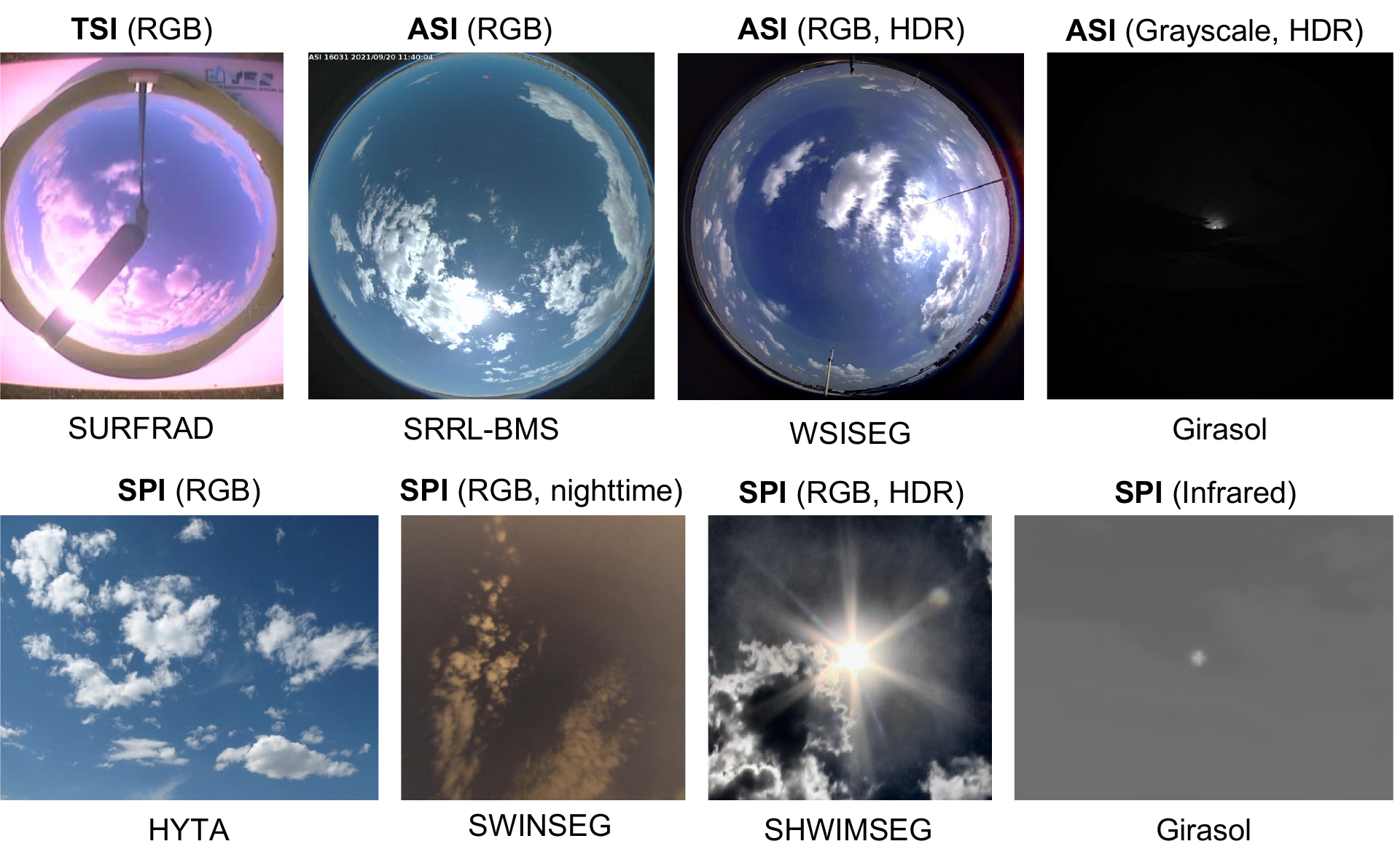}
	  \caption{Sky image samples of different kinds (Note the images presented here do not reflect the actual resolution of the images. The actual resolution of images from different datasets can be found in Table \ref{tab:data_spec_sf_cmp}.)}\label{fig:sky_image_samples}
\end{figure}

\paragraph{\textbf{Sky image}} Three general types of sky images are found in the identified datasets, namely, total sky image (TSI), all sky image (ASI) and sky patch image (SPI). Figure \ref{fig:sky_image_samples} show a few samples of sky images of different kinds from different datasets, with the general sky image type annotated above the samples and other characteristics of the images shown in the bracket, such as RGB, nighttime, infrared and high dynamic range (HDR). 

Each type of sky image has its unique features. TSIs are captured by total sky imagers (e.g., Yankee Environmental Systems's TSI-880 instrument), which use a hemispherical chrome-plated mirror to reflect the sky into a downward-pointing charge-coupled device (CCD) camera located above the mirror, and a sun tracking shadow band is equipped on the mirror to protect the CCD optical sensor from the effects of solar reflection \cite{inman2013solar}. TSIs are generally of low resolution and the presence of a black sun-blocking shadow band and a camera supporting arm prevent getting a total view of the sky dome, going against the fact that the information contained in the circumsolar area is critical to image-based solar forecasting \cite{palettaSPINSimplifyingPolar2021}. To help alleviate this, studies use linear interpolation of surrounding pixels \cite{jiang2020ultra} to fill the missing part of the sky or cloud motion displacement based on two temporally near frames (e.g., 5 min difference in time stamp) to back-calculate the missing pixels in the previous frame by assuming a smooth cloud motion process \cite{wang2020minutely}. Early efforts using TSIs for solar forecasting featured by works done by researchers from University of California at San Diego (UCSD), which propagate clouds using wind vectors extracted from TSIs and build physical deterministic models \cite{chow2011intra,marquez2013intra,quesada-ruizCloudtrackingMethodologyIntrahour2014a} or use features extracted from sky images and train machine learning models \cite{chuHybridIntrahourDNI2013a,Chu2015realtime,Chu2015reforcast,pedroAdaptiveImageFeatures2019}. Some recent works also use TSIs to build end-to-end deep learning models for solar forecasting \cite{Feng2020,Feng2022}. 

In comparison, ASIs tend to have much higher pixel resolution than TSI and do not have a shadow band in the images, so they can provide the whole view of the sky and are used more frequently in recent deep-learning-based short-term solar forecasting models \cite{Sun2018,Zhang2018,Sun2019,Venugopal2019,palettaConvolutionalNeuralNetworks2020,Nie2020,Nie2021,palettaBenchmarkingDeepLearning2021,Paletta2021eclipse}. All sky imagers are typically developed based on a camera equipped with a fish-eye lens and protected by a weatherproof enclosure \cite{chauvin2015cloud}. Industrial sky imagers such as EKO's SRF-02 and ASI-16 are commonly used in sky image collection, e.g., \cite{SRRL1981,SIRTA2005}. Surveillance network cameras, such as Hikvision DS-2CD6365GOE-IVS and MOBOTIX Q25, are also used by researchers for solar forecasting \cite{nie2022skippd, Blum_2021}. Though much cheaper than industrial sky imagers, they work well for providing images for solar forecasting \cite{Sun2018,Sun2019,Venugopal2019,Nie2020,Nie2021}. Some researchers also built their own camera system with customized setups to suit the need for cloud detection or short-term solar forecasting, such efforts including advanced all sky imagers developed by researchers from research institutions, e.g., University of Girona \cite{long2006retrieving}, Universidad de Granada \cite{cazorla2010development}, UCSD \cite{yang2014solar}, PROMES-CNRS \cite{chauvin2015cloud}, and low-cost Raspberry Pi-based sky cameras by University of Texas at San Antonio \cite{Richardson_2017} and University College Dublin \cite{Jain_2022}.

SPIs are either taken by cameras equipped with a normal lens (in contrast to fish-eye lenses used for capturing ASI) or obtained from patches of unwrapped ASIs. They only cover a portion of the sky (i.e. a limited range of zenith and azimuth angles) unlike ASIs and TSIs, which could capture the panoramic view of the sky. Additionally, SPIs do not have distorted regions, which are present in TSIs and ASIs because the lens' field of view is wider than the size of the image sensor. Thus, the use of SPIs for solar forecasting tasks is limited, as it probably needs information from different parts of the sky for models to learn to anticipate the motion of clouds, while they are more frequently used in tasks such as cloud segmentation or classification, which do not require the model's attention on the cloud motion. 

For most studies, these three types of sky images are captured by visible light sensors, while there are studies using images taken by infrared sensors, e.g., \cite{terren2021segmentation,terren2021explicit,terren2021comparative,ajith2021deep}. The advantages of using infrared sensor over visible sensor include avoiding the saturation problem of the circumsolar region as well as allowing the derivation of valuable cloud properties such as the temperature and altitude \cite{terren2021comparative}. High dynamic range (HDR) is a technique used to balance the light of images by blending several images taken with different exposure time into one with high contrast. In sky imaging, it is useful in lots of scenes such as very bright direct sunlight and extreme shade of clouds. Although most sky images are captured during the day time, some are taken at night for application purposes such as weather reporting and prediction, aviation, and satellite communication \cite{dev2017nighttime}.

\paragraph{\textbf{Irradiance and PV power generation}} Irradiance and PV power generation are highly correlated and can both be used as target variables for solar forecasting. Generally, irradiance data is more common than PV data in terms of openness, due to privacy restriction or energy security in accessing residential or utility PV data. The same trend has been found in our identified open-source dataset, 45 datasets provide irradiance data while only 5 datasets provide PV generation data. Moreover, these PV data are from small-scale rooftop PV system or outdoor testing facility for research purposes. On the other hand, compared with irradiance forecasting, which can be adapted to different locations, PV forecasting model is hard to be adapted to other locations given the different specifications of the PV systems. Different types of solar irradiance are found in the datasets. Some commonly measured irradiance components include the global horizontal irradiance (GHI), the direct normal irradiance (DNI) and the diffuse horizontal irradiance (DHI). DNI, or beam irradiance, is the incident light directly from the Sun measured at the surface of the Earth normal to the optical path. DNI is critical for concentrated solar technologies and can be greatly affected by cloud cover and aerosol content. DHI is the irradiance received by a horizontal surface at the Earth's surface which has been scattered or diffused by the atmosphere. GHI is the total irradiance falling on a surface horizontal to the surface of the earth, which can be calculated based on DNI and DHI using the following equation, Where $\theta_z$ is the solar zenith angle (Equation~\ref{equation:ghi}).
\begin{equation}
    \mathrm{GHI} = \mathrm{DNI}\times \cos (\theta_z)+\mathrm{DHI}
    \label{equation:ghi}
\end{equation}
While GHI and DHI can be both measured using pyranometers placed on a horizontal plane, for measuring DHI, a black ball or disc needs to be installed together with a sun tracker to remove the direct component from the sun light. The DNI can be measured via a pyrheliometer. To maximize the PV power production, solar panels are usually tilted at an angle to maximize the irradiance reaching the panel, which is referred to as global irradiance in the plane of array $\mathrm{G_{POA}}$ or global tilted irradiance (GTI). $\mathrm{G_{POA}}$ or GTI is crucial for modeling the performance of PV system, and consists of three components, namely, a beam (direct) component ($\mathrm{G_{B,POA}}$), a sky-diffuse component ($\mathrm{G_{D,POA}}$) and a ground-reflected component ($\mathrm{G_{R,POA}}$). $\mathrm{G_{POA}}$ can either be measured using a pyranometer placing on the tilted panel surface or can be calculated based on transposition of GHI, DNI and DHI to the three components of $\mathrm{G_{POA}}$ \cite{lave2015evaluation}.

\begin{figure}[htb!]
	\centering
		\includegraphics[width=.8\textwidth]{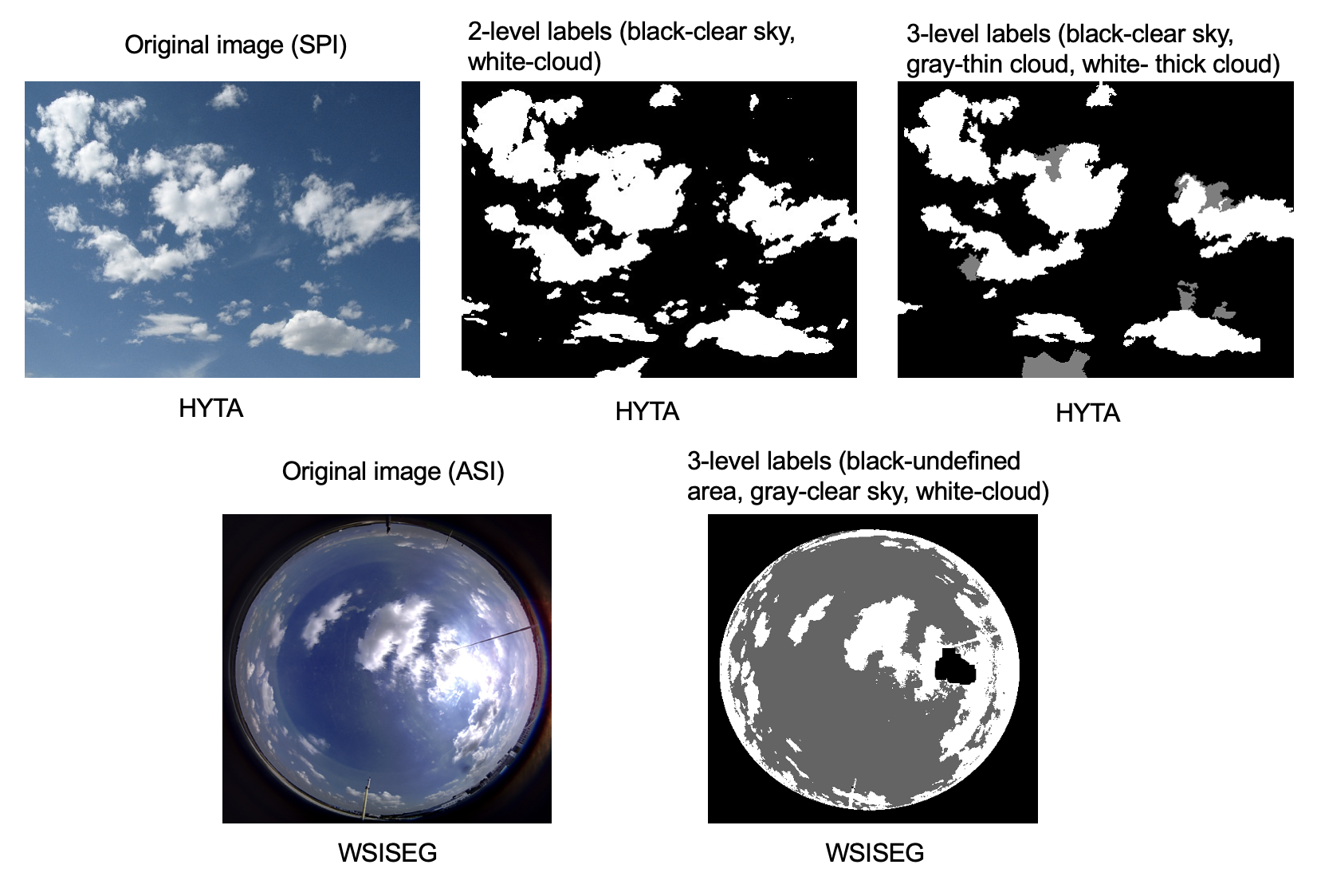}
	  \caption{Segmentation map samples of different kinds (Note the samples presented here do not reflect the actual resolution of the images. The actual resolution of images from different datasets can be found in Table \ref{tab:data_spec_sf_cmp}.)}
	  \label{fig:sm_samples}
\end{figure}

\paragraph{\textbf{Segmentation map}} Two types of segmentation maps are identified across the datasets. The first type corresponds to segmentation maps generated by algorithms like red-blue-ratio thresholding. This type of segmentation map usually comes from datasets released by national labs or government scientific organizations, i.e. type 'O' datasets. The second type is segmentation map manually labeled by human experts, that is generally for datasets that specifically focusing on cloud segmentation task. Besides, datasets provide different levels of segmentation for the cloud images. Algorithm generated segmentations almost only provide the binary segmentation of sky and clouds. Human labelled segmentation contains more diversity. Although most of them provide binary segmentation maps, some datasets provide more than 2 levels of segmentation by differentiating the cloud types at pixel level, so these datasets can also be used in cloud type classification tasks. These datasets include HYTA (Ternary) \cite{dev2015multi}, which provides a 3-level segmentation of sky, thin and thick clouds; FGCDR \cite{ye2019supervised}, which provides a 8-level segmentation of 8 different cloud categories at pixel level; WMD \cite{krauz2020assessing}, which provides a 4-level segmentation of high-level clouds, low-level cumulus type clouds, rain clouds and clear sky and PSA Fabel \cite{fabel2022applying}, which provides a 3-level segmentation of low-, middle- and high-layer clouds. Figure \ref{fig:sm_samples} shows segmentation map samples of 2-level and 3-level labels from HYTA and WSISEG.

%\textcolor{red}{sample images for segmentation map?}

\begin{figure}[htb!]
	\centering
		\includegraphics[width=.8\textwidth]{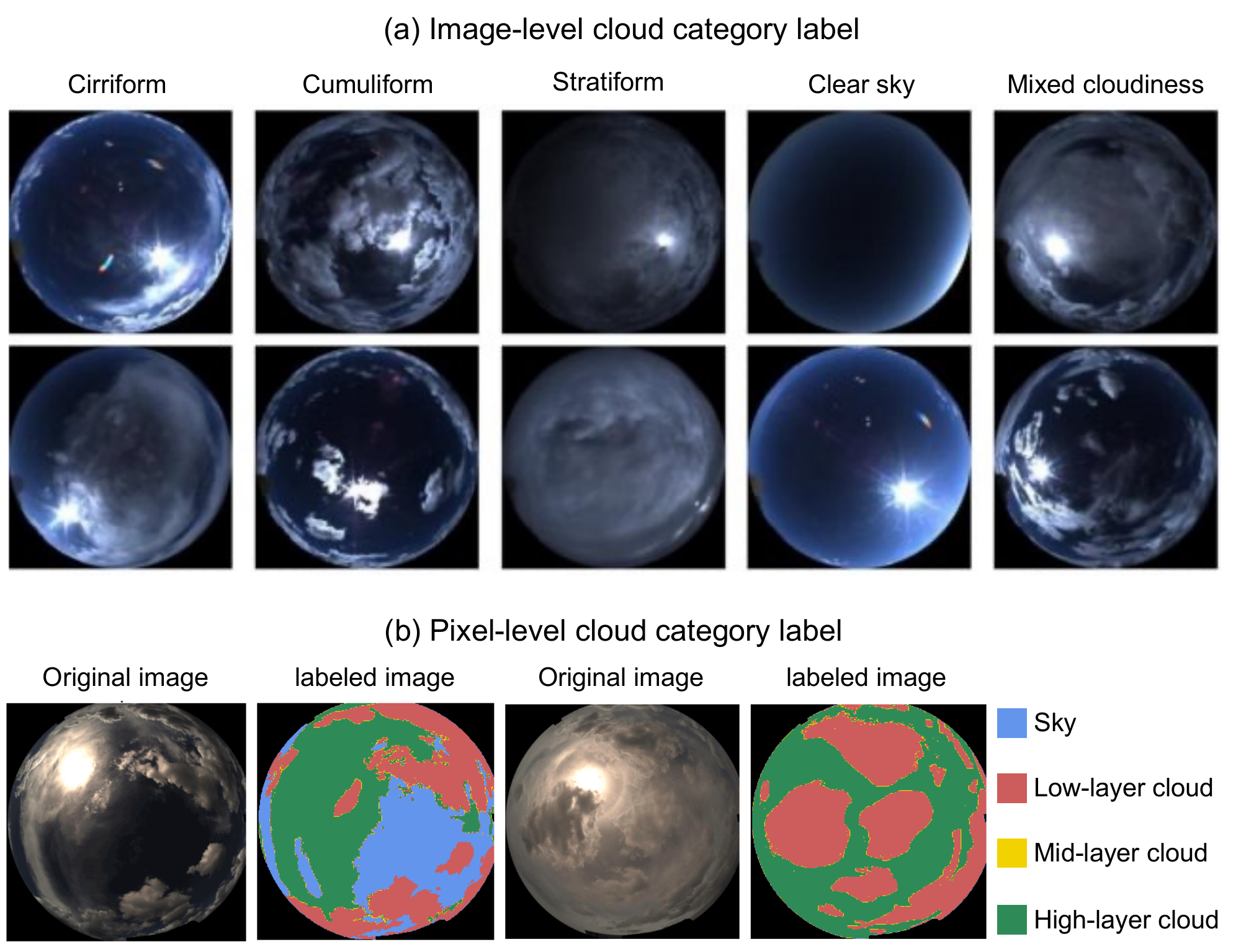}
	  \caption{Cloud category label samples of different kinds (Note: (1) (a) shows the samples of image-level cloud category label of five sky condition classes adapted from \cite{li2016pixels}, and (b) shows the samples of pixel-level cloud category label adapted from \cite{fabel2022applying}; (2)the images presented here do not reflect the actual resolution of the images. The actual resolution of images from different datasets can be found in Table \ref{tab:data_spec_sf_cmp}.)}
	  \label{fig:cloud_label_samples}
\end{figure}

\paragraph{\textbf{Cloud category label}} Another type of label data is cloud category labels, which can be further divided into image-level labels and pixel-level labels. As the name suggests, image-level labels classify images into certain cloud types, while pixel-level labels distinguish for each pixel in the image, what cloud type this pixel belongs to. The pixel-level cloud category label can also be used for the cloud segmentation tasks as mentioned before. There is no consistent agreement on the cloud type classification criteria, some are based on the main cloud genera recommended by the World Meteorological Organization (WMO) \cite{zhuo2014cloud,CCSN2018,ye2019supervised} (e.g., Cirrus, Cumulus, Stratus, Nimbus), some are based on the visual characteristics of clouds \cite{SWIMCAT2015,li2016pixels,luo2018ground} (e.g., patterned clouds, thick dark clouds, thin white clouds, veil clouds) and some are based on the height of clouds \cite{fabel2022applying} (e.g., high-level clouds, medium-level clouds, low-level clouds). Figure \ref{fig:cloud_label_samples} shows samples of cloud category label data both at image-level and pixel-level.

%\textcolor{red}{sample images for cloud category label?}

\paragraph{\textbf{Meteorological measurements}} Meteorological data such as temperature, wind speed and direction, pressure, humidity, etc., have varying impacts on the solar irradiance reaching the ground and the amount of power able to be generated from that irradiance. Making use of these information can potentially contribute to the development of robust solar forecasting models. However, the challenge is to make the deep learning systems pay due attention to both imagery and sensor measurement data. Imagery data is often high dimensional, i.e. an array with hundreds of thousands of entries, which might catch more attention from the deep systems than sensor measurements that might just be a vector of 10s of entries. This is actually an active field of research called data fusion, applications like self-driving \cite{xu2018pointfusion} and robotics \cite{zhou2017incorporating} have explored this thoroughly, while very limited studies has been found investigating this problem for sky-image-base solar forecasting. In 2019, \citet{Venugopal2019} first systematically compared 28 methods of fusion (MoF) for integrating hybrid input of sky image sequences and PV output history to forecast 15-min-ahead PV power output by using CNN as backbone. Although no other sensor data are investigated beside PV power generation, it provides a general guideline for fusing sensor data with imagery data for solar forecasting. Similarly, \citet{ajithDeepLearningBased2021} developed a multi-modal fusion network based on CNN and LSTM for studying solar irradiance forecasts using IR images and past solar irradiance data with forecasting horizon ranging from 15 to 150s. More recently, \citet{terren-serranoDeepLearningIntraHour2022} explored fusing information from IR sky images with various sensors including pyranometer, solar tracker and weather station using a multi-task deep learning architecture based on RNNs.

\subsection{Spatial and temporal coverage of datasets}
The geographic locations, collection period as well as size of each dataset are shown in Table \ref{tab:dataset_spatial_temporal_cover}. If the information about the size of the dataset is not provided, we estimate it based on the data collection period and the temporal resolution of the data if available. The size of the datasets varies from 733 KB to 16 TB, depending on the number of sites and time duration for data collection, the temporal resolution of data logging as well as the type of data included. For example, SkyCam dataset logs data every 10 seconds from 3 different location for 365 days, and SKIPP'D dataset provides sky video footage. Most of the type O datasets (see Table \ref{tab:dataset_basic_info}) have multi-year efforts in collecting data, while for type R datasets (see Table \ref{tab:dataset_basic_info}), the temporal coverage of the data varied from one to another. Ideally, we would want the datasets collection spans multiple years to catch the annual variability in local weather conditions.

\begin{landscape}
\begin{ThreePartTable}
\begin{TableNotes}[flushleft]
\item Note: for size estimation column (Size est.), Img. stands for only the size of image data, and Tot. represents the total dataset size including image and all other types of data such as sensor measurements, segmentation map and cloud category label.
\end{TableNotes}

\begin{longtable}{>{\raggedright}p{0.15\linewidth}>{\raggedright}p{0.10\linewidth}>{\raggedright}p{0.27\linewidth}>{\raggedright}p{0.10\linewidth}>{\raggedright}p{0.2\linewidth}>{\raggedright\arraybackslash}p{0.08\linewidth}}
\caption{Spatial and temporal coverage of the dataset.} \label{tab:dataset_spatial_temporal_cover} \\

\toprule
\multirow{2}{*}{Dataset} & Spatial coverage (\# of sites) & \multirow{2}{*}{Locations} & Temporal coverage (yrs) & \multirow{2}{*}{Collection period} & \multirow{2}{*}{Size est. (GB)} \\ % Table header row
\midrule 
\endfirsthead

\caption {Spatial and temporal coverage of the dataset (continued).}\\
\toprule
\multirow{2}{*}{Dataset} & Spatial coverage (\# of sites) & \multirow{2}{*}{Locations} & Temporal coverage (yrs) & \multirow{2}{*}{Collection period} & \multirow{2}{*}{Size est. (GB)} \\ % Table header row
\midrule 
\endhead

\hline
\multicolumn{6}{r}{{Continued on next page}} \\ \hline
\endfoot

\bottomrule
\insertTableNotes
\endlastfoot

\multirow{2}{*}{SRRL-BMS\cite{SRRL1981}} & \multirow{2}{*}{1} & \multirow{2}{*}{Golden, Colorado, US} & TSI: 18.0 & 2004.07.14-present & 4.7 (Img.)\\ 
 & & & ASI: 4.8 & 2017.09.26-present & 9.2 (Img.) \\ \noalign{\vspace{1mm}} \hline \noalign{\vspace{1mm}}
\multirow{7}{*}{SURFRAD \cite{SURFRAD2000}} & \multirow{7}{*}{7} & Bondville, Illinois, US & 4.7 & 2006.09.01-2011.05.25 &  0.4 (Tot.) \\ 
& & Boulder, Colorado, US & 5.3 & 2006.09.01-2011.11.30 & 0.9 (Tot.) \\ 
& & Dessert Rock, Nevada, US & 8.4 & 2006.09.01-2015.01.25 & 0.9 (Tot.) \\ 
& & Fort Peck, Montana, US & 8.4 & 2006.09.01-2015.01.25 & 1.0 (Tot.) \\ 
& & Goodwin Creek, Mississippi, US & 4.7 & 2006.09.01-2011.04.28 & 0.5 (Tot.) \\
& & Sioux Falls, South Dakota, US & 2.7 & 2008.10.11-2011.06.08 & 0.3 (Tot.)\\ 
& & PSU, Pennsylvania, US & 8.4 & 2006.09.01-2015.01.19 & 0.9 (Tot.)\\ \noalign{\vspace{1mm}} \hline \noalign{\vspace{1mm}}
\multirow{2}{*}{SIRTA \cite{SIRTA2005}} & \multirow{2}{*}{1} & \multirow{2}{*}{Palaiseau, France} & TSI: 6.7 & 2008.10.23-2015.06.24 & N/A \\ 
 & & & ASI: 7.6 & 2014.12-present & N/A \\ \noalign{\vspace{1mm}} \hline \noalign{\vspace{1mm}}
ARM-MASRAD \cite{ARM_MASRAD2005} & 1 & Point Reyes, California, US & 0.6 & 2005.02.01-2005.09.15 & 6.9 (Img.)\\ \noalign{\vspace{1mm}} \hline \noalign{\vspace{1mm}}
ARM-RADAGAST \cite{ARM_RADAGAST2008} & 1 & Niamey, Niger & 11.2 & 2005.11.24-2017.01.07 & 14.5 (Img.)\\ \noalign{\vspace{1mm}} \hline \noalign{\vspace{1mm}}
ARM-STORMVEX \cite{ARM_STORMVEX2010} & 1 & Steamboat Springs, Colorado, US & 0.6 & 2010.09.23-2011.04.22 & 11.1 (Img.)\\ \noalign{\vspace{1mm}} \hline \noalign{\vspace{1mm}}
\multirow{2}{*}{HYTA (Binary)\cite{li2011hybrid} } & \multirow{2}{*}{2} & Beijing, China & N/A & N/A & \multirow{2}{*}{0.001 (Tot.)}\\
& & Conghua, Guangdong, China & N/A & N/A & \\ \noalign{\vspace{1mm}} \hline \noalign{\vspace{1mm}}
ARM-AMIE-GAN \cite{ARM_AMIE_GAN2011} & 1 & Gan Island, Maldives & 0.4 & 2011.09.25-2012.02.09  & 10.1 (Img.) \\ \noalign{\vspace{1mm}} \hline \noalign{\vspace{1mm}}
ARM-COPS \cite{ARM_COPS2011} & 1 & Black Forest, Germany & 0.8 & 2007.03.14-2008.01.01 &  13.2 (Img.)\\ \noalign{\vspace{1mm}} \hline \noalign{\vspace{1mm}}
ARM-HFE \cite{ARM_EAST_AIRC2011} & 1 & Shouxian, Anhui, China & 0.6 & 2008.05.08-2008.12.28 & 7.1 (Img.) \\ \noalign{\vspace{1mm}} \hline \noalign{\vspace{1mm}}
ARM-GVAX \cite{ARM_GVAX2013} &1 & Nainital, Uttarkhand, India & 0.7 & 2011.07.16-2012.04.01 & 20.4 (Img.) \\ \noalign{\vspace{1mm}} \hline \noalign{\vspace{1mm}}
SWIMCAT \cite{SWIMCAT2015} & 1 & Nanyang Technological University, Singapore & 1.3 & 2013.01-2014.05  & 0.01 (Tot.)\\ \noalign{\vspace{1mm}} \hline \noalign{\vspace{1mm}}
\multirow{2}{*}{HYTA (Ternary) \cite{dev2015multi}} & \multirow{2}{*}{2} & Beijing, China & N/A & N/A  & \multirow{2}{*}{0.0007 (Tot.)}\\ 
& & Conghua, Guangdong, China & N/A & N/A & \\ \noalign{\vspace{1mm}} \hline \noalign{\vspace{1mm}}
ARM-CAP-MBL \cite{ARM_CAP_MBL2015} & 1 & Graciosa Island, Azores, Portugal & 1.7 & 2009.04.14-2011.01.05  & 23.6 (Img.)\\ \noalign{\vspace{1mm}} \hline \noalign{\vspace{1mm}}
ARM-TCAP \cite{ARM_TCAP2015} & 1 & Highland Center, Massachusetts, US & 1.0 & 2012.06.29-2013.07.08 & 29.9 (Img.) \\ \noalign{\vspace{1mm}} \hline \noalign{\vspace{1mm}}
TCIS \cite{li2016pixels} & 1 & Tibet, China & 2.0 & 2012.08-2014.07 & N/A\\ \noalign{\vspace{1mm}} \hline \noalign{\vspace{1mm}}
NCU \cite{cheng2017cloud} & 1 & National Central University, Taiwan & 0.5 & 2014.01-2014.06 & N/A \\ \noalign{\vspace{1mm}} %\hline \noalign{\vspace{1mm}}
ARM-BAECC \cite{BAECCA2016} & 1 & Hyytiälä, Finland & 0.6 & 2014.02.01-09.13  & 22.6 (Img.)\\ \noalign{\vspace{1mm}} \hline \noalign{\vspace{1mm}}
\multirow{2}{*}{ARM-ACAPEX \cite{ARM_ACAPEX2016}} & \multirow{2}{*}{2} & Honolulu, Hawaii, US & 0.1 & 2015.01.09-02.12 & \multirow{2}{*}{2.6 (Img.)} \\ 
 & & San Diego, California, US & 0.1 & 2015.01.09-02.12 & \\ \noalign{\vspace{1mm}} \hline \noalign{\vspace{1mm}}
\multirow{2}{*}{ARM-MAGIC \cite{ARM_MAGIC2016}} & \multirow{2}{*}{2} & Los Angeles, California, US & \multirow{2}{*}{1.0} & \multirow{2}{*}{2012.10.01-2013.09.26} & \multirow{2}{*}{19.3 (Img.)} \\
 & & Honolulu, Hawaii, US & & &  \\ \noalign{\vspace{1mm}} \hline \noalign{\vspace{1mm}}
ARM-GoAmazon \cite{ARM_GoAmazon2016}  & 1 & Manacapuru, Amazonas, Brazil & 1.9 & 2014.01.01-2015.12.01 & 54.2 (Img.)\\ \noalign{\vspace{1mm}} \hline \noalign{\vspace{1mm}}
ARM-NSA \cite{ARM_NSA2016} & 1 & Barrow, Alaska, US & 16.3 & 2006.04.25-2022.08.20 & 350.2 (Img.) \\ \noalign{\vspace{1mm}} \hline \noalign{\vspace{1mm}}
\multirow{3}{*}{ARM-SGP \cite{ARM_SGP2016}} & \multirow{3}{*}{3} & \multirow{3}{*}{Lamont, Oklahoma, US} & Site 1: 22.1 &  2000.07.02-2022.08.14 & 443.7 (Img.) \\
& &  & Site 2: 5.7 & 2016.12.15-2022.08.14 & 159.6 (Img.) \\
& &  & Site 3: 0.7 & 2016.04.20-2016.12.15 & 16.0 (Img.) \\ \noalign{\vspace{1mm}} \hline \noalign{\vspace{1mm}}
\multirow{3}{*}{ARM-TWP \cite{ARM_TWP2016}} & \multirow{3}{*}{3} & Manus, Papua New Guinea & 10.5 & 2003.11.30-2014.06.10 & 143.9 (Img.) \\ 
&  & Nauru Island, Nauru & 10.8 &2002.11.12-2013.09.09 & 177.1 (Img.)\\
&  & Darwin, Australia & 12.5 & 2002.07.16-2015.01.04 & 222.1 (Img.)\\ \noalign{\vspace{1mm}} \hline \noalign{\vspace{1mm}}
SWIMSEG \cite{SWIMSEG2017} & 1 & Nanyang Technological University, Singapore & 1.7 & 2013.10-2015.07 & 0.2 (Tot.)\\ \noalign{\vspace{1mm}} \hline \noalign{\vspace{1mm}}
SWINSEG \cite{dev2017nighttime} & 1 & Nanyang Technological University, Singapore & 1.0 & 2016.01-2016.12  & 0.003 (Tot.) \\ \noalign{\vspace{1mm}} \hline \noalign{\vspace{1mm}}
SHWIMSEG \cite{SHWIMSEG2018} & 1 & Nanyang Technological University, Singapore & N/A & N/A & 0.2 (Tot.) \\ \noalign{\vspace{1mm}} \hline \noalign{\vspace{1mm}}
Zenithal \cite{luo2018ground} & 1 & Nanjing, China & N/A & N/A & N/A \\ \noalign{\vspace{1mm}} \hline \noalign{\vspace{1mm}}
CCSN \cite{CCSN2018} & N/A & N/A & N/A & N/A & 0.09 (Tot.) \\ \noalign{\vspace{1mm}} \hline \noalign{\vspace{1mm}}
ARM-LASIC \cite{ARM_LASIC2018} & 1 & Ascension Island, South Atlantic Ocean & 1.5 & 2016.05.02-2017.10.31 & 36.9 (Img.)\\ \noalign{\vspace{1mm}} \hline \noalign{\vspace{1mm}}
NAO-CAS \cite{shi2019diurnal} & 1 & China & N/A & N/A & N/A\\ \noalign{\vspace{1mm}} \hline \noalign{\vspace{1mm}}
\multirow{2}{*}{FGCDR \cite{ye2019supervised}} & \multirow{2}{*}{2} & Hangzhou, China & N/A & N/A & N/A\\
& & Lijiang, China & N/A & N/A & N/A \\ \noalign{\vspace{1mm}} \hline \noalign{\vspace{1mm}}
LES dataset \cite{caldas2019very} & 1 & Salto, Uruguay & 0.1 (22 days) & 2016.05-2016.11 & N/A\\ \noalign{\vspace{1mm}} \hline \noalign{\vspace{1mm}}
SWINySEG \cite{dev2019cloudsegnet} & 1 & Nanyang Technological University, Singapore & 2.7 & 2013.10-2015.07 \& 2016.01-12 & 0.06 (Tot.) \\ \noalign{\vspace{1mm}} \hline \noalign{\vspace{1mm}}
UCSD-Folsom \cite{UCSD2019} & 1 & Folsom, California, US & 3.0 & 2014.01-2016.12 & 49.8 (Tot.) \\ \noalign{\vspace{1mm}} \hline \noalign{\vspace{1mm}}
UoH \cite{dandiniHaloRatioGroundbased2019} & 6 & Bayfordbury, Hertfordshire, UK & 10.0 & 2012.07.25-present & N/A \\ 
\multirow{5}{*}{UoH \cite{dandiniHaloRatioGroundbased2019}} & \multirow{5}{*}{6} & Hemel Hempstead, Hertfordshire, UK & 12.0 & 2010.07.09-present &  N/A\\
 &  & Exmoor, UK & 1.6 & 2011.10.27-2013.05.15 & N/A\\
& & Niton, Isle of Wight & 6.3 & 2010.08.10-2016.11.07 & N/A\\
& & Cromer, Norfolk, UK & 0.4 & 2010.09.02-2011.01.15 & N/A\\
&  & Guernsey, UK & 3.7 & 2011.08.09-2015.04.30 & N/A\\ \noalign{\vspace{1mm}} \hline \noalign{\vspace{1mm}}
\multirow{2}{*}{P2OA-RAPACE \cite{lothon2019elifan}} & \multirow{2}{*}{2} & Lannemezan, Midi-Pyrénées, France & 16.4 & TSI: 2006.02-present  & N/A  \\ 
& & Pic du Midi de Bigorre, France & 2.5 & ASI: 2017.07-2019.12 & N/A \\ \noalign{\vspace{1mm}} \hline \noalign{\vspace{1mm}}
OHP \cite{OHP} & 1 & Saint-Michel-l'Observatoire, France & 1.5 & 2015.01.01-2016.06.15 & N/A \\ \noalign{\vspace{1mm}} \hline \noalign{\vspace{1mm}}
OPAR \cite{OPAR} & 1 & MAIDO, France & 1.5 & 2015.01.01-2016.06.15 & N/A \\ \noalign{\vspace{1mm}} \hline \noalign{\vspace{1mm}}
ARM-CACTI \cite{ARM_CACTI2019} & 1 & Cordoba, Argentina & 0.6 &  2018.09.23-2019.05.01 & 20.0 (Img.) \\ \noalign{\vspace{1mm}} \hline \noalign{\vspace{1mm}}
ARM-HOU \cite{ARM_HOU2019} & 1 & Houston, TX, US & 1.0 & 2021.08.02-2022.08.20 & 32.1 (Img.) \\ \noalign{\vspace{1mm}} \hline \noalign{\vspace{1mm}}
ARM-MARCUS \cite{ARM_MARCUS2019} & 1 & Hobart, Australia  & 0.4 & 2017.10.29-2018.03.24 & 12.8 (Img.) \\ \noalign{\vspace{1mm}} \hline \noalign{\vspace{1mm}}
El Arenosillo \cite{trigo2005development, El_Arenosillo} & 1 & Huelva, Spain & N/A & N/A & N/A \\ \noalign{\vspace{1mm}} \hline \noalign{\vspace{1mm}}
MGCD \cite{liu2020multi} & 1 & Tianjin, China & 1.8 & 2017.03-2018.12  & N/A\\ \noalign{\vspace{1mm}} \hline \noalign{\vspace{1mm}}
GRSCD \cite{liu2020ground} & 1 & Tianjin, China & 1.0 & 2017-2018  & N/A\\ \noalign{\vspace{1mm}} \hline \noalign{\vspace{1mm}}
WSISEG \cite{xie2020segcloud} & 1 & China & N/A & N/A  & 0.1 (Tot.) \\\noalign{\vspace{1mm}} \hline \noalign{\vspace{1mm}}
\multirow{2}{*}{WMD \cite{krauz2020assessing}} & \multirow{2}{*}{1} & Jarošov nad Nežárkou, South Bohemia, Czechia & \multirow{2}{*}{0.2} & \multirow{2}{*}{2016.2.22-2016.5.18} & \multirow{2}{*}{29.3 (Tot.)} \\\noalign{\vspace{1mm}} \hline \noalign{\vspace{1mm}}
CO-PDD \cite{baray2020cezeaux} & 1 & Cézeaux, Opme, Puy de Dôme, France & 2.3 & 2015.12-2018.03 & N/A \\ \noalign{\vspace{1mm}} \hline \noalign{\vspace{1mm}}
\multirow{2}{*}{ARM-AWARE \cite{ARM_AWARE2020}} & \multirow{2}{*}{2} & McMurdo Station Ross Ice Shelf, Antarctica & 1.1 & 2015.12.01-2017.01.01 & 28.7 (Img.)\\ 
& & West Antarctic Ice Sheet, Antarctica & 0.1 & 2015.12.06-2016.01.17  & 5.2 (Img.) \\ \noalign{\vspace{1mm}} \hline \noalign{\vspace{1mm}}
\multirow{2}{*}{GCD \cite{liu2021ground}} & \multirow{2}{*}{9} & Tianjin, Anhui, Sichuan, Gansu, Shandong, Hebei, Liaoning, Jiangsu, Hainan in China & \multirow{2}{*}{1.0} & \multirow{2}{*}{N/A} & \multirow{2}{*}{N/A} \\ \noalign{\vspace{1mm}} \hline \noalign{\vspace{1mm}}
BASS \cite{c2021feasibility} & 1 & Valencia, Spain & 0.4 & 2020.02-2021.06 & N/A \\ \noalign{\vspace{1mm}} \hline \noalign{\vspace{1mm}}
NIMS-KMA \cite{kim2021twenty} & 1 & Daejeon, Korea & 1.0 & 2019.01-12 & N/A\\ \noalign{\vspace{1mm}} \hline \noalign{\vspace{1mm}}
Girasol \cite{Girasol2021} & 1 & University of New Mexico, New Mexico, US & 0.7 (244 days) &  2017-2019  & 110.0 (Tot.) \\ \noalign{\vspace{1mm}}

\multirow{3}{*}{SkyCam \cite{SkyCam2021}} & \multirow{3}{*}{3} & Neuchâtel, Switzerland & 1.0 & 2018.01.01-12.31 & \multirow{3}{*}{16000 (Tot.) } \\
& & Bern, Switzerland & 1.0 & 2018.01.01-12.31  & \\
& & Alpnach, Switzerland & 1.0 & 2018.01.01-12.31 & \\ \noalign{\vspace{1mm}} \hline \noalign{\vspace{1mm}}
SIPM \cite{SIPM2021} & 1 & Rio de Janeiro, Brazil & 0.1 (26 days) & 2019.02.25-03.23 & 0.3 (Tot.)\\ \noalign{\vspace{1mm}} \hline \noalign{\vspace{1mm}}
SIPPMIF \cite{UOW2021} & 2 (1.9 km dist.) & UoW, New South Wales, Australia & 0.0 (2 days) & 2019.09.10-09.12 & 7.2 (Tot.)\\ \noalign{\vspace{1mm}} \hline \noalign{\vspace{1mm}}
Waggle \cite{park2021prediction} & 1 & Lemont, Illinois, US & 0.0 (14 days) & 2020.06 & N/A\\ \noalign{\vspace{1mm}} \hline \noalign{\vspace{1mm}}
\multirow{2}{*}{TAN1802 Voyage \cite{TAN1802_Voyage2021}} & \multirow{2}{*}{1} & Wellington, New Zealand and Ross Sea, Antarctica & \multirow{2}{*}{0.1} & \multirow{2}{*}{2018.02.08-03.21}  & \multirow{2}{*}{2.7 (Img.)} \\ \noalign{\vspace{1mm}} \hline \noalign{\vspace{1mm}}
ARM-SAIL/GUC \cite{ARM_GUC2021} & 1 & Gunnison, CO, US & 1.1 &  2021.07.07-2022.08.20  & 55.0 (Img.)\\ \noalign{\vspace{1mm}} \hline \noalign{\vspace{1mm}}
SKIPP'D \cite{nie2022skippd} & 1 & Stanford University, California, US & 2.7 & 2017.03-2019.10 & 1700 (Tot.) \\ \noalign{\vspace{1mm}} \hline \noalign{\vspace{1mm}}
\multirow{2}{*}{TCDD \cite{zhang2021ground}} & \multirow{2}{*}{9} & Tianjin, Anhui, Sichuan, Gansu, Shandong, Hebei, Liaoning, Jiangsu, Hainan in China & \multirow{2}{*}{2.0} & \multirow{2}{*}{2019-2021} &  \multirow{2}{*}{N/A} \\  \noalign{\vspace{1mm}} \hline \noalign{\vspace{1mm}}
PSA Fabel \cite{fabel2022applying} & 1 & Plataforma Solar de Almeria, Spain & 1.0 & 2017.01-2017.12 & N/A\\ \noalign{\vspace{1mm}} \hline \noalign{\vspace{1mm}}
NAO-CAS XJ \cite{li2022all} & 1 & Xinjiang, China & 1.9 & 2019.01-2020.11 & N/A\\\noalign{\vspace{1mm}} \hline \noalign{\vspace{1mm}}
\multirow{2}{*}{TLCDD \cite{zhang2022ground}} & \multirow{2}{*}{9} & Tianjin, Anhui, Sichuan, Gansu, Shandong, Hebei, Liaoning, Jiangsu, Hainan in China & \multirow{2}{*}{2.0} & \multirow{2}{*}{2019-2021} & \multirow{2}{*}{N/A} \\ \noalign{\vspace{1mm}} \hline \noalign{\vspace{1mm}}
ACS WSI \cite{ye2022self} & N/A & China & 1.0 & 2013.01-2013.12 & N/A \\ \noalign{\vspace{1mm}} \hline \noalign{\vspace{1mm}}
Orion StarShoot \cite{Orion_Starshoot} & 1 & Tito Scalo, Italy & N/A & N/A & N/A\\ \noalign{\vspace{1mm}} \hline \noalign{\vspace{1mm}}
LTR \cite{warsaw} & 1 & Warsaw, Poland & 2.0 & 2019-2021 & N/A\\ \noalign{\vspace{1mm}} \hline \noalign{\vspace{1mm}}
LOA \cite{LOA} & 1 & Villeneuve d'Ascq, France & 9.1 & 2009.09.16-2018.11.05 & N/A \\ \noalign{\vspace{1mm}} \hline \noalign{\vspace{1mm}}
ARM-COMBLE \cite{ARM_COMBLE2022} & 2 & Andenes and Bear Island in Norway & 0.3 & 2020.01.28-06.01  & 18.0 (Img.) \\  \noalign{\vspace{1mm}} \hline \noalign{\vspace{1mm}}
ARM-MOSAIC \cite{ARM_MOSAIC2022} & 1 & Central Arctic, Arctic Circle & 0.5 & 2020.03.23-09.20  & 16.0 (Img.) \\ \noalign{\vspace{1mm}} \hline \noalign{\vspace{1mm}}
ARM-ACE-ENA \cite{ARM_ACE_ENA2022} & 1 & Graciosa Island, Azores, Portugal & 8.8 & 2013.10.01-2022.08.01  & 153.2 (Img.) \\ \noalign{\vspace{1mm}} \hline \noalign{\vspace{1mm}}
ARM-OLI \cite{ARM_OLI} & 1 & Oliktok Point, AK, US & 7.7 & 2013.10.01-2021.06.15 & 107.8 (Img.) \\
\end{longtable}
\end{ThreePartTable}
\end{landscape}

\begin{figure}[!htbp]
	\centering
		\includegraphics[width=1.0\textwidth]{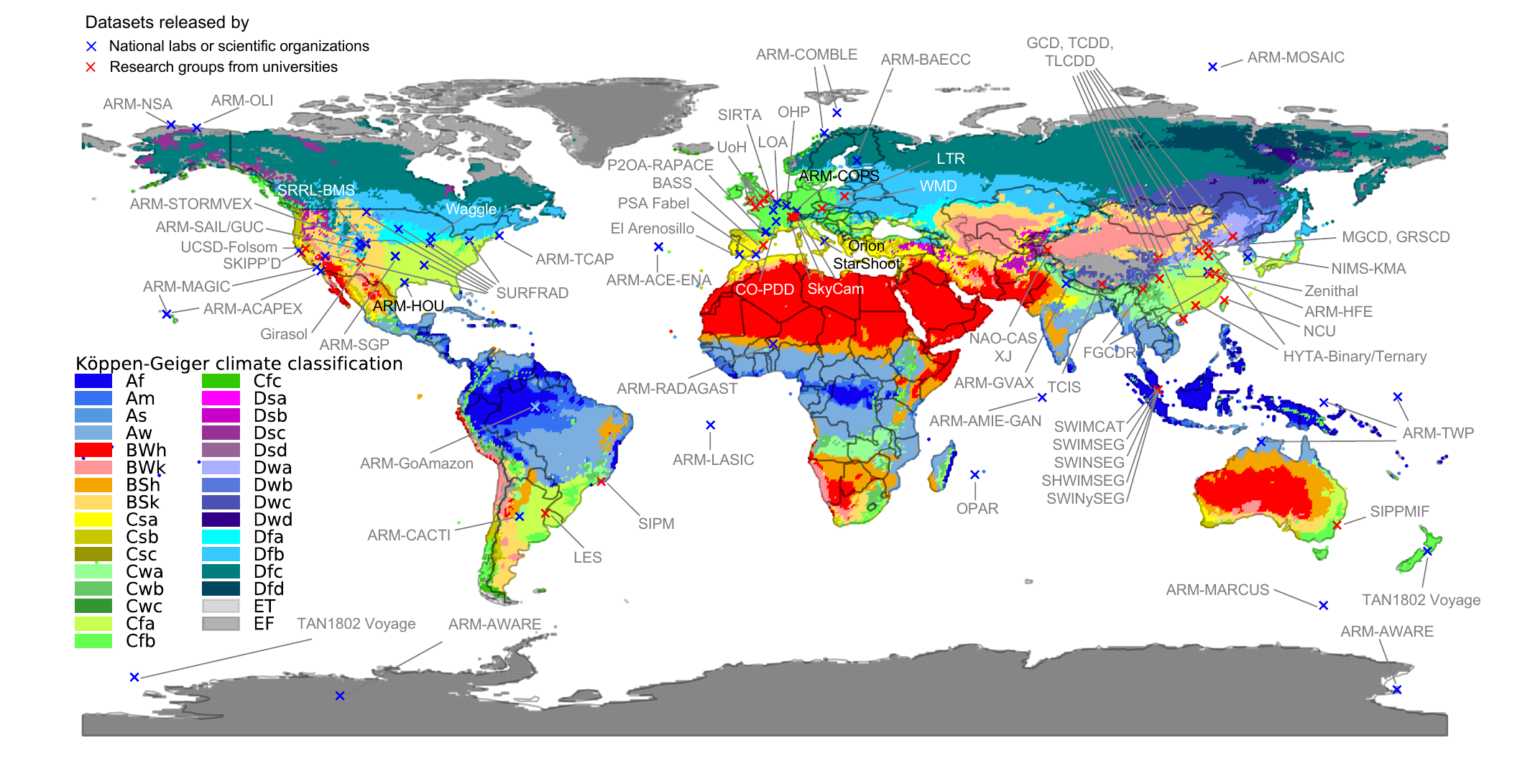}
	  \caption{Köppen–Geiger climate map with geographic locations of the 72 open-source ground-based sky image datasets annotated by blue (for datasets released by national labs or scientific organizations) and red (for datasets released by research groups from universities) cross. A few datasets without geographic coordinates are not shown in the figure, including NAO-CAS, ACS-WSI, CCSN and WSISEG.} \label{fig:dataset_location}
\end{figure}

Figure \ref{fig:dataset_location} shows the geographic locations of the 72 open-source ground-based sky image datasets annotated by blue (for datasets released by national labs or scientific organizations, i.e., type O datasets) and red (for datasets released by research groups from universities, i.e., type R datasets) cross. The map is colored by Köppen–Geiger climate classes, a widely used classification system for climate. It should be noted that here we use the long-term average climate from 1901 to 2010 with the data obtained from \cite{chen2013using}. It can be observed that the datasets' locations cover all 7 continents on the Earth as well as a wide range of climate classes.

Deep learning model performance is heavily relied on the data used for training, not only the amount of data, but also the diversity of the samples. The success of ImageNet \cite{ImageNet2009} in boosting computer vision research has provided a template for solar forecasting and related research fields. The open-source sky image datasets identified in this study can potentially be used for constructing a global dataset consisting of massive and diversified sky image samples with various weather conditions. Moreover, with such a large-scale sky image dataset, pre-training a general solar forecasting model on it and transferring the knowledge to local sites can save data and get a jump start in local model development. A recent study by \citet{transferlearningpaper} suggests that transferring learning from a large and diversified source dataset to a local target dataset can save up to 80\% of the training data while achieving comparable performance for 15-min-ahead solar forecasting. A selection of the suitable datasets, processing and reconciliation are thus critical for building this large-scale dataset. Also, challenges still need to be addressed for transfer learning as the data collected from different locations are more or less heterogeneous, for example, difference in prediction variable (i.e. PV power generation versus solar irradiance), different distribution of measurements given different local weather conditions, different camera types (TSI versus ASI) and orientations, etc. 

\subsection{Datasets usage analysis}
\label{subsec:data_usage_analysis}
For analysis of the datasets usage, we first count their citations, total usage and usage involving sky images as described in section \ref{subsec:dataset_evaluation_criteria}. Table \ref{tab:dataset_usage} shows these three metrics for each one of the open-source datasets we identified. It should be noted that the counts presented here are based on statistics from Google Scholar by 2022-07-01. It's hard to track the data sharing activities between researchers before the datasets publication, and thus we do not track the pre-publication usages of the datasets. Two exceptions are UCSD-Folsom and SKIPP'D, as they mention in their dataset papers \cite{UCSD2019,nie2022skippd} about some research publications enabled by the datasets before the datasets were published. For all 72 open-source datasets, although majority of them have citations, only 25 has been used by the scientific community and 24 has usage counts involving sky images. It can be observed that some datasets, especially those released by national labs or government scientific organizations, such as SURFRAD, SIRTA, and many datasets from ARM program like ARM-COPS, ARM-HFE, ARM-CAP-MBL and ARM-GoAmazon, tend to have high citation and total usage counts, but much less usage involving sky images. These datasets generally provide a wide spectrum of atmospheric and meteorological observations, and according to our further investigation, they are much more frequently used for climate and atmospheric modeling, e.g., surface radiation/temperature modeling, or remote sensing model validation, while limited efforts have been seen using these datasets for image-based solar forecasting or cloud modeling. The underlying reasons for that could be: (1) low temporal resolution image data that can hardly satisfy the need of short-term forecasting, for example, SURFRAD only provide imagery data in 1 hour resolution, although radiation and meteorological measurements are provided in 1$\sim$3 minutes resolution; (2) low exposure of the datasets to solar forecasting community. Some of these datasets indeed have high quality data, for example, SIRTA provide multi-year 1-min ASI as well as meteorological measurement data, but is not well-known by the solar forecasting community. Most of usage is associated with datasets released before 2020, and only 2 datasets released after 2020 have usage counts, which are Girasol and SKIPP'D. A further look shows the usage of these recent released datasets are mostly internal, i.e. the dataset is mainly used by researchers from the same research group who published the datasets. Some frequently used datasets for image-based solar forecasting (and potentially for cloud motion prediction) are SRRL-BMS (14), UCSD-Folsom (11), SKIPP'D (8), SIRTA (7), for cloud segmentation are SWIMSEG (13), HYTA (Binary) (12) and SWINSEG (6) and for cloud classification are SWIMCAT (19), CCSN (10) and TCIS (4).

\def\mybar#1{%%
{\color{blue}\rule{#1mm}{8pt}} #1} 

\begin{ThreePartTable}
\begin{TableNotes}[flushleft]
\item %Note: For datasets that could potentially be used for different applications, e.g., Girasol for SF/CMP and CC, FGCDR for CS and CC, the usages involving sky image in the table are aggregated for different applications rather than separated.
\end{TableNotes}

\begin{longtable}{>{\raggedright}p{0.10\linewidth}>{\raggedright}p{0.23\linewidth}>{\raggedright}>{\raggedright}p{0.08\linewidth}>{\raggedright}>{\raggedright}p{0.10\linewidth}>{\raggedright}p{0.1\linewidth}>{\raggedright\arraybackslash}p{0.23\linewidth}}
\caption{Datasets citations and usages by 2022-07-01} \label{tab:dataset_usage} \\

\toprule
Application & Dataset & Year & Citations & Total usage & Usage involving sky images \\ % Table header row
\midrule 
\endfirsthead

\caption {Datasets citations and usages by 2022-07-01 (continued)}\\
\toprule
Application & Dataset & Year & Citations & Total usage & Usage involving sky images \\ % Table header row
\midrule 
\endhead

\hline
\multicolumn{6}{r}{{Continued on next page}} \\ \hline
\endfoot

\bottomrule
\insertTableNotes
\endlastfoot
\multirow{29}{*}{SF/CMP} & SRRL-BMS \cite{SRRL1981} & 1981 & 45 & 34 & \mybar{14} \\
& SIRTA \cite{SIRTA2005} & 2005 & 235 & 201 & \mybar{7}  \\
& ARM-MASRAD \cite{ARM_MASRAD2005} & 2005 & 2 & 0 & 0 \\
& ARM-RADAGAST \cite{ARM_RADAGAST2008} & 2008 & 54 & 0 & 0  \\
& ARM-STORMVEX \cite{ARM_STORMVEX2010} & 2010 & 2 & 1 & \mybar{1}  \\
& ARM-AMIE-GAN \cite{ARM_AMIE_GAN2011} & 2011 & 4 & 0 & 0  \\
& ARM-COPS \cite{ARM_COPS2011} & 2011 & 209 & 1 & \mybar{1}  \\
& ARM-HFE \cite{ARM_EAST_AIRC2011} & 2011 & 182 & 3 & \mybar{1}  \\
& ARM-GVAX \cite{ARM_GVAX2013} & 2013 & 2 & 0 & 0  \\
& ARM-CAP-MBL \cite{ARM_CAP_MBL2015} & 2015 & 109 & 0 & 0  \\
& ARM-TCAP \cite{ARM_TCAP2015} & 2015 & 37 & 0 & 0  \\
& ARM-BAECC \cite{BAECCA2016} & 2016  & 64 & 1 & \mybar{1}  \\
 & ARM-ACAPEX \cite{ARM_ACAPEX2016} & 2016 & 0 & 0 & 0  \\
& ARM-MAGIC \cite{ARM_MAGIC2016} & 2016 & 5 & 1 & \mybar{1}   \\
& ARM-GoAmazon \cite{ARM_GoAmazon2016} & 2016 & 248 & 1 & \mybar{1}  \\
& ARM-NSA \cite{ARM_NSA2016} & 2016 & 44 & 0 & 0  \\
& ARM-SGP \cite{ARM_SGP2016} & 2016 & 84 & 1 & \mybar{1}   \\
& ARM-TWP \cite{ARM_TWP2016} & 2016 & 27 & 0 & 0 \\
 & ARM-LASIC \cite{ARM_LASIC2018} & 2018  & 1 & 0 & 0  \\
& LES dataset \cite{caldas2019very} & 2019 & 49 & 2 & \mybar{1} \\
 & UCSD-Folsom \cite{UCSD2019} & 2019 & 49 & 18 & \mybar{11}  \\
& UoH \cite{dandiniHaloRatioGroundbased2019} & 2019 & 4 & 0 & 0 \\
& P2OA-RAPACE \cite{lothon2019elifan} & 2019 & 12 & 1 & \mybar{1} \\
& OHP \cite{OHP} & 2019 & 12 & 0 & 0  \\
& OPAR \cite{OPAR} & 2019 & 12 & 0 & 0  \\
& ARM-CACTI \cite{ARM_CACTI2019} & 2019 & 10 & 0 & 0 \\
& ARM-HOU \cite{ARM_HOU2019} & 2019 & 3 & 0 & 0 \\
& ARM-MARCUS \cite{ARM_MARCUS2019} & 2019 & 4 & 0 & 0 \\
& El Arenosillo \cite{trigo2005development, El_Arenosillo} & 2020 & 1 & 0 & 0 \\
 & CO-PDD \cite{baray2020cezeaux} & 2020 & 11 & 0 & 0 \\
\multirow{17}{*}{SF/CMP} & ARM-AWARE \cite{ARM_AWARE2020} & 2020 & 29 & 1 & \mybar{1} \\
& BASS \cite{c2021feasibility} & 2021 & 2 & 1 & \mybar{1} \\
& Girasol \cite{Girasol2021} & 2021 & 7 & 7 & \mybar{2} \\
& SkyCam \cite{SkyCam2021} & 2021 & 0 & 0 & 0 \\
& SIPM \cite{SIPM2021} & 2021 & 0 & 1 & \mybar{1} \\
& SIPPMIF \cite{UOW2021} & 2021 & 2 & 1 & \mybar{1}  \\
& Waggle \cite{park2021prediction} & 2021 & 8 & 1 & \mybar{1} \\
& TAN1802 Voyage \cite{TAN1802_Voyage2021} & 2021 & 3 & 0 & 0 \\
& ARM-SAIL/GUC \cite{ARM_GUC2021} & 2021 & 4 & 0 & 0  \\
& SKIPP'D \cite{nie2022skippd} & 2022 & 0 & 8 & \mybar{8} \\
& ARM-COMBLE \cite{ARM_COMBLE2022} & 2022 & 1 & 0 & 0 \\
& ARM-MOSAIC \cite{ARM_MOSAIC2022} & 2022 & 26 & 0 & 0 \\
& ARM-ACE-ENA \cite{ARM_ACE_ENA2022} & 2022 & 10 & 0 & 0 \\
& Orion StarShoot \cite{Orion_Starshoot} & N/A & N/A & N/A & N/A \\
& LTR \cite{warsaw} & N/A & N/A & N/A & N/A   \\
& LOA \cite{LOA} & N/A & N/A & N/A & N/A   \\
& ARM-OLI \cite{ARM_OLI} & N/A & N/A & N/A &  N/A \\ \noalign{\vspace{1mm}} \hline \noalign{\vspace{1mm}}

\multirow{17}{*}{CS} & SURFRAD \cite{SURFRAD2000} & 2000 & 429 & 286 & \mybar{3}  \\ 
& HYTA (Binary) \cite{li2011hybrid} & 2011 & 182 & 12 & \mybar{12}  \\
& HYTA (Ternary) \cite{dev2015multi} & 2015 & 18 & 3 & \mybar{3}  \\
& NCU \cite{cheng2017cloud} & 2017 & 33 & 1 & \mybar{1} \\
& SWIMSEG \cite{SWIMSEG2017} & 2017 & 87 & 14 & \mybar{14} \\
& SWINSEG \cite{dev2017nighttime} & 2017 & 16 & 7 & \mybar{7} \\ 
& SHWIMSEG \cite{SHWIMSEG2018} & 2018 & 10 & 2 & \mybar{2} \\
& NAO-CAS \cite{shi2019diurnal} & 2019 & 10 & 1 & \mybar{1}  \\
& FGCDR \cite{ye2019supervised} & 2019 & 21 & 2 & \mybar{2}  \\
& SWINySEG \cite{dev2019cloudsegnet} & 2019 & 26 & 5 & \mybar{5} \\
& WSISEG \cite{xie2020segcloud} & 2020 & 20 & 1 & \mybar{1} \\
& WMD \cite{krauz2020assessing} & 2020 & 3 & 1 & \mybar{1} \\
& Girasol \cite{Girasol2021} & 2021 & 7 & 7 & \mybar{5} \\
& TCDD \cite{zhang2021ground} & 2021 & 5 & 1 & \mybar{1} \\
& PSA Fabel \cite{fabel2022applying} & 2022  & 2 & 1 & \mybar{1} \\
& TLCDD \cite{zhang2022ground} & 2022 &  1 & 1 & \mybar{1} \\
& ACS WSI \cite{ye2022self} & 2022 & 0 & 1 & \mybar{1} \\ \hline \noalign{\vspace{1mm}}

\multirow{13}{*}{CC} & SWIMCAT \cite{SWIMCAT2015} & 2015 & 46 & 20 & \mybar{20}  \\
& TCIS \cite{li2016pixels} & 2016 & 21 & 5 & \mybar{5} \\
& Zenithal \cite{luo2018ground} & 2018 & 4 & 1 & \mybar{1} \\
 & CCSN \cite{CCSN2018} & 2018 & 114 & 11 & \mybar{11} \\
& FGCDR \cite{ye2019supervised} & 2019 & 21 & 2 & \mybar{2}  \\
& MGCD \cite{liu2020multi} & 2020 & 8 & 4 & \mybar{4} \\
& GRSCD \cite{liu2020ground} & 2020 & 12 & 1 & \mybar{1} \\
& WMD \cite{krauz2020assessing} & 2020 & 3 & 1 & \mybar{1} \\
& GCD \cite{liu2021ground} & 2021  & 3 & 1 & \mybar{1} \\
& NIMS-KMA \cite{kim2021twenty} & 2021 & 0 & 1 & \mybar{1} \\
& Girasol \cite{Girasol2021} & 2021 & 7 & 7 & \mybar{5} \\
& PSA Fabel \cite{fabel2022applying} & 2022  & 2 & 1 & \mybar{1} \\
& NAO-CAS XJ \cite{li2022all} & 2022 &  2 & 1 & \mybar{1} \\
\end{longtable}
\end{ThreePartTable}

\begin{figure}[h!]
\centering
%\includeinkscape[inkscapelatex=false,width=1.0\textwidth]{Figure/Dataset_citing_studies_dendrogram_final.svg}
%\includesvg[inkscapelatex=false,width=1.\textwidth]{Figure/Dataset_citing_studies_dendrogram_final.svg}
\includegraphics[width=1\textwidth]{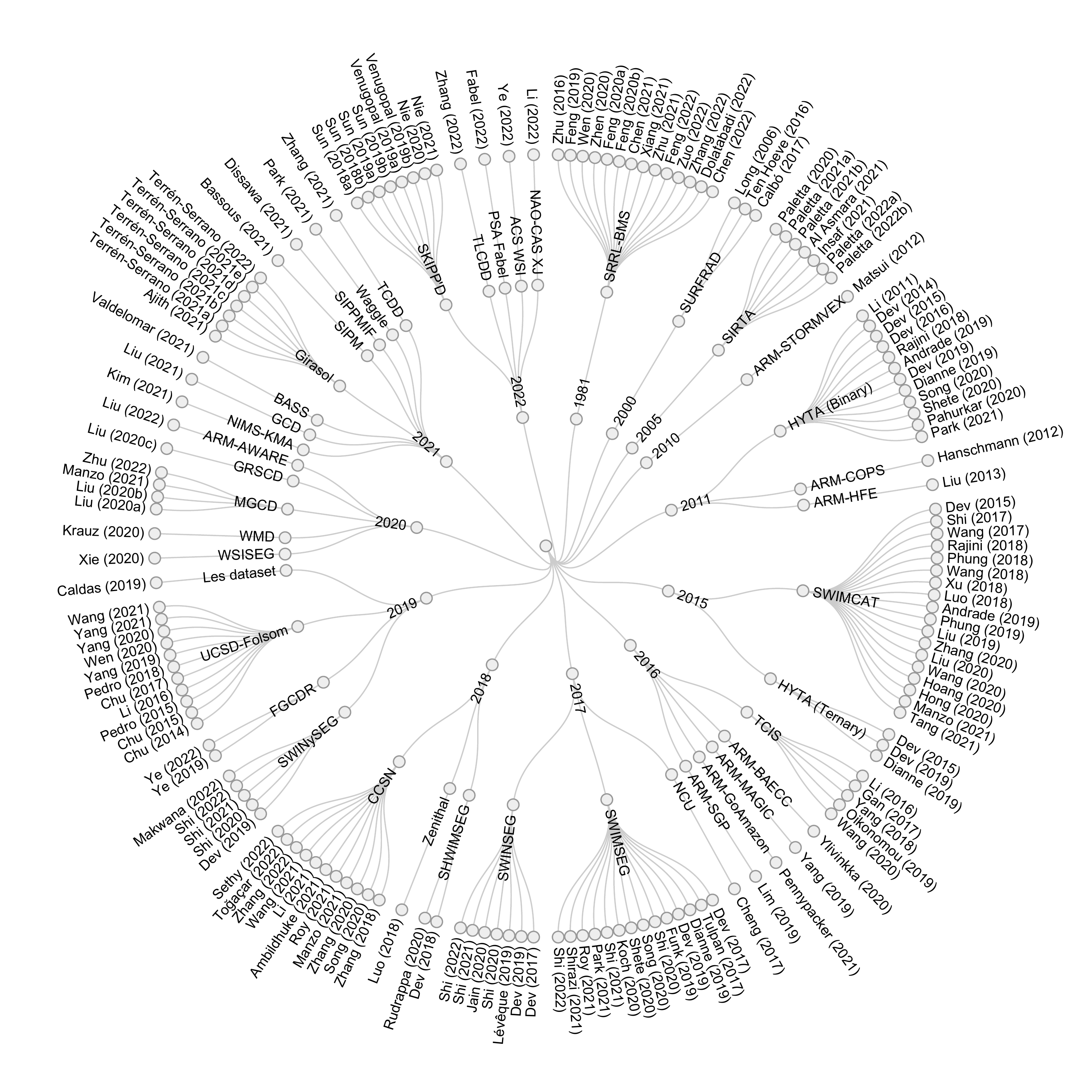}
\caption{Papers that use the open-source datasets involving the use of sky image data in the datasets. The datasets are clustered by their published year. Papers are represented by the last name of the first author as well as the publication years. (detailed references to the papers are presented in the Table \ref{tab:studies_use_sky_image_datasets} in Appendix \ref{sec:appendixB})} \label{fig:papers_use_datasets}
\end{figure}

For datasets with usage involving sky images, we further dive into the studies that use these datasets and see what research topics these studies investigate and what methods they apply. Figure \ref{fig:papers_use_datasets} shows studies that use the datasets involving the usage of sky images with the detail references to these studies listed in Table \ref{tab:studies_use_sky_image_datasets} in Appendix \ref{sec:appendixB}. Figure \ref{fig:topics_1} presents the research topics and applied methods of studies for each dataset that is used. It should be noted that here we ignore the datasets with only one usage count, which in most case is self-usage by the researchers who published the research articles along with their datasets. We categorize the research topics/applications into solar forecasting, cloud segmentation, cloud classification and cloud motion prediction and methods into: prediction type (deterministic and/or probabilistic) and methodology: cloud motion vectors (CMVs), which includes particle image velocimetry and optical flow; machine learning (ML) with or without feature engineering, which include artificial neural networks (ANN), support vector machine (SVM), Random forecast, nearest neighbors, $k$-means and XGBoost; deep learning (DL), which includes CNN, LSTM, 3D-CNN, ConvLSTM, U-Net and transformers; and ensemble modelling which combines multiple types of prediction models to achieve better performance. 

As shown in Figure \ref{fig:topics_1}, the datasets are much less frequently used for cloud motion prediction compared with the other three research topics, solar forecasting, cloud segmentation and classification. To make it clear, the cloud motion prediction mentioned in this study is mainly for deep-learning-based motion prediction based on context sky images, and thus does not include studies using the cloud motion vector methods which linearly extrapolate cloud motion based on consecutive image frames. It's not because the datasets cannot meet the requirement for such research, while it more points to the fact that deep-learning-based cloud motion prediction using sky images is under-studied. It can be also observed that some datasets are versatile and can be used in multiple applications (e.g., SRRL-BMS, SIRTA, etc.), while some are only used in single area (e.g., SURFRAD, HYTA (Ternary), SWINSEG, etc.). This is mostly aligned with our inference based on the data inclusion and temporal resolution of data, although some special cases are found. For example, studies used HYTA (Binary) not only for cloud segmentation \cite{dev2014systematic, dianne2019deep}, but also solar prediction \cite{park2021prediction}. \citet{park2021prediction} first trained and validated a deep learning model to segment sky patches using Waggle, SWIMSEG and HYTA (Binary) datasets, and then the cloud cover ratio estimated by the segmentation model is then used to predict the corresponding solar irradiance level with the Waggle dataset. In terms of the prediction type, much more studies have focused on deterministic prediction (i.e., predict determined values, e.g., irradince values, segmentation map, etc.) than probabilistic prediction (i.e., predict value distributions or probability map, e.g., irradiance prediction interval, probabilistic segmentation map, etc.), although the latter provides more valuable information for risk management for the gird. The underlying reason for this is not yet clear, while \citet{hong2016probabilistic} suggested that it might be the case that probabilistic forecasts were evaluated using the same performance metrics as the deterministic forecasts but perform worse than their deterministic counterpart. More details on the common performance metrics for deterministic and probabilistic forecasting of PV power production can be found in work by \citet{van2018review}. Although we mainly focus on datasets suitable for machine-learning or deep-learning-based methods for solar forecasting and cloud modeling, due to the scope of this study, we do not provide a review of different methodology. Instead, we point the readers to the references of these studies in Table \ref{tab:studies_use_sky_image_datasets} in Appendix \ref{sec:appendixB} for more details.

\begin{figure}[h!] 
\centering

\begin{minipage}[b]{0.32\textwidth}
    \includegraphics[width=1\textwidth]{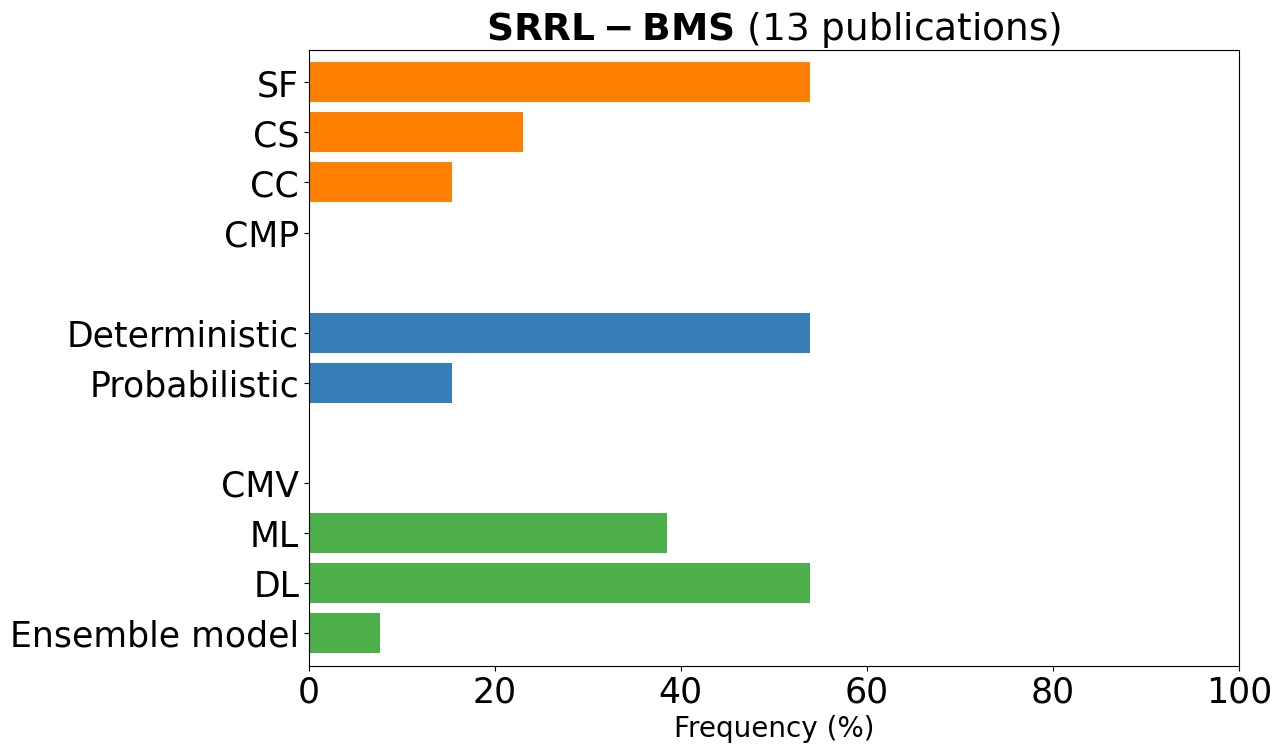}
  \end{minipage}
  \begin{minipage}[b]{0.32\textwidth}
    \includegraphics[width=1\textwidth]{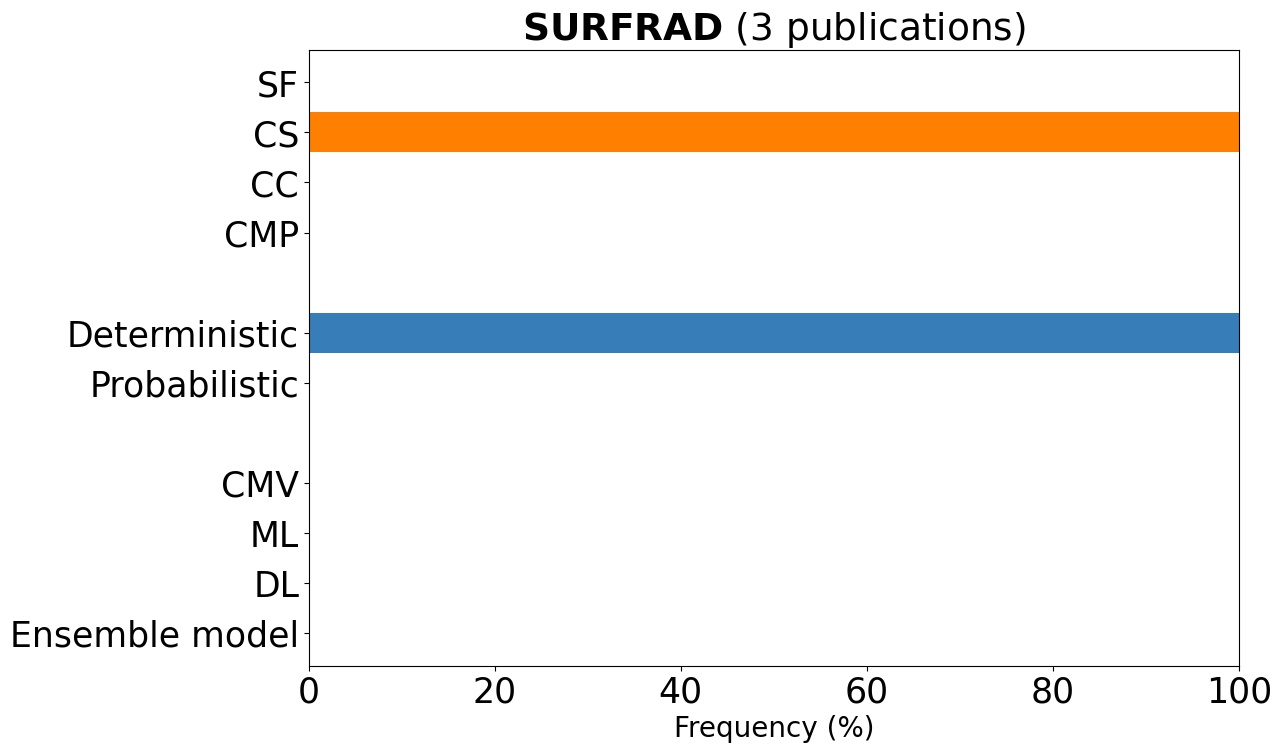}
  \end{minipage}
\begin{minipage}[b]{0.32\textwidth}
    \includegraphics[width=1\textwidth]{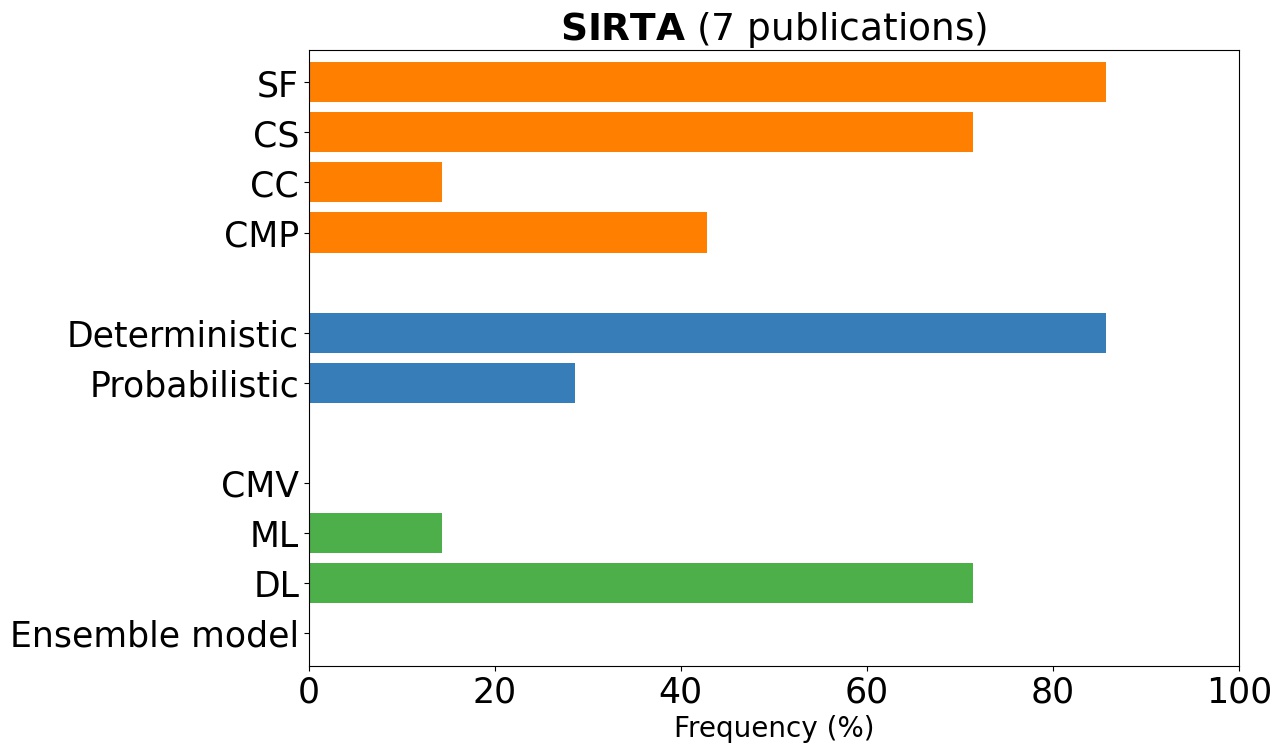}
\end{minipage} 

\begin{minipage}[b]{0.32\textwidth}
    \includegraphics[width=1\textwidth]{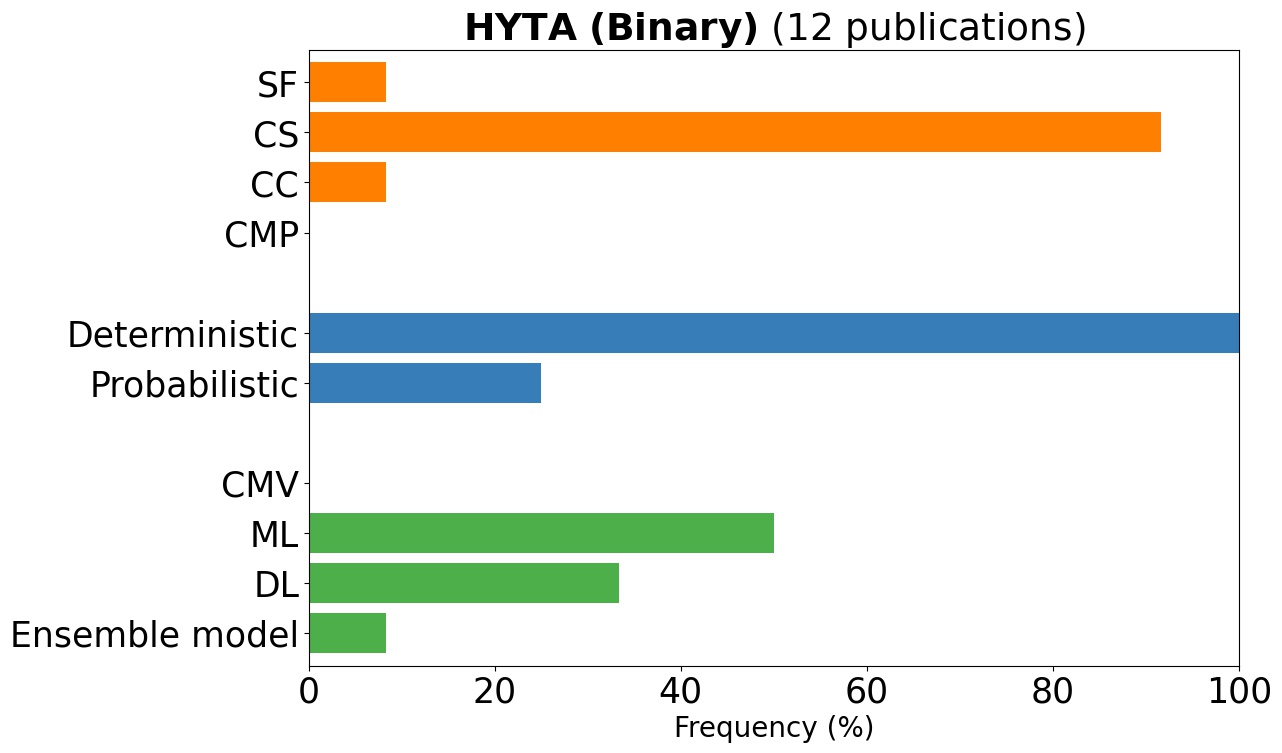}
  \end{minipage}
  \begin{minipage}[b]{0.32\textwidth}
    \includegraphics[width=1\textwidth]{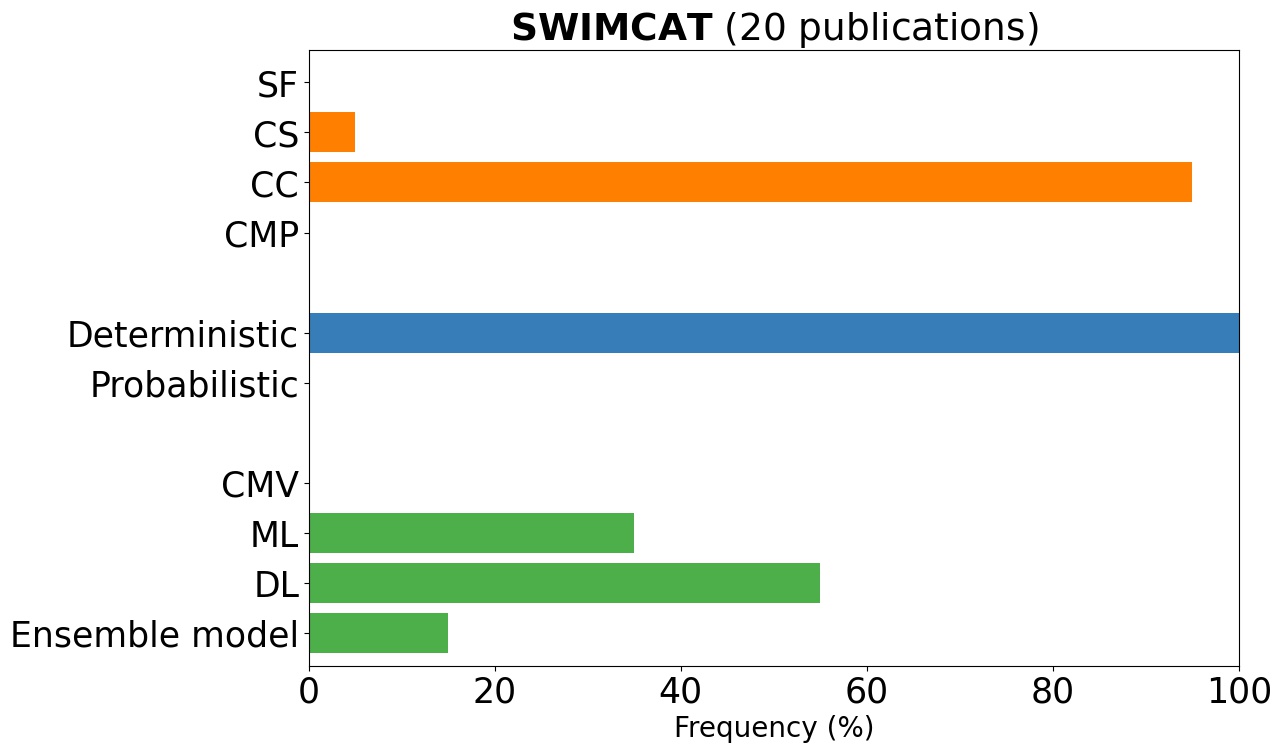}
  \end{minipage}
\begin{minipage}[b]{0.32\textwidth}
    \includegraphics[width=1\textwidth]{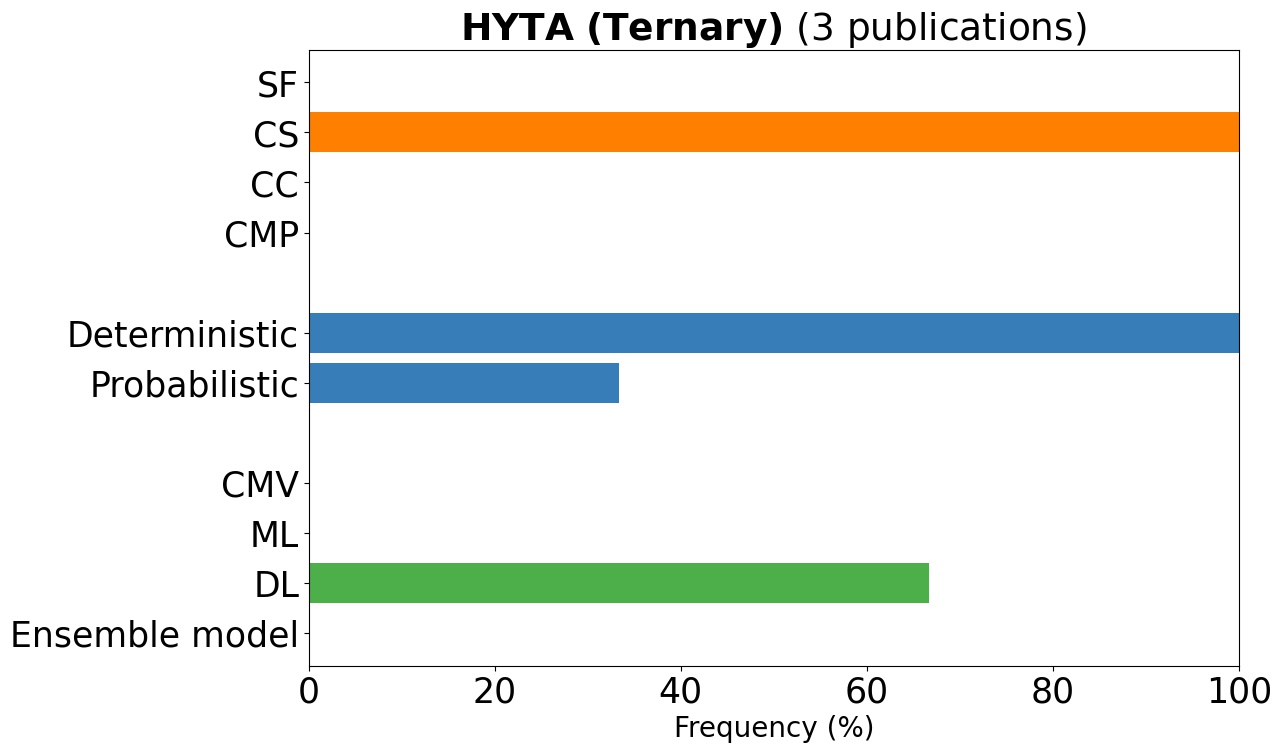}
\end{minipage}

\begin{minipage}[b]{0.32\textwidth}
    \includegraphics[width=1\textwidth]{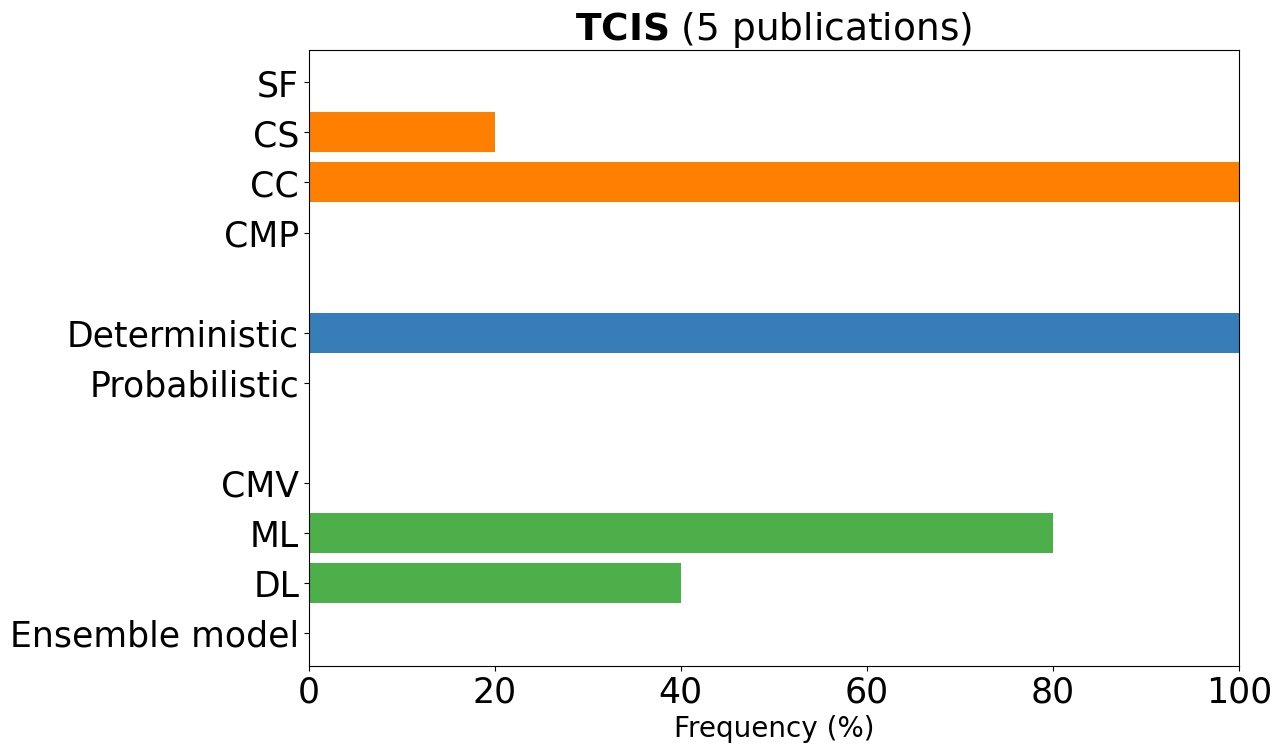}
  \end{minipage}
  \begin{minipage}[b]{0.32\textwidth}
    \includegraphics[width=1\textwidth]{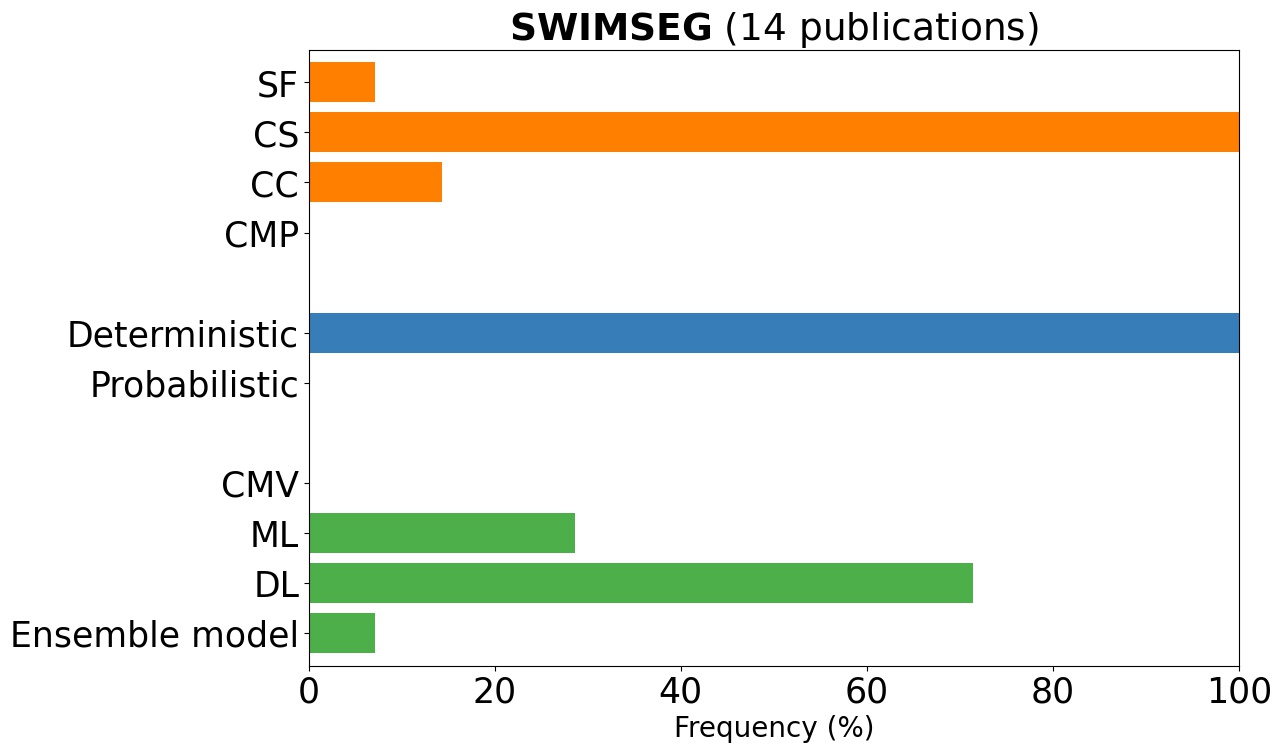}
  \end{minipage}
\begin{minipage}[b]{0.32\textwidth}
    \includegraphics[width=1\textwidth]{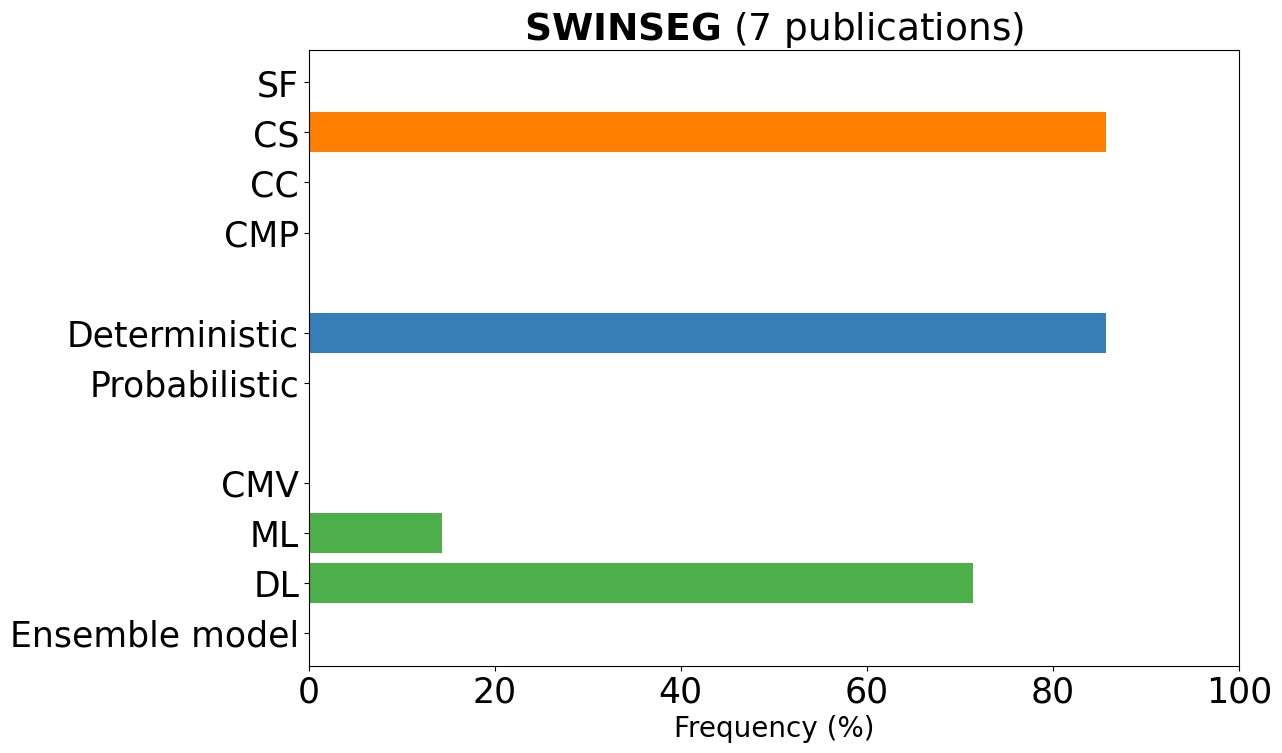}
\end{minipage}

\begin{minipage}[b]{0.32\textwidth}
    \includegraphics[width=1\textwidth]{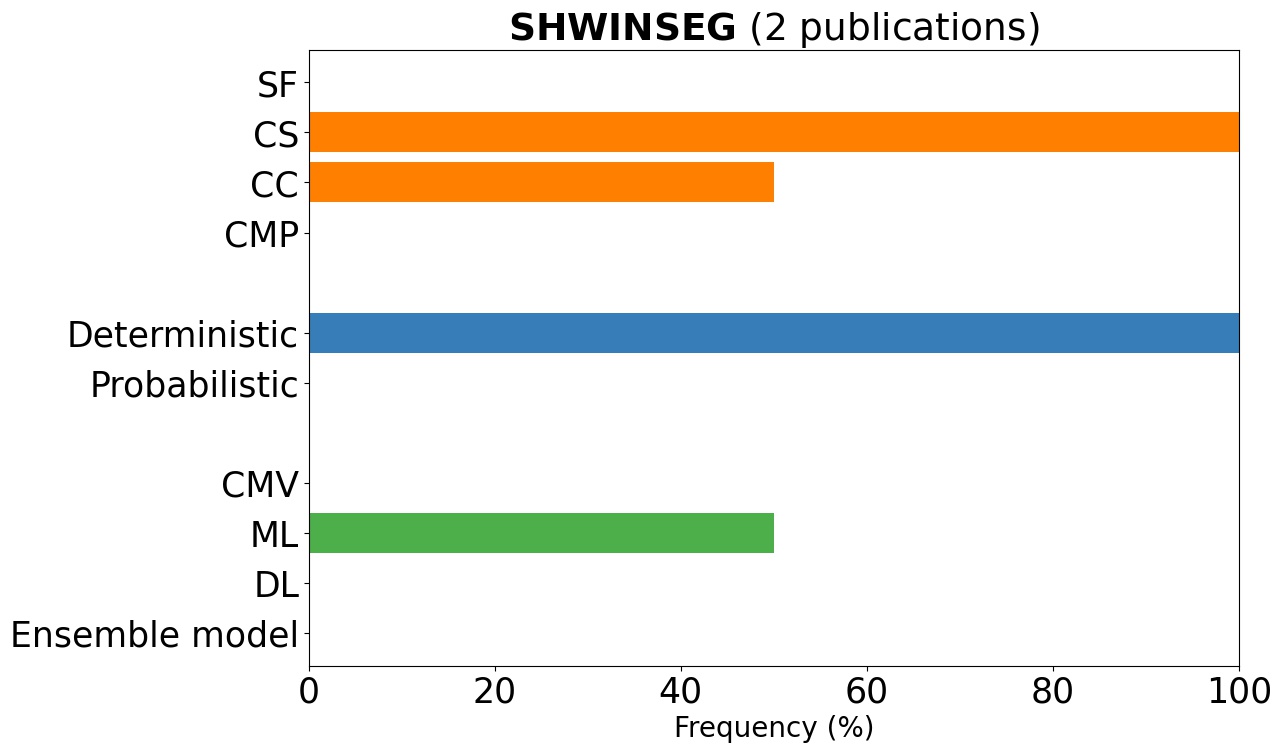}
  \end{minipage}
  \begin{minipage}[b]{0.32\textwidth}
    \includegraphics[width=1\textwidth]{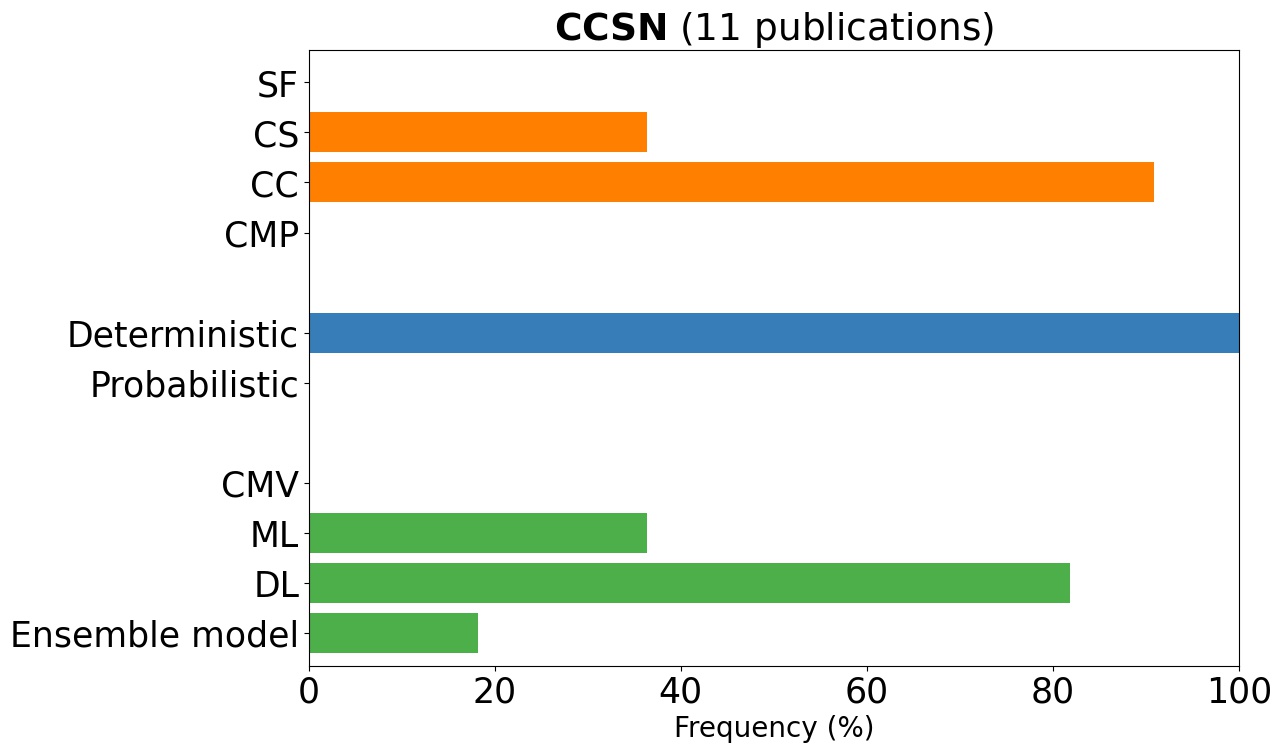}
  \end{minipage}
\begin{minipage}[b]{0.32\textwidth}
    \includegraphics[width=1\textwidth]{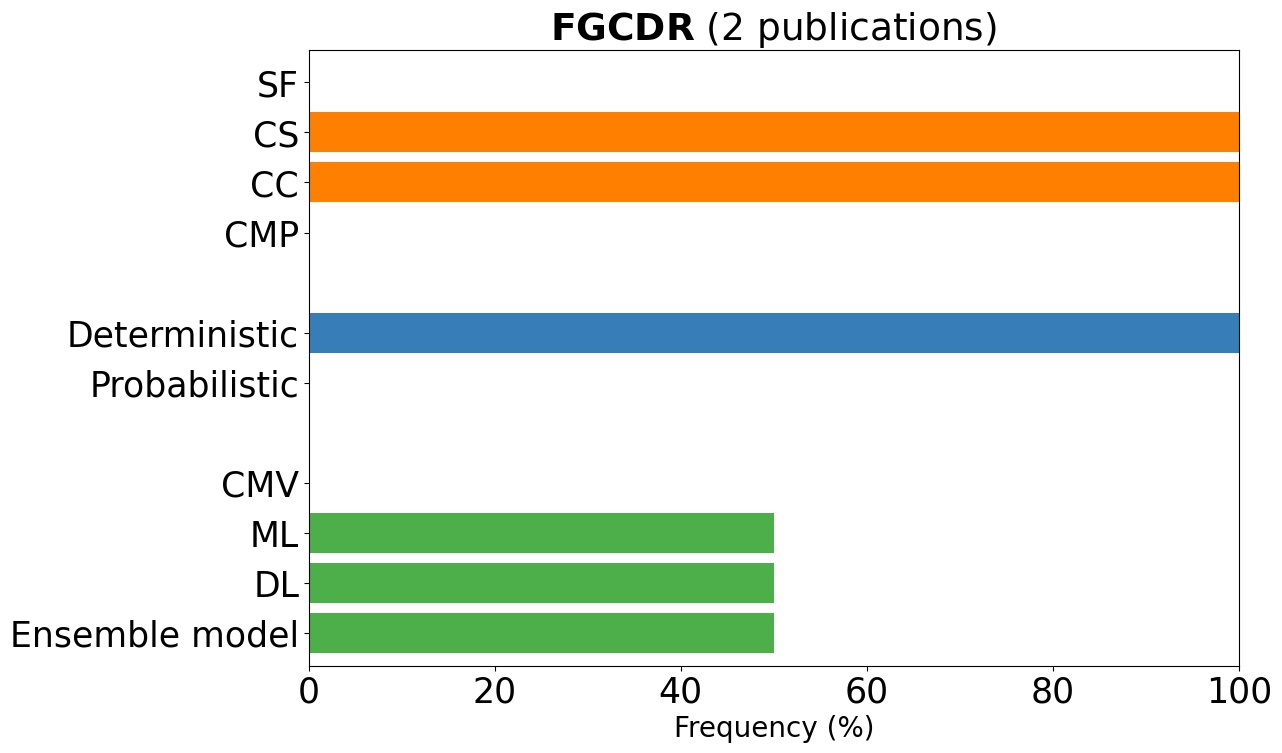}
\end{minipage}

\begin{minipage}[b]{0.32\textwidth}
    \includegraphics[width=1\textwidth]{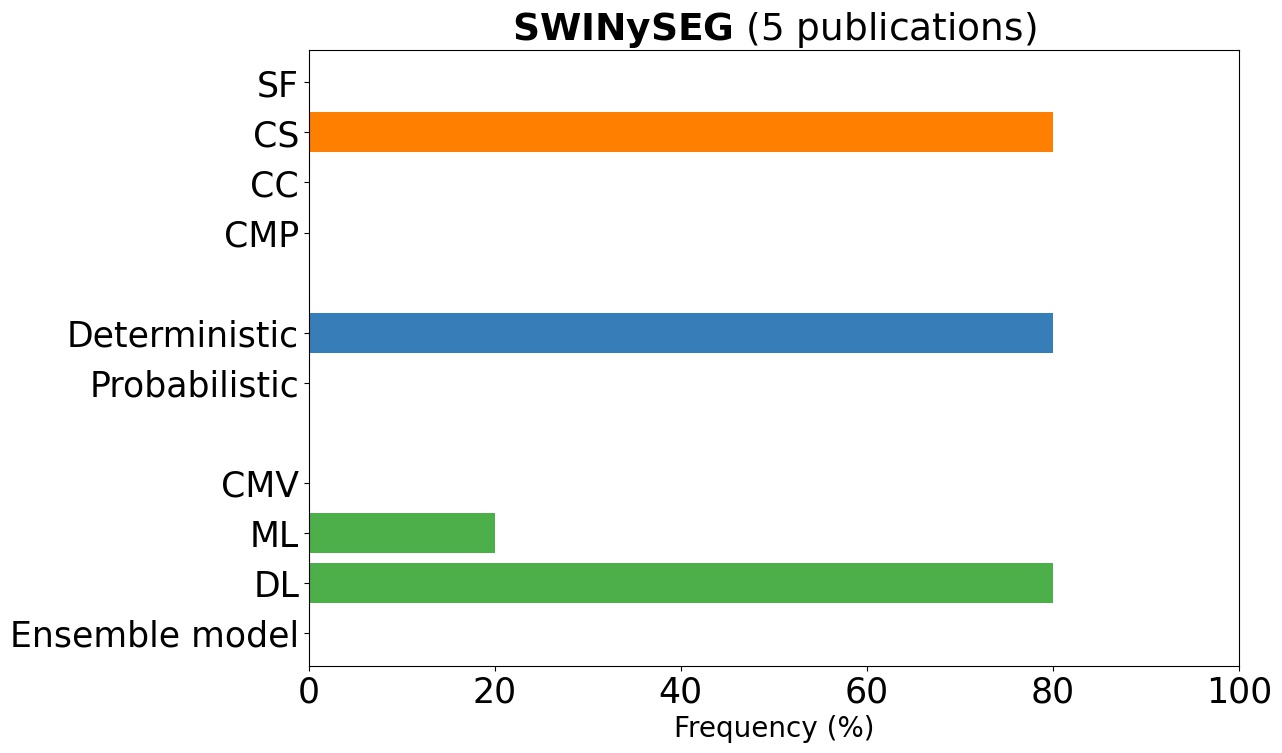}
  \end{minipage}
  \begin{minipage}[b]{0.32\textwidth}
    \includegraphics[width=1\textwidth]{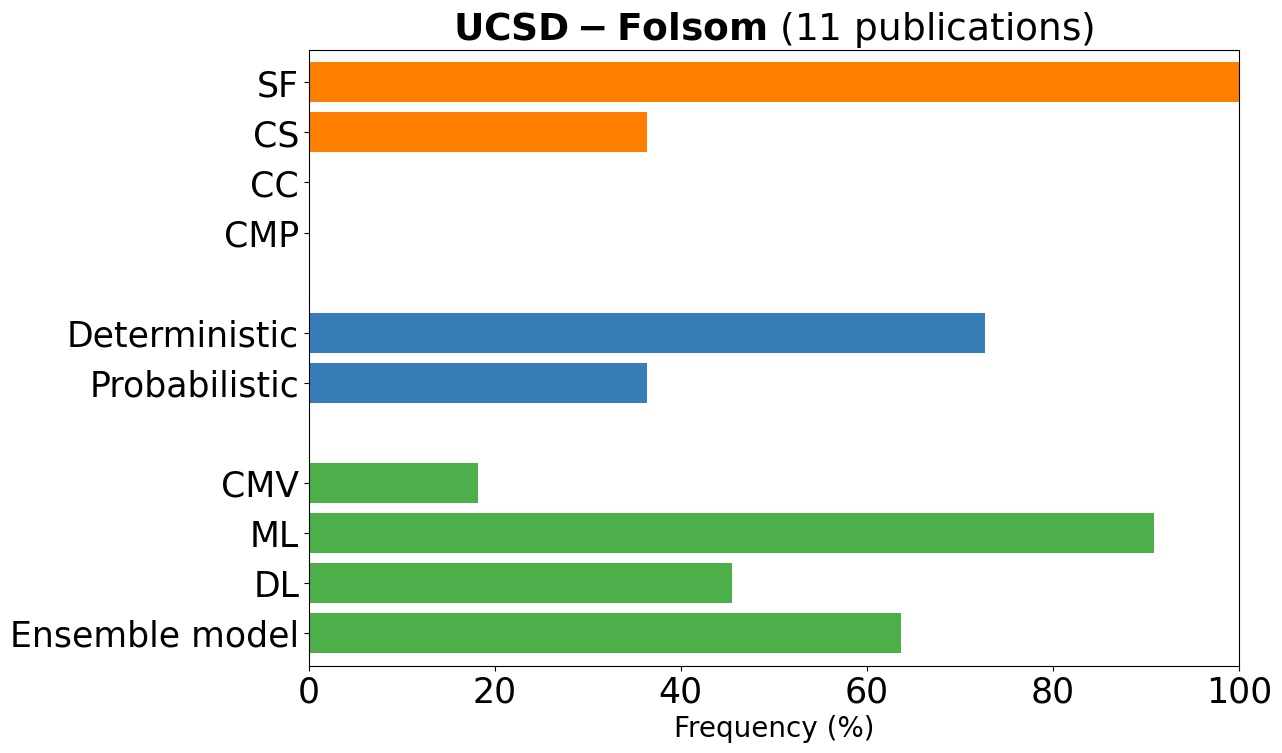}
  \end{minipage}
\begin{minipage}[b]{0.32\textwidth}
    \includegraphics[width=1\textwidth]{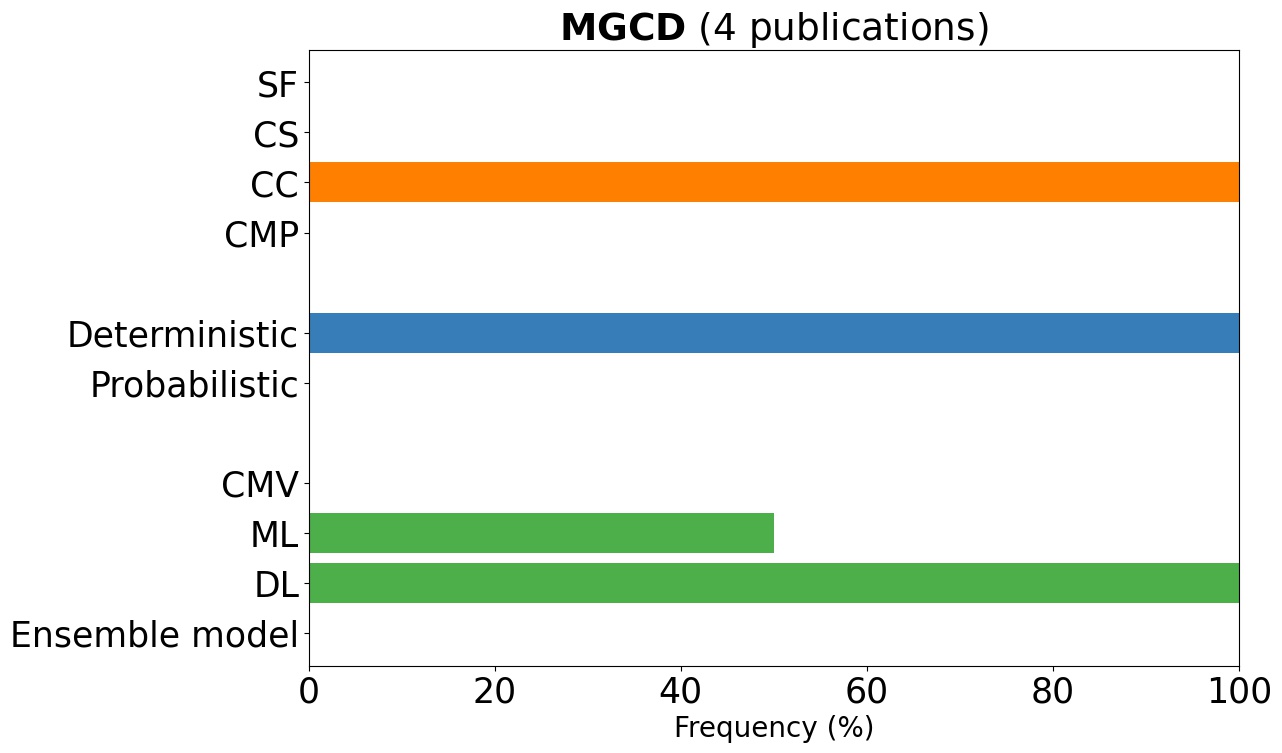}
\end{minipage}

  \begin{minipage}[b]{0.32\textwidth}
    \includegraphics[width=1\textwidth]{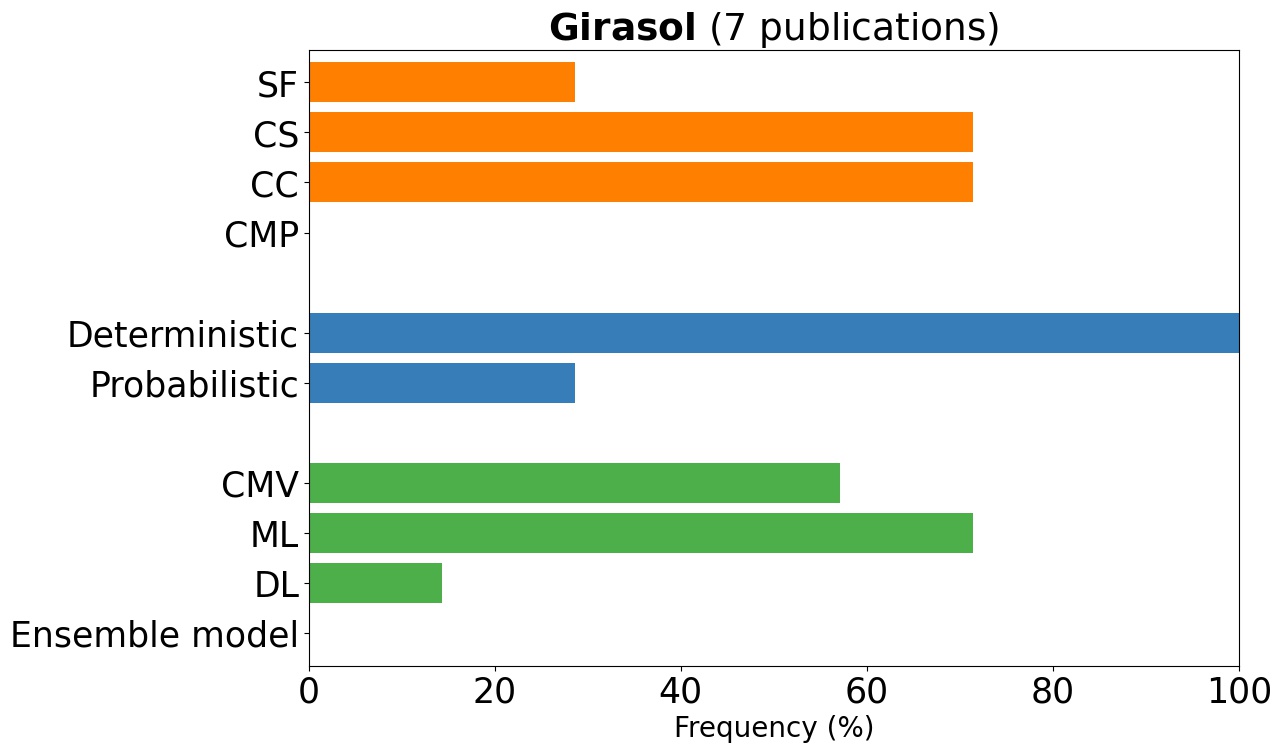}
  \end{minipage}
\begin{minipage}[b]{0.32\textwidth}
    \includegraphics[width=1\textwidth]{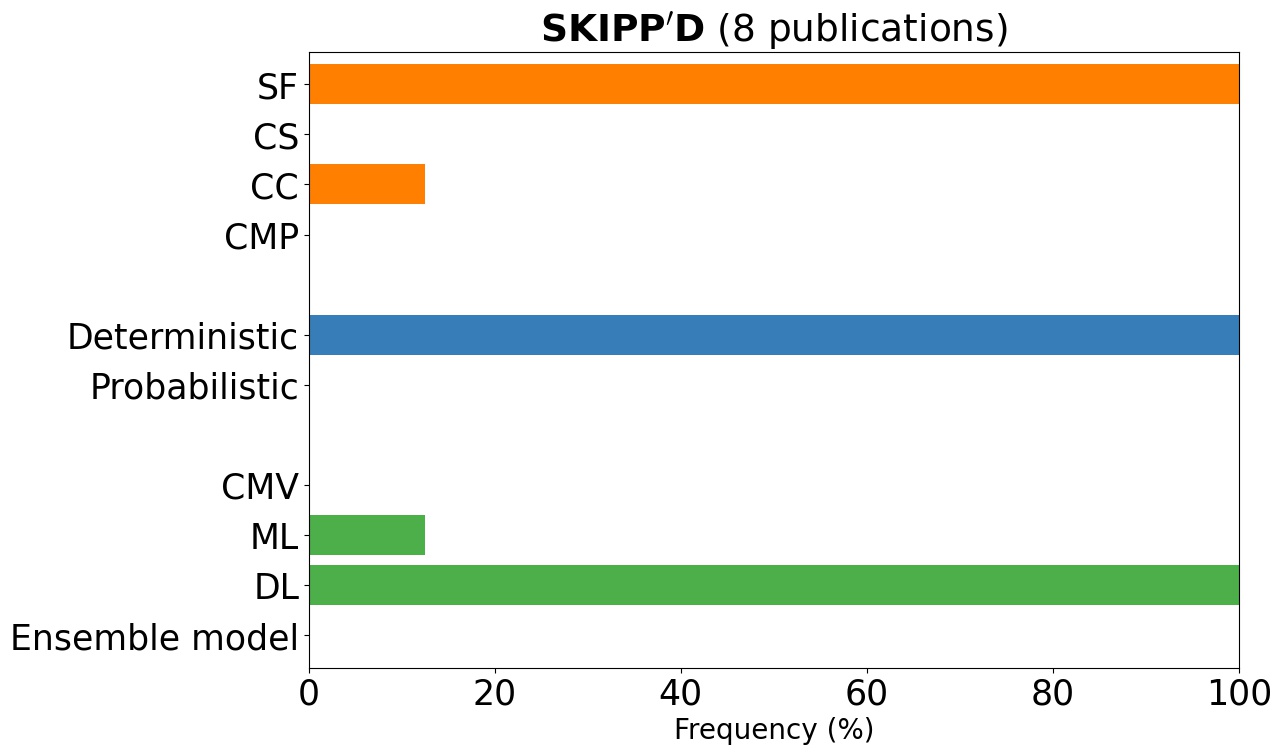}
\end{minipage}

\begin{minipage}[b]{0.40\textwidth}
    \includegraphics[width=1\textwidth]{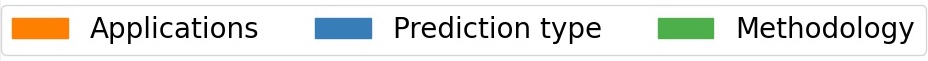}
  \end{minipage}

\caption{Analysis of the research topics and methods of studies that use the datasets involving the usage of sky images. }
\label{fig:topics_1}
\end{figure}

\subsection{Comprehensive dataset evaluation}
In this study, we provide a comprehensive evaluation of each dataset based on the multi-criteria ranking system described in Section \ref{subsec:dataset_evaluation_criteria}. Datasets with different applications are evaluated separately with slightly different criteria. Figure \ref{fig:radar_chart} shows the radar plots of each dataset for (a) solar forecasting and cloud motion prediction, (b) cloud segmentation and (c) cloud classification. It should be noted that for certain evaluation dimensions, if some datasets do not have such information, for example, dataset OPAR \cite{OPAR} does not release any information about the image resolution, it is marked as N/A in the center of the corresponding subfigure. Meanwhile, the names of these datasets (bold font above each radar plot) are marked with a exclamation mark (!). Readers should be cautious when looking at the dimensions of these datasets and do not confuse it with a low rank, and thus should not compare them with the same dimensions of other datasets. 

For solar forecasting and cloud motion prediction, there are totally 47 datasets included in the evaluation, among which 26 are released by ARM programs. It should be noted here SURFRAD are excluded \cite{SURFRAD2000} from the evaluation as it provides low temporal resolution (1 hour) imagery data, which does not satisfy the requirement for very short-term solar forecasting and cloud motion prediction. For non-ARM  datasets, SRRL-BMS, SIRTA, UCSD-Folsom, P2OA-RAPACE, SKIPP'D are among the top choices for ASI datasets, with relatively balanced performance for each dimension. Although most of them suffer from single-site data collection, ways to assemble these datasets can help solve this problem and help build a large-scale centralized datasets. One thing needs to be noted is that SRRL-BMS open sources multi-years of 10-min high resolution sky images from two different sky cameras. The relatively low temporal resolution of the image data can barely satisfy the need of very short-term solar forecasting due to clouds volatility. However, SRRL-BMS provides 1-min high resolution sky images live view on their website \footnote{\url{https://midcdmz.nrel.gov/apps/sitehome.pl?site=SRRLASI}} without archiving them. Web scrapping these 1-min sky images and corresponding irradiance and meteorological measurements could significantly improve the data quality. Girasol provide special infrared images with sun tracking so the sun is always in the center of images. ARM datasets mostly contains high temporal resolution data (30s) and all of them provide TSIs for the imagery data, some ARM datasets with good performance in most of the dimensions are ARM-SGP, ARM-TWP.

For cloud segmentation, no all-round ideal datasets are identified, and all datasets have some limitations. Also, note that a considerable amount of datasets miss the temporal coverage information. SWIMSEG and TLCDD are two good choices among all datasets while both datasets provide SPIs. HYTA dataset are widely used by the cloud segmentation community, while it is limited in dataset size (only 32 images). Datasets provide ASIs are: NCU, NAO-CAS, FGCDR, WSISEG, WMD, PSA Fabel, ACS WSI. Although most of these datasets provide high resolution ASIs, one problem is the limited number of labeled samples, and also they generally provide data collected from a single location. Assembly of these dataset would be potentially be a good solution. 

For cloud classification, SWIMCAT and CCSN are so far the top-two most used dataset, while they are limited in dataset size and image resolution. Also, SPIs are provided by SWIMCAT and CCSN. Some recently released datasets such as FGCDR, MGCD, GRSCD and GCD provide a large amount of samples. Among these datasets, only GCD provides SPIs. All other datasets provide high resolution ASIs. It should be noted that datasets FGCDR, PSA Fabel and WMD provide pixel-level labels for each image samples, hence can also be used for cloud segmentation, while other datasets provide image-level labels.

\section{Conclusion}
\label{sec:conclusion}
In this study, we conducted a survey of open-source sky image datasets, which can potentially be applied in the areas of solar forecasting, cloud segmentation, cloud classification as well as cloud motion prediction. For solar forecasting and cloud motion prediction, we focus on very short-term prediction with forecasting horizon less than 30 minutes. Based on that, we have identified a total of 72 open-source datasets around the world which cover a wide range of climate conditions and include different specifications of imagery and sensor measurement data. We collect extensive information about each dataset and meanwhile provide the ways to access these datasets. According to our screening, 47 datasets are found suitable for solar forecasting and cloud motion prediction, 15 datasets can be used for cloud segmentation and 13 datasets are for cloud classification. We then propose a multi-criteria ranking system for evaluating all the datasets we identified, which covers different aspects of the datasets: comprehensiveness, quality control, temporal and spatial coverage, temporal resolution, image resolution, dataset size as well as dataset usage by the scientific community. The assessment are conducted separately for individual applications based on slightly different criteria. According to the evaluation, we provide insights to the users on the choices of datasets for each of these application fields. Further, we highly suggest the efforts in assembling multiple suitable datasets to build a centralized large-scale dataset to overcome the limitation of individual datasets, e.g., the spatial and temporal coverage as well as the dataset size. We hope this paper can serve as a one-stop shop for researchers who are looking for datasets for training deep learning models for very short-term solar forecasting and relevant areas.

\begin{figure}[h]
\centering
\captionsetup[subfigure]{justification=centering}
\begin{subfigure}[b]{1.0\textwidth}
\centering
\includegraphics[width=1.0\textwidth]{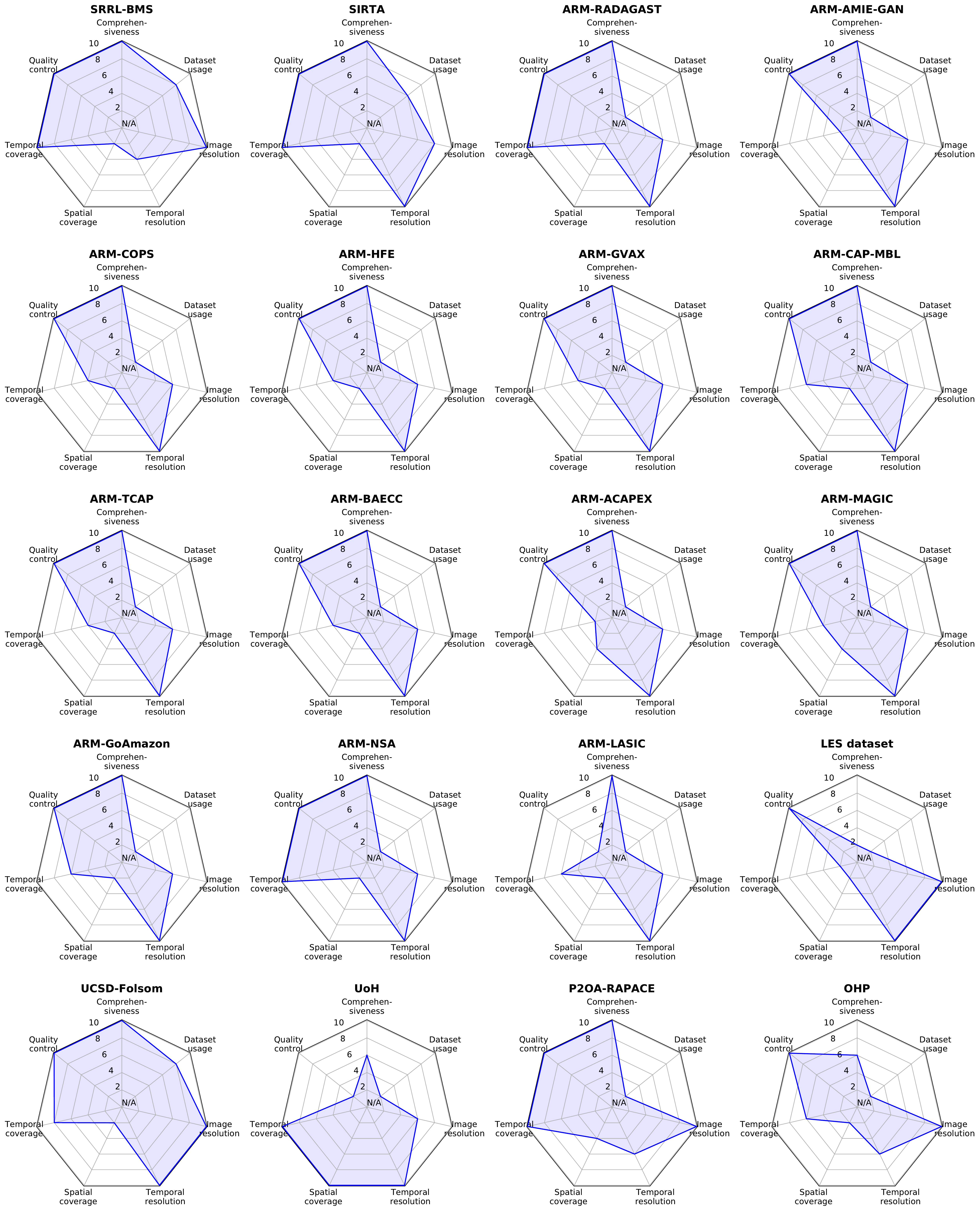}
\subcaption{Datasets for solar forecasting and cloud motion prediction} \label{fig:Radar_map_SFandCMP}
\end{subfigure}
\end{figure}

\begin{figure}[h]
\centering
\captionsetup[subfigure]{justification=centering}
\begin{subfigure}[b]{1.0\textwidth}
\centering
\includegraphics[width=1.0\textwidth]{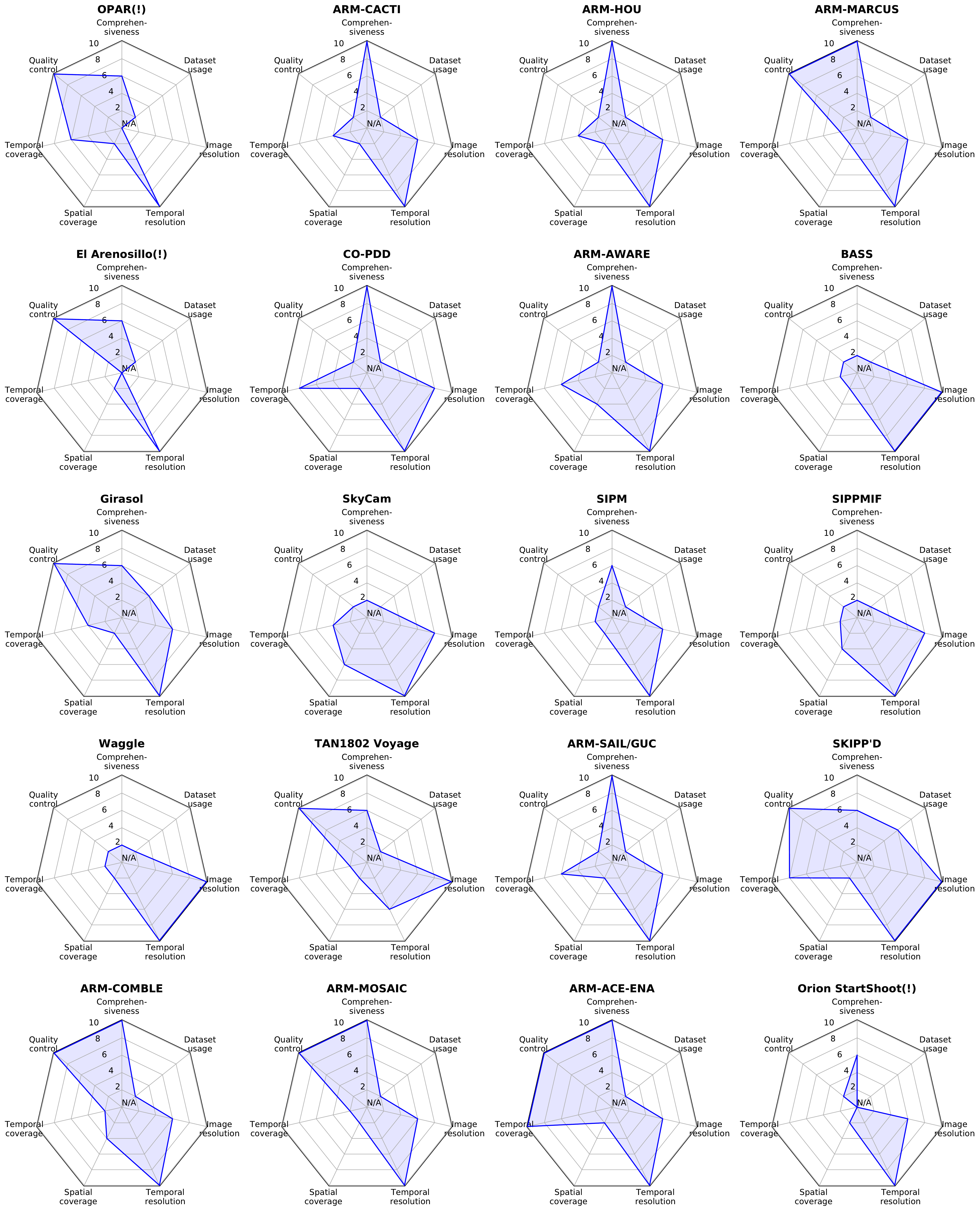}
\subcaption{Datasets for solar forecasting and cloud motion prediction (continued)}
\end{subfigure}
\end{figure}

\begin{figure}[h]
\centering
\captionsetup[subfigure]{justification=centering}
\begin{subfigure}[b]{1.0\textwidth}
\centering
\includegraphics[width=1.0\textwidth]{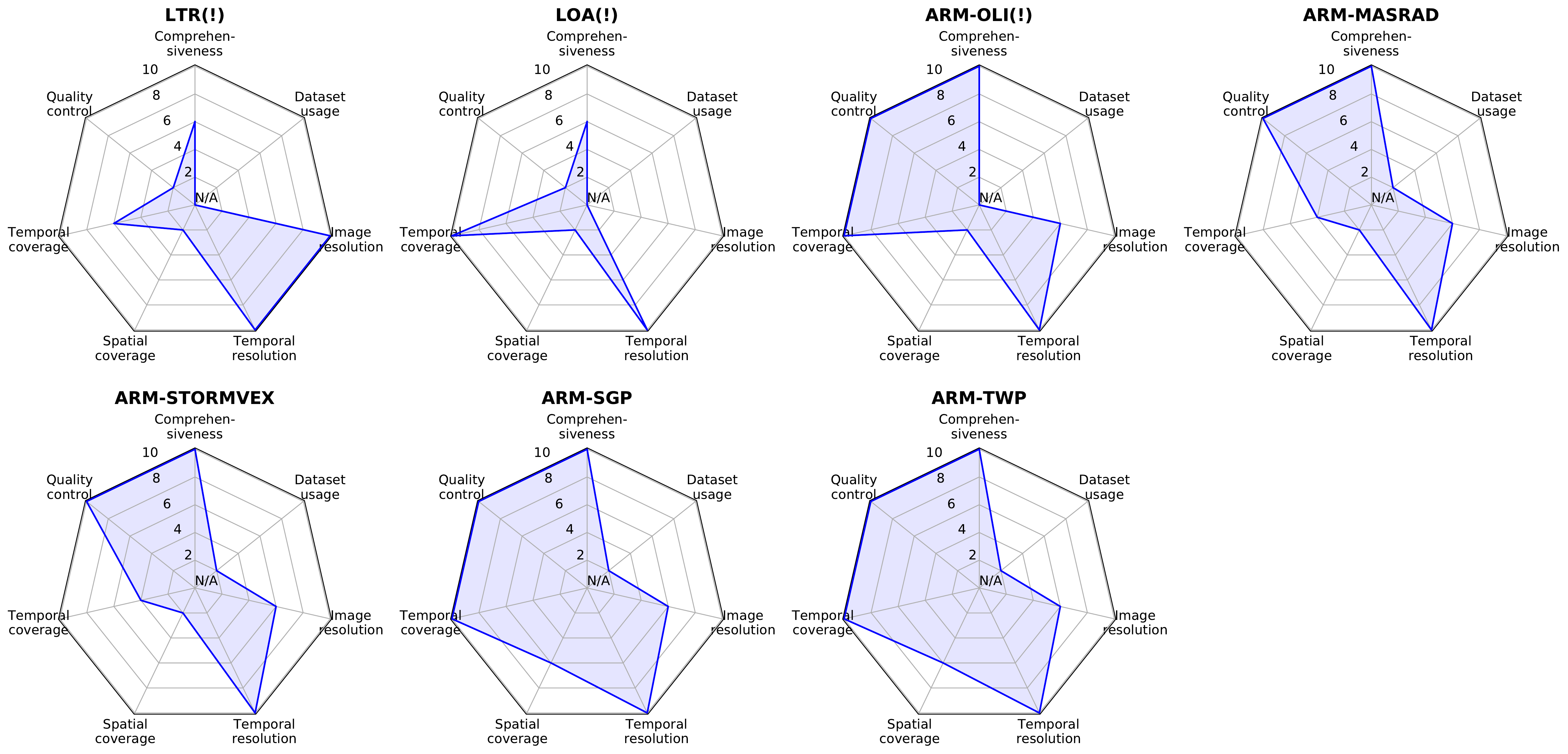}
\subcaption{Datasets for solar forecasting and cloud motion prediction (continued)}
\end{subfigure}
\end{figure}

\begin{figure}[h]
\ContinuedFloat
\captionsetup[subfigure]{justification=centering}
\begin{subfigure}[b]{1.0\textwidth}
	\centering
		\includegraphics[width=1.0\textwidth]{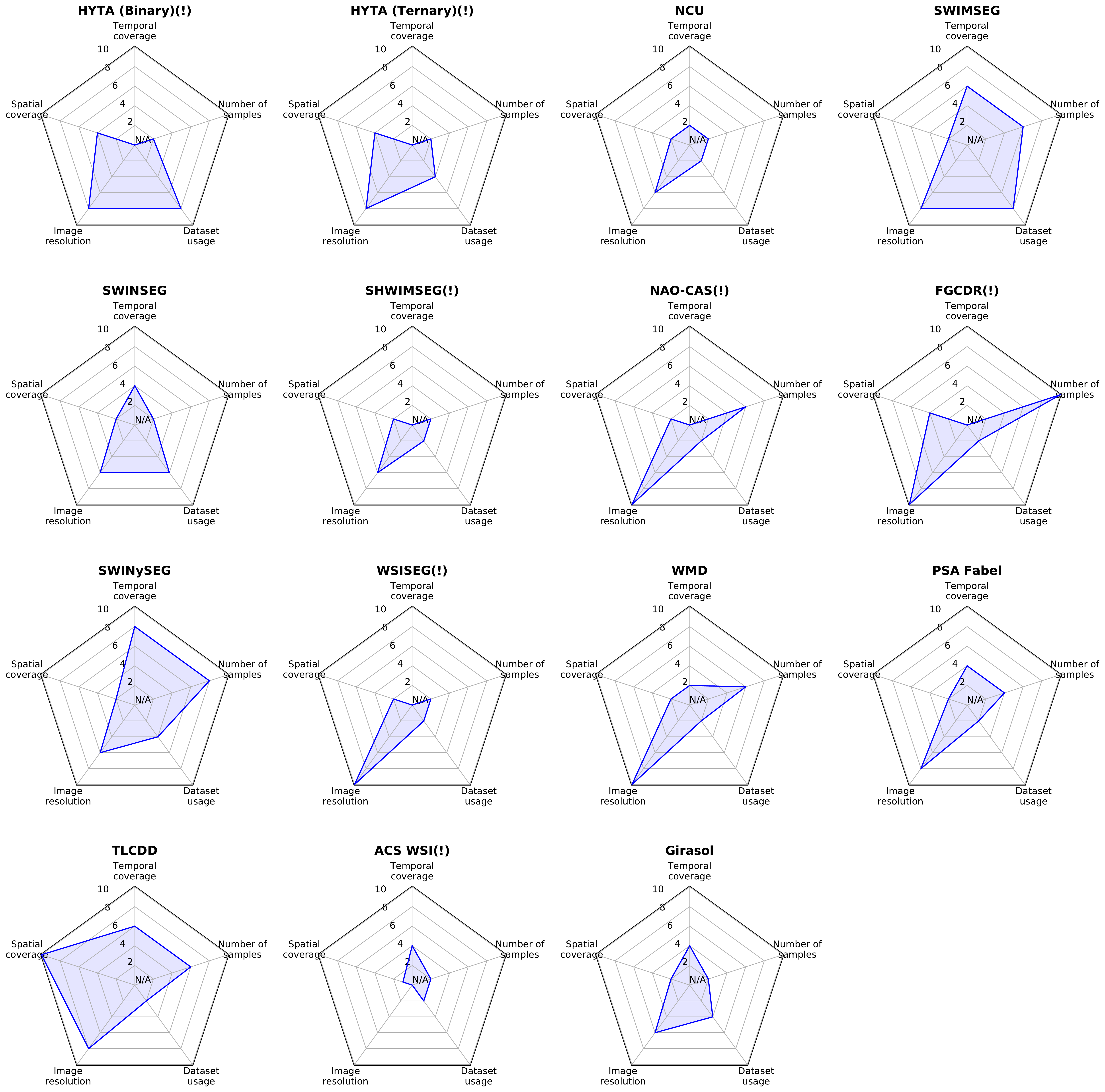}
	  \subcaption{Datasets for cloud segmentation}
\end{subfigure}
\end{figure}

\renewcommand*{\thesubfigure}{c}
\begin{figure}[h]
\vspace{-1.5cm}
\captionsetup[subfigure]{justification=centering}
\ContinuedFloat
\begin{subfigure}[b]{1.0\textwidth}
	\centering
		\includegraphics[width=1.0\textwidth]{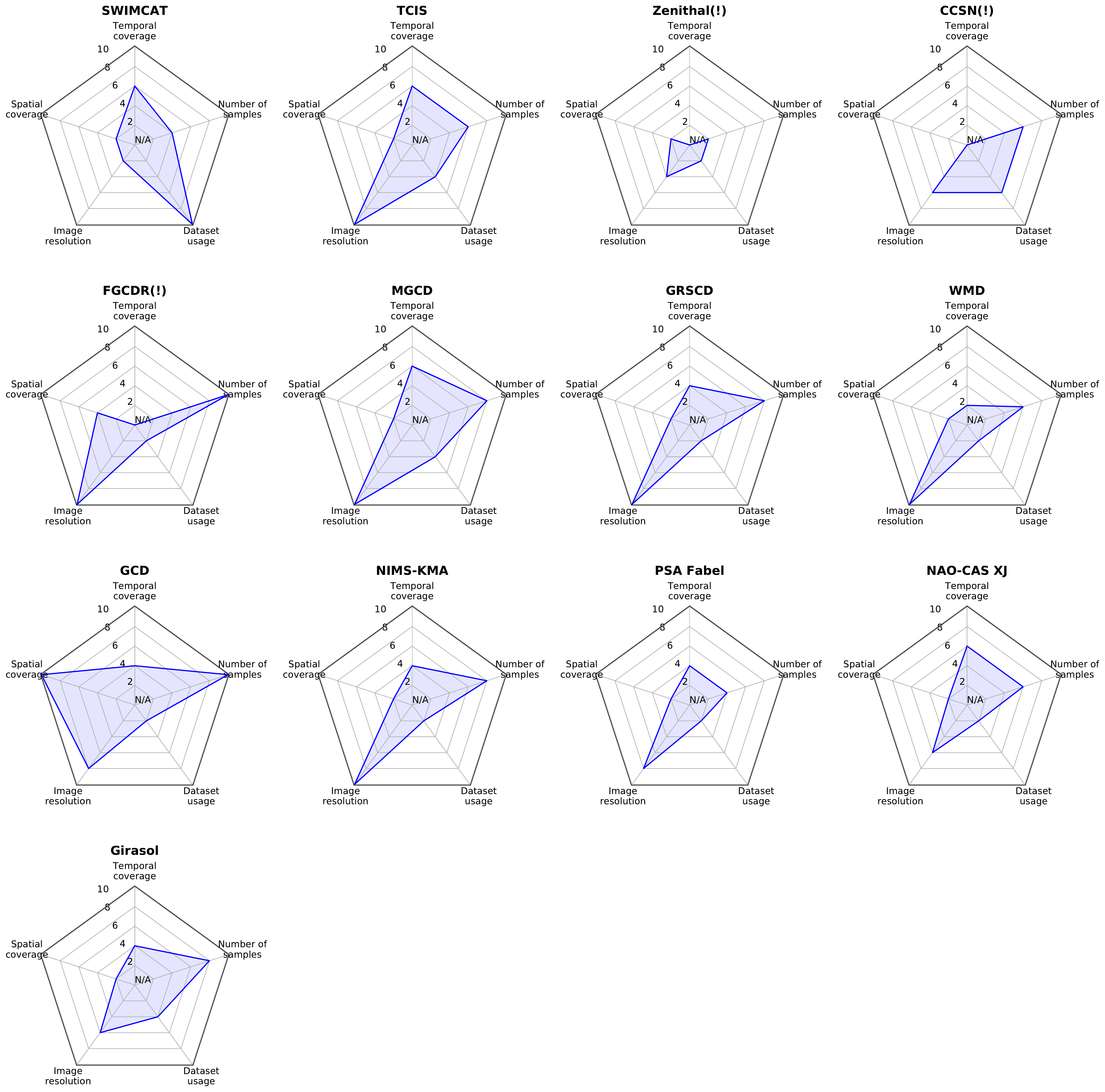}
	  \subcaption{Datasets for cloud classification}
\end{subfigure}
\caption{Comprehensive evaluation of open-source datasets for different applications. (a) datasets for solar forecasting and cloud motion prediction, (b) cloud segmentation and (c) cloud classification. Note the evaluation criteria is a bit different for each application. For certain evaluation dimensions, if datasets do not have such information, it is marked as N/A in the center of the corresponding subfigure and the names of these datasets are marked with a exclamation mark (!). Readers should be cautious when looking at the dimensions of these datasets and do not confuse it with a low rank.} \label{fig:radar_chart}
\end{figure}

\printcredits

\section*{Acknowledgement}
The research was supported by the Dubai Electricity and Water Authority (DEWA) through their membership in the Stanford Energy Corporate Affiliates (SECA) program. This research was sponsored by ENGIE Lab CRIGEN, EPSRC (EP/R513180/1) and the University of Cambridge.

\appendix
\renewcommand{\thetable}{\thesection.\arabic{table}}

\section{Data specifications of sky image datasets}
\label{sec:appendixA}

The data specifications of sky image datasets intended for solar forecasting/cloud motion prediction, cloud segmentation and cloud classification are listed in Table \ref{tab:data_spec_sf_cmp}, \ref{tab:data_spec_cs}, \ref{tab:data_spec_cc}, respectively.

\begin{landscape}
\begin{ThreePartTable}
\begin{TableNotes}[flushleft]
\item $^*$ SRRL-BMS has a live-view of ASI which updates minutely on its website, but these live-view images are not archived.
\item $^{**}$ The segmentation maps provided in Waggle dataset is generated mainly by algorithm with the pixels difficult to separate manually labeled.
\end{TableNotes}
\begin{longtable}{L>{\raggedright}p{0.15\linewidth}>{\raggedright}p{0.15\linewidth}>{\raggedright}p{0.2\linewidth}>{\raggedright}p{0.10\linewidth}>{\raggedright}p{0.03\linewidth}>{\raggedright\arraybackslash}p{0.1\linewidth}}
\caption{Data specifications in sky image datasets intended for using primarily in solar forecasting and cloud motion prediction} \label{tab:data_spec_sf_cmp} \\
\toprule
Dataset & Sky image & Irrad./PV  & Meteo. & SM & CCL & Others \\ 
\midrule 
\endfirsthead

\caption{Data specifications in sky image datasets intended for using primarily in solar forecasting and cloud motion prediction (continued)}\\
\toprule
Dataset & Sky image & Irrad./PV  & Meteo. & SM & CCL & Others \\ 
\midrule 
\endhead

\hline
\multicolumn{7}{r}{{Continued on next page}} \\ \hline
\endfoot

\bottomrule
\insertTableNotes
\endlastfoot

SRRL-BMS\cite{SRRL1981} & 1536$\times$1536 RGB ASI, normal and under exposure: 10-min freq.$^*$; 288$\times$352 RGB TSI: 10-min freq. &  GHI, DNI, DHI, GTI, etc.: 1-min freq.  & Sun angles, temperature, wind speed and direction, pressure, relative humidity, etc.: 1-min freq. & 2-level segmentation (cloud and sky) generated by algorithm & N/A & Cloud fraction derived from sky image \\ \hline \noalign{\vskip 1mm}
SIRTA \cite{SIRTA2005} & 768$\times$1024 RGB ASI, long (1/100 sec) and short (1/2000 sec) exposure: 1 to 2-min freq.; 480$\times$640 RGB TSI: 1-min freq. & Downwelling GHI, DNI, DHI, infrared irradiance; Upwelling GHI, infrared irradiance: 1-min freq.; PV panel testbench  & Temperature, wind speed and direction, pressure, relative humidity, precipitation rate, etc.: 1-min freq.; Aerosol optical depth: $\sim$15-min freq. & 2-level segmentation (cloud and sky) generated by algorithm \cite{lothon2019elifan} & N/A & Cloud fraction derived from sky image \\ \hline \noalign{\vskip 1mm}
ARM-MASRAD \cite{ARM_MASRAD2005} & 480$\times$640 RGB TSI: 30-sec freq. & Shortwave broadband total downwelling irradiance, longwave broadband downwelling irradiance, etc: 1-min freq. & Cloud optical depth, aerosol optical depth, atmospheric temperature, cloud base and top height, radar reflectivity, atmospheric pressure, radar Doppler, etc. & 2-level segmentation (cloud and sky) generated by algorithm & N/A & Cloud fraction derived from sky image \\ \hline \noalign{\vskip 1mm}
ARM-RADAGAST \cite{ARM_RADAGAST2008} & 480$\times$640 RGB TSI: 30-sec freq. & Shortwave narrowband total downwelling irradiance, longwave broadband downwelling irradiance, etc: 1-min freq. & Cloud optical depth, aerosol optical depth, atmospheric temperature, cloud base and top height, radar reflectivity, atmospheric pressure, radar Doppler, etc. & 2-level segmentation (cloud and sky) generated by algorithm & N/A & Cloud fraction derived from sky image \\ \hline \noalign{\vskip 1mm}
ARM-STORMVEX \cite{ARM_STORMVEX2010} & 480$\times$640 RGB TSI: 30-sec freq. & Shortwave broadband total downwelling irradiance, longwave broadband downwelling irradiance, etc: 1-min freq. & Cloud optical depth, aerosol optical depth, atmospheric temperature, cloud base and top height, radar reflectivity, atmospheric pressure, radar Doppler, etc. & 2-level segmentation (cloud and sky) generated by algorithm & N/A & Cloud fraction derived from sky image \\ \hline \noalign{\vskip 1mm}
ARM-AMIE-GAN \cite{ARM_AMIE_GAN2011} & 480$\times$640 RGB TSI: 30-sec freq. & Shortwave broadband total downwelling irradiance, longwave broadband downwelling irradiance, etc: 1-min freq. & Cloud optical depth, aerosol optical depth, atmospheric temperature, cloud base and top height, radar reflectivity, atmospheric pressure, radar Doppler, etc. & 2-level segmentation (cloud and sky) generated by algorithm & N/A & Cloud fraction derived from sky image \\ 

ARM-COPS \cite{ARM_COPS2011} & 480$\times$640 RGB TSI: 30-sec freq. & Shortwave broadband total downwelling irradiance, longwave broadband downwelling irradiance, etc: 1-min freq.  & Cloud base and top height, atmospheric pressure, moisture, temperature, aerosol optical depth, radar Doppler, precipitation, etc. & 2-level segmentation (cloud and sky) generated by algorithm & N/A & Cloud fraction derived from sky image \\ \hline \noalign{\vskip 1mm}
ARM-HFE\cite{ARM_EAST_AIRC2011} & 480$\times$640 RGB TSI: 30-sec freq. & Shortwave broadband total downwelling irradiance, Longwave broadband downwelling irradiance, etc: 1-min freq. & Cloud optical depth, aerosol optical depth, atmospheric temperature, cloud base and top height, radar reflectivity, atmospheric pressure, radar Doppler. & 2-level segmentation (cloud and sky) generated by algorithm & N/A & Cloud fraction derived from sky image \\ \hline \noalign{\vskip 1mm}
ARM-GVAX\cite{ARM_GVAX2013} & 480$\times$640 RGB TSI: 30-sec freq. & Shortwave broadband total downwelling irradiance, Longwave broadband downwelling irradiance, etc: 1-min freq. & Cloud optical depth, aerosol optical depth, atmospheric temperature, cloud base and top height, radar reflectivity, atmospheric pressure, radar Doppler, etc. & 2-level segmentation (cloud and sky) generated by algorithm & N/A & Cloud fraction derived from sky image \\ \hline \noalign{\vskip 1mm}
ARM-CAP-MBL \cite{ARM_CAP_MBL2015} & 480$\times$640 RGB TSI: 30-sec freq. & Shortwave broadband total downwelling irradiance, Longwave broadband downwelling irradiance, etc: 1-min freq. & Cloud optical depth, aerosol optical depth, atmospheric temperature, cloud base and top height, radar reflectivity, atmospheric pressure, radar Doppler, etc. & 2-level segmentation (cloud and sky) generated by algorithm & N/A & Cloud fraction derived from sky image \\  \hline \noalign{\vskip 1mm}
ARM-TCAP \cite{ARM_TCAP2015} & 480$\times$640 RGB TSI: 30-sec freq. & Shortwave broadband total downwelling irradiance, Longwave broadband downwelling irradiance, etc: 1-min freq. & Cloud optical depth, aerosol optical depth, atmospheric temperature, cloud base and top height, radar reflectivity, atmospheric pressure, radar Doppler, etc. & 2-level segmentation (cloud and sky) generated by algorithm & N/A & Cloud fraction derived from sky image \\ \hline \noalign{\vskip 1mm}
ARM-BAECC \cite{BAECCA2016} & 480$\times$640 RGB TSI: 30-sec freq. & GHI, DNI, DHI: 1-min freq.& Cloud optical depth, aerosol optical depth, atmospheric temperature, cloud base and top height, radar reflectivity, atmospheric pressure, radar Doppler, etc. & 2-level segmentation (cloud and sky) generated by algorithm & N/A & Cloud fraction derived from sky image \\

ARM-ACAPEX \cite{ARM_ACAPEX2016} & 480$\times$640 RGB TSI: 30-sec freq. & Shortwave and longwave broadband irradiance, etc: 1-sec freq. & Cloud optical depth, aerosol optical depth, atmospheric temperature, cloud base and top height, radar reflectivity, atmospheric pressure, radar Doppler, etc. & 2-level segmentation (cloud and sky) generated by algorithm & N/A & Cloud fraction derived from sky image \\ \hline \noalign{\vskip 1mm}
ARM-MAGIC \cite{ARM_MAGIC2016} & 480$\times$640 RGB TSI: 30-sec freq. & Shortwave and longwave broadband downwelling irradiance, etc: 6-sec freq. & Cloud optical depth, aerosol optical depth, atmospheric temperature, cloud base and top height, radar reflectivity, atmospheric pressure, radar Doppler, etc. & 2-level segmentation (cloud and sky) generated by algorithm & N/A & Cloud fraction derived from sky image \\ \hline \noalign{\vskip 1mm}
ARM-GoAmazon \cite{ARM_GoAmazon2016} & 480$\times$640 RGB TSI: 30-sec freq. & Shortwave and longwave broadband downwelling irradiance, etc: 1-min freq. & Cloud optical depth, aerosol optical depth, atmospheric temperature, cloud base and top height, radar reflectivity, atmospheric pressure, radar Doppler, etc. & 2-level segmentation (cloud and sky) generated by algorithm & N/A & Cloud fraction derived from sky image \\ \hline \noalign{\vskip 1mm}
ARM-NSA \cite{ARM_NSA2016} & 480$\times$640 RGB TSI: 30-sec freq. & Shortwave and longwave broadband downwelling irradiance, etc: 1-min freq. & Cloud optical depth, aerosol optical depth, atmospheric temperature, cloud base and top height, radar reflectivity, atmospheric pressure, radar Doppler, etc. & 2-level segmentation (cloud and sky) generated by algorithm & N/A & Cloud fraction derived from sky image \\ \hline \noalign{\vskip 1mm}
ARM-SGP \cite{ARM_SGP2016} & 480$\times$640 RGB TSI: 30-sec freq. & Shortwave broadband total downwelling irradiance: 1-min freq. & Cloud optical depth, aerosol optical depth, atmospheric temperature, cloud base and top height, radar reflectivity, atmospheric pressure, radar Doppler, etc.  & 2-level segmentation (cloud and sky) generated by algorithm & N/A & Cloud fraction derived from sky image \\ \hline \noalign{\vskip 1mm}
ARM-TWP \cite{ARM_TWP2016} & 480$\times$640 RGB TSI: 30-sec freq. & Shortwave broadband total downwelling irradiance: 1-min freq. & Cloud optical depth, aerosol optical depth, atmospheric temperature, cloud base and top height, radar reflectivity, atmospheric pressure, radar Doppler, etc.  & 2-level segmentation (cloud and sky) generated by algorithm & N/A & Cloud fraction derived from sky image \\ 

ARM-LASIC \cite{ARM_LASIC2018} & 480$\times$640 RGB TSI: 30-sec freq. & Shortwave broadband total downwelling irradiance, Longwave broadband downwelling irradiance, etc: 5-sec freq. & Cloud optical depth, aerosol optical depth, atmospheric temperature, cloud base and top height, radar reflectivity, atmospheric pressure, radar Doppler, etc. & 2-level segmentation (cloud and sky) generated by algorithm & N/A & Cloud fraction derived from sky image  \\ \hline \noalign{\vskip 1mm}
LES dataset \cite{caldas2019very} & 1920$\times$1280 RGB ASI: 1-min freq. & GHI and other radiation components: 1-min freq. & N/A & N/A & N/A & N/A \\ \hline \noalign{\vskip 1mm}
UCSD-Folsom \cite{UCSD2019} & 1536$\times$1536 RGB ASI: 1-min freq.; Satellite imagery: GOES-15 visible and infrared band, spatial resolution 1km, temporal resolution 30 min; & GHI, DNI, DHI: 1-min freq. &  Temperature, wind speed and direction, maximum wind speed, relative humidity, pressure, precipitation: 1-min freq. & N/A & N/A & NWP; Extraceted sky image and irradiance features \\ \hline \noalign{\vskip 1mm}
UoH \cite{dandiniHaloRatioGroundbased2019} & 640$\times$480 daytime and nighttime RGB ASI: 30-sec freq.  & Irradiance: 1-min freq. & Brightness temperature & N/A & N/A & N/A \\ \hline \noalign{\vskip 1mm}
P2OA-RAPACE \cite{lothon2019elifan} & 2048$\times$1536 RGB ASI: 5-min freq. (15-min freq. until 2017); 2272$\times$1704 RGB ASI: 5-min freq. & Irradiance, infra-red radiation: 1-min freq. & Wind speed, wind profiles, rain, pressure, temperature, humidity: 1-min freq. & 2-level segmentation (cloud and sky) generated by algorithm \cite{lothon2019elifan} & N/A & Cloud fraction derived from sky image  \\ \hline \noalign{\vskip 1mm}
OHP \cite{OHP} & 2048$\times$1536 RGB TSI: 5-min freq. & Irradiance: 1-min freq. & Aerosols & N/A & N/A & N/A \\ \hline \noalign{\vskip 1mm}
OPAR \cite{OPAR} & Alcor System: RGB images: freq. N/A & Irradiance: 1-min freq. & Aerosols & N/A & N/A & N/A \\  \hline \noalign{\vskip 1mm}
ARM-CACTI\cite{ARM_CACTI2019} & 480$\times$640 RGB TSI: 30-sec freq. & Shortwave broadband total downwelling irradiance, Longwave broadband downwelling irradiance, etc: 1-min freq. & Cloud optical depth, aerosol optical depth, atmospheric temperature, cloud base and top height, radar reflectivity, atmospheric pressure, radar Doppler, etc. & 2-level segmentation (cloud and sky) generated by algorithm & N/A & Cloud fraction derived from sky image \\ 

ARM-HOU\cite{ARM_HOU2019} & 480$\times$640 RGB TSI: 30-sec freq. & Shortwave broadband total downwelling irradiance, Longwave broadband downwelling irradiance, etc: 1-min freq. & Cloud optical depth, aerosol optical depth, atmospheric temperature, cloud base and top height, radar reflectivity, atmospheric pressure, radar Doppler, etc. & 2-level segmentation (cloud and sky) generated by algorithm & N/A & Cloud fraction derived from sky image \\ \hline \noalign{\vskip 1mm}
ARM-MARCUS\cite{ARM_MARCUS2019} & 480$\times$640 RGB TSI: 30-sec freq. & Shortwave broadband total downwelling irradiance, Longwave broadband downwelling irradiance, etc: 1-min freq. & Cloud optical depth, aerosol optical depth, atmospheric temperature, cloud base and top height, radar reflectivity, atmospheric pressure, radar Doppler, etc. & 2-level segmentation (cloud and sky) generated by algorithm & N/A & Cloud fraction derived from sky image \\ \hline \noalign{\vskip 1mm}
El Arenosillo \cite{trigo2005development, El_Arenosillo} & RGB images, Night images: freq. N/A & Irradiance: GHI, DNI: 1-min freq. & Lidar profile: 10-min freq. & N/A & N/A & N/A \\ \hline \noalign{\vskip 1mm}
CO-PDD \cite{baray2020cezeaux} & 1024$\times$768 RGB ASI, long and short time exposures: 1 to 2-min freq.; 600$\times$800 webcam images: 10-min freq.  & Irradiance: freq. & Pressure, humidity, temperature, precipitation : 5-min freq.,wind speed and wind profiles, raindrop size distribution, water profiles and columns: 15-min freq., aerosols : 1-min to 1-day freq. & 2-level segmentation (cloud and sky) generated by algorithm \cite{lothon2019elifan}: 10-min freq. & N/A & Cloud fraction derived from sky image \\ \hline \noalign{\vskip 1mm}
ARM-AWARE\cite{ARM_AWARE2020} & 480$\times$640 RGB TSI: 30-sec freq. & Shortwave broadband total downwelling irradiance, Longwave broadband downwelling irradiance, etc: 5-sec freq. & Cloud optical depth, aerosol optical depth, atmospheric temperature, cloud base and top height, radar reflectivity, atmospheric pressure, radar Doppler, etc. & 2-level segmentation (cloud and sky) generated by algorithm  & N/A & Cloud fraction derived from sky image \\ \hline \noalign{\vskip 1mm}
BASS \cite{c2021feasibility} & 1600$\times$1200 HDR RGB ASI: 1-min freq. & GHI, DHI, direct sun and sky radiance in different bands: 1-min freq. & N/A & N/A & N/A & N/A\\ \hline \noalign{\vskip 1mm}
Girasol \cite{Girasol2021} & 450$\times$450 visible HDR gray-scale ASI and 60$\times$80 infrared SPI, with the sun centered in the frame: 15-sec freq. & GSI: 4 to 6 samples per sec. & Sun positions, temperature, dew point, atmospheric pressure, wind direction, wind velocity and relative humidity: 10-min freq. & N/A & N/A & N/A \\ 

SkyCam \cite{SkyCam2021} & 600$\times$600 HDR RGB ASI: 10-sec freq. & Irradiance: 10-sec freq. & N/A & N/A & N/A & N/A \\ \hline \noalign{\vskip 1mm}
SIPM \cite{SIPM2021} & 640$\times$480 RGB ASI: 1-sec freq. & PV power output calculated based on PV voltage and current measurement: 1-sec freq. & N/A & N/A & N/A & PV voltage, current and temperature measurement: 1-sec freq. \\ \hline \noalign{\vskip 1mm}
SIPPMIF \cite{UOW2021} & 1024$\times$768 RGB ASI (two-camera system): 10-sec freq. & PV power output: 1-min freq. & N/A & N/A & N/A & N/A \\ \hline \noalign{\vskip 1mm}
Waggle \cite{park2021prediction} & 2304$\times$1536 (raw), 300$\times$300 (resized) RGB SPI: 15-sec freq. & Irradiance: 15-min freq.; PV power output: 5-sec freq.  & N/A & 2-level segmentation (cloud and sky) generated mainly by algorithm $^{**}$ & N/A & N/A \\ \hline \noalign{\vskip 1mm}
TAN1802 Voyage \cite{TAN1802_Voyage2021} & 3096$\times$2080 HDR RGB ASI: 5-min freq. & Downwelling shortwave and downwelling infrared radiation: 1-min freq. & Air
temperature, dew-point temperature, pressure, wind speed, wind direction, relative humidity, sea surface temperatur: 1-min freq. & 2-level segmentation (cloud and sky) generated by algorithm & N/A & N/A \\ \hline \noalign{\vskip 1mm}
ARM-SAIL/GUC \cite{ARM_GUC2021} & 480$\times$640 RGB TSI: 30-sec freq. & Shortwave broadband total downwelling irradiance: 1-min freq. & Cloud optical depth, aerosol optical depth, atmospheric temperature, cloud base and top height, radar reflectivity, atmospheric pressure, radar Doppler, etc. & 2-level segmentation (cloud and sky) generated by algorithm & N/A & Cloud fraction derived from sky image \\ \hline \noalign{\vskip 1mm}
SKIPP'D \cite{nie2022skippd} & 2048$\times$2048 (raw), 64$\times$64 (resized) RGB ASI: 1-min freq. & PV power output: 1-min freq. & N/A & N/A & N/A & Sky video: 20 frames per sec. \\ 

ARM-COMBLE \cite{ARM_COMBLE2022} & 480$\times$640 RGB TSI: 30-sec freq. & Net broadband total irradiance: 5-sec freq. & Cloud optical depth, aerosol optical depth, atmospheric temperature, cloud base and top height, radar reflectivity, atmospheric pressure, radar Doppler, etc.  & 2-level segmentation (cloud and sky) generated by algorithm & N/A & Cloud fraction derived from sky image \\ \hline \noalign{\vskip 1mm}
ARM-MOSAIC \cite{ARM_MOSAIC2022} & 480$\times$640 RGB TSI: 30-sec freq. & Shortwave broadband total downwelling irradiance: 1-min freq. & Cloud optical depth, aerosol optical depth, atmospheric temperature, cloud base and top height, radar reflectivity, atmospheric pressure, radar Doppler, etc.  & 2-level segmentation (cloud and sky) generated by algorithm & N/A & Cloud fraction derived from sky image \\ \hline \noalign{\vskip 1mm}
ARM-ACE-ENA \cite{ARM_ACE_ENA2022} & 480$\times$640 RGB TSI: 30-sec freq. & Shortwave broadband total downwelling irradiance: 1-min freq. & Cloud optical depth, aerosol optical depth, atmospheric temperature, cloud base and top height, radar reflectivity, atmospheric pressure, radar Doppler, etc.  & 2-level segmentation (cloud and sky) generated by algorithm & N/A & Cloud fraction derived from sky image \\ \hline \noalign{\vskip 1mm}
Orion StarShoot \cite{Orion_Starshoot} & 768$\times$494 RGB ASI, daytime and nighttime: 1-min freq. & GHI, DNI, Diffuse: freq. & Pressure, temperature, humidity, wind, water vapour, aerosols, cloud properties, cloud base: 15-sec to 1-min freq. & N/A & N/A & N/A \\ \hline \noalign{\vskip 1mm}
LTR \cite{warsaw} & 3096$\times$2080 RGB ASI (visible images), 4 cameras (1280$\times$720): freq. N/A & Irradiance, short- and longwave radiation: freq. & Air temperature, pressure, relative humidity, vapor pressure, wind speed and direction, precipitation: 1-min freq. & N/A & N/A & N/A \\ \hline \noalign{\vskip 1mm}
LOA \cite{LOA} & Visible images: freq. & Infrared radiation: freq. & Weather station: aerosol, cloud profile: freq. N/A  & N/A & N/A & N/A \\ \hline \noalign{\vskip 1mm}
ARM-OLI \cite{ARM_OLI} & 480$\times$640 RGB TSI: 30-sec freq. & Shortwave broadband total downwelling irradiance: 1-min freq. & Cloud optical depth, aerosol optical depth, atmospheric temperature, cloud base and top height, radar reflectivity, atmospheric pressure, radar Doppler, etc.  & 2-level segmentation (cloud and sky) generated by algorithm & N/A & Cloud fraction derived from sky image \\ 
\end{longtable}
\end{ThreePartTable}
\end{landscape}

\begin{landscape}
\begin{ThreePartTable}
\begin{TableNotes}[flushleft]
\item $^*$ No image pixel resolution information is provided by the authors of FGCDR dataset. Here, we put an estimation here based on the camera model Nikon D90, which is used for the whole sky imager.
\item $^{**}$ The Girasol dataset article \cite{Girasol2021} does not mention that it includes labeled segmentation maps. However, a follow-up study by the authors \cite{terren2021segmentation} mentioned that it used 12 labeled segmentation maps.
\end{TableNotes}
\begin{longtable}{L>{\raggedright}p{0.15\linewidth}>{\raggedright}p{0.1\linewidth}>{\raggedright}p{0.1\linewidth}>{\raggedright}p{0.2\linewidth}>{\raggedright}p{0.12\linewidth}>{\raggedright\arraybackslash}p{0.05\linewidth}}
\caption{Data specifications in sky image datasets intended for using primarily in cloud segmentation} \label{tab:data_spec_cs} \\
\toprule
Dataset & Sky image & Irrad./PV  & Meteo. & SM & CCL & Others \\ 
\midrule 
\endfirsthead

\caption{Data specifications in sky image datasets intended for using primarily in solar forecasting and cloud motion prediction (continued)}\\
\toprule
Dataset & Sky image & Irrad./PV  & Meteo. & SM & CCL & Others \\ 
\midrule 
\endhead

\hline
\multicolumn{7}{r}{{Continued on next page}} \\ \hline
\endfoot

\bottomrule
\insertTableNotes
\endlastfoot

SURFRAD \cite{SURFRAD2000} & 288$\times$352 RGB TSI: 1-hour freq.  & Downwelling GHI, DNI, DHI, Upwelling GHI, etc.: 1 to 3-min freq. & Sun angles, temperature, wind speed and direction, pressure, relative humidity, etc.: 1 to 3-min freq. & 2-level segmentation (cloud and sky) generated by algorithm & N/A & Cloud fraction derived from sky image \\ \hline \noalign{\vskip 1mm}
HYTA (Binary) \cite{li2011hybrid} & 32 682$\times$512 RGB SPIs & N/A & N/A & 2-level segmentation (cloud and sky) labeled manually & N/A & N/A \\  \hline \noalign{\vskip 1mm}
HYTA (Tenary) \cite{dev2015multi} & 32 682$\times$512 RGB SPIs & N/A & N/A & 3-level segmentation (sky, thin cloud and thick cloud) labeled manually & N/A & N/A \\ \hline \noalign{\vskip 1mm}
NCU \cite{cheng2017cloud} & 250 640$\times$480 RGB ASIs & N/A  & N/A & 2-level segmentation (cloud and sky) labeled manually & N/A & N/A \\ \hline \noalign{\vskip 1mm}
SWIMSEG \cite{SWIMSEG2017} & 1013 600$\times$600 RGB SPIs & N/A & N/A & 2-level segmentation (cloud and sky) labeled manually & N/A & N/A \\ \hline \noalign{\vskip 1mm}
SWINSEG \cite{dev2017nighttime} & 115 500$\times$500 nighttime SPIs & N/A & N/A & 2-level segmentation (cloud and sky) labeled manually & N/A & N/A \\ \hline \noalign{\vskip 1mm}
SHWIMSEG \cite{SHWIMSEG2018} & 52 500$\times$500 HDR RGB SPIs & N/A & N/A & 2-level segmentation (cloud and sky) labeled manually & N/A & N/A \\ \hline \noalign{\vskip 1mm}
NAO-CAS \cite{shi2019diurnal} & 1124 (369 diurnal+755 nocturnal) 1408$\times$1408 RGB ASIs & N/A & N/A & 2-level segmentation (cloud and sky) labeled manually & N/A & N/A \\ \hline \noalign{\vskip 1mm}
FGCDR \cite{ye2019supervised} & 28638 4288$\times$2848$^*$ ASIs & N/A & N/A & 8-level segmentation (cumulus, stratocumulus, stratus, altostratus, altocumulus, cirrocumulus, cirrostratus, cirrus) labeled manually & 8 categories (pixel-level label): same as segmentation map & N/A \\

SWINySEG \cite{dev2019cloudsegnet} & 6768 daytime (1013 original+5056 augmented) and nighttime (115 original+575 augmented) 300$\times$300 SPIs & N/A & N/A & 2-level (cloud and sky) segmentation labeled manually & N/A & N/A \\ \hline \noalign{\vskip 1mm}
WSISEG \cite{xie2020segcloud} & 400 2000$\times$1944 (raw), 480$\times$450 (resized) HDR RGB ASIs & N/A & N/A & 3-level segmentation (cloud, sky and undefined area) labeled manually & N/A & N/A \\  \hline \noalign{\vskip 1mm}
WMD \cite{krauz2020assessing} & 2044 3264$\times$4928 (raw), 1200$\times$800 (resized) RGB ASIs  & N/A & N/A & 4-level segmentation (high-level clouds, low-level cumulus type clouds, rain clouds and clear sky) labeled manually & 4 categories by cloud height (pixel-level label): same as segmentation map & N/A \\ \hline \noalign{\vskip 1mm}
Girasol \cite{Girasol2021} & 12 60$\times$80 infrared SPIs$^{**}$  & N/A & N/A & 2-level segmentation (clouds and clear sky) labeled manually & N/A & N/A \\ \hline \noalign{\vskip 1mm}
TCDD \cite{zhang2021ground} & 2300 512$\times$512 RGB SPIs & N/A & N/A & N/A & 2-level segmentation (cloud and sky) manually labeled & N/A \\ \hline \noalign{\vskip 1mm}
PSA Fabel \cite{fabel2022applying} & 512$\times$512 RGB ASI: 770 labeled and 286477 unlabeled images (for un-supervised pre-training)  & N/A & N/A & 3-level segmentation (low-, middle- and high-layer clouds) manually labeled  & 3 categories (pixel-level label): same as segmentation map & N/A \\ \hline \noalign{\vskip 1mm}
TLCDD \cite{zhang2022ground} & 5000 512$\times$512 RGB SPIs & N/A & N/A & 2-level segmentation (cloud and sky) labeled manually & N/A & N/A \\ \hline \noalign{\vskip 1mm}
ACS WSI \cite{ye2022self} & 10000 RGB ASIs (500 labeled segmentation) & N/A & N/A & 2-level segmentation (cloud and sky) labeled manually & N/A & N/A \\ 

\end{longtable}
\end{ThreePartTable}
\end{landscape}

\begin{landscape}
\begin{ThreePartTable}
\begin{TableNotes}[flushleft]
\item $^*$ No image pixel resolution information is provided by the authors of FGCDR dataset. Here, we put an estimation here based on the camera model Nikon D90, which is used for the whole sky imager.
\item $^{**}$ The Girasol dataset article \cite{Girasol2021} does not mention that it includes cloud category labels. However, a follow-up study by the authors \cite{terren2021comparative} mentioned that it used 8200 labeled images with 4 different sky conditions.
\end{TableNotes}
\begin{longtable}{L>{\raggedright}p{0.15\linewidth}>{\raggedright}p{0.05\linewidth}>{\raggedright}p{0.05\linewidth}>{\raggedright}p{0.16\linewidth}>{\raggedright}p{0.25\linewidth}>{\raggedright\arraybackslash}p{0.05\linewidth}}
\caption{Data specifications in sky image datasets intended for using primarily in cloud classification} \label{tab:data_spec_cc} \\
\toprule
Dataset & Sky image & Irrad./PV  & Meteo. & SM & CCL & Others \\ 
\midrule 
\endfirsthead

\caption{Data specifications in sky image datasets intended for using primarily in cloud classification (continued)}\\
\toprule
Dataset & Sky image & Irrad./PV  & Meteo. & SM & CCL & Others \\ 
\midrule 
\endhead

\hline
\multicolumn{7}{r}{{Continued on next page}} \\\hline
\endfoot

\bottomrule
\insertTableNotes
\endlastfoot

SWIMCAT \cite{SWIMCAT2015} & 784 125$\times$125 RGB SPIs & N/A & N/A & N/A & 5 categories (image-level label) by visual characteristics: clear sky, patterned clouds, thick dark clouds, thick white clouds, and veil
clouds & N/A \\ \hline \noalign{\vskip 1mm}
TCIS \cite{li2016pixels} & 5000 1392$\times$1040 (raw), 821$\times$821 (resized),
ASIs & N/A & N/A & N/A & 5 categories by visual characteristics (image-level label): cirriform, cumuliform, stratiform, clear sky and mixed cloudiness & N/A \\ \hline \noalign{\vskip 1mm}
Zenithal \cite{luo2018ground} & 500 320$\times$240 infrared SPIs & N/A & N/A & N/A & 5 categories by visual characteristics (image-level label): stratiform, cumuliform, waveform, and cirriform clouds and clear sky & N/A \\ \hline \noalign{\vskip 1mm}
CCSN \cite{CCSN2018} & 2543 256$\times$256 RGB SPIs & N/A & N/A & N/A & 11 categories by WMO (image-level label): cirrus, cirrostratus, cirrocumulus, altocumulus, altostratus, cumulus, cumulonimbus, nimbostratus, stratocumulus, stratus, contrail & N/A \\ \hline \noalign{\vskip 1mm}
FGCDR \cite{ye2019supervised} & 28638 4288$\times$2848$^*$ ASIs & N/A & N/A & 8-level segmentation (cumulus, stratocumulus, stratus, altostratus, altocumulus, cirrocumulus, cirrostratus, cirrus) labeled manually & 8 categories (pixel-level label): same as segmentation map & N/A \\ \hline \noalign{\vskip 1mm}
MGCD \cite{liu2020multi} & 8000 1024$\times$1024 RGB ASIs & N/A & N/A & N/A & 7 categories by WMO (image-level label): cumulus, altocumulus and cirrocumulus, cirrus and cirrostratus, clear sky, stratocumulus, stratus and altostratus, cumulonimbus and nimbostratus and mixed cloud) & N/A \\

GRSCD \cite{liu2020ground} & 8000 1024$\times$1024 RGB ASIs & N/A & N/A & N/A & 7 cloud categories by WMO (image-level label): cumulus, altocumulus and cirrocumulus, cirrus and cirrostratus, clear sky, stratocumulus and stratus and altostratus, cumulonimbus and nimbostratus, mixed cloud & N/A \\ \hline \noalign{\vskip 1mm} 
WMD \cite{krauz2020assessing} & 2044 3264$\times$4928 (raw), 1200$\times$800 (resized) RGB ASIs.  & N/A & N/A & 4-level segmentation (high-level clouds, low-level cumulus type clouds, rain clouds and clear sky) labeled manually & 4 categories by cloud height (pixel-level label): same as segmentation map & N/A \\  \hline \noalign{\vskip 1mm}
GCD \cite{liu2021ground} & 19000 512$\times$512 RGB SPIs & N/A & N/A & N/A & 7 categories by WMO (image-level label) cumulus, altocumulus and cirrocumulus, cirrus and cirrostratus, clear sky, stratocumulus, stratus and altostratus, cumulonimbus and nimbostratus, and mixed cloud & N/A \\ \hline \noalign{\vskip 1mm}
NIMS-KMA \cite{kim2021twenty} & 7402 2432$\times$2432 daytime and nighttime RGB ASIs & N/A & N/A & N/A & 10 categories based on cloud cover (image-level label) labeled manually  & N/A \\ \hline \noalign{\vskip 1mm}
Girasol \cite{Girasol2021} & 8200 60$\times$80 infra-red SPIs$^{**}$ & N/A & N/A & N/A & 4 categories: clear-sky, cumulus, stratus or nimbus cloud (image-level label) labeled manually  & N/A \\ \hline \noalign{\vskip 1mm}
PSA Fabel \cite{fabel2022applying} & 512$\times$512 RGB ASI: 770 labeled and 286477 unlabeled images  & N/A & N/A & 3-level segmentation (low-, middle- and high-layer clouds) manually labeled  & 3 categories (pixel-level label): same as segmentation map & N/A \\ \hline \noalign{\vskip 1mm}
NAO-CAS XJ \cite{li2022all} & 5000 370$\times$370 RGB ASIs & N/A & N/A & N/A & 4 categories based on cloud cover (image-level label): clear, outer, inner, covered & N/A \\ 
\end{longtable}
\end{ThreePartTable}
\end{landscape}

\section{Open-source datasets usage involving sky images}
\label{sec:appendixB}
The studies that use the sky image datasets involving the use of sky images are listed in Table \ref{tab:studies_use_sky_image_datasets} below.

\begin{longtable}{|l|l|>{\raggedright\arraybackslash}p{0.63\textwidth}|}
\caption{List of studies that use the sky image datasets involving the use of sky images} \label{tab:studies_use_sky_image_datasets} \\

\hline \textbf{Application} & \textbf{Dataset} & \textbf{Studies use the dataset involving using sky images}  \\ \hline 
\endfirsthead

\caption {List of studies that use the sky image datasets involving the use of sky images (continued)}\\
\hline \textbf{Application} & \textbf{Dataset} & \textbf{Studies use the dataset involving using sky images}  \\ \hline 
\endhead \\ \hline 
\endhead

%\multicolumn{3}{|r|}{{Continued on next page}} \\ \hline
%\endfoot

%\hline \hline
%\endlastfoot

\multirow{18}{*}{SF/CMP} & SRRL-BMS \cite{SRRL1981}  & \cite{zhu2016method,fengOpenSolarPromotingOpenness2019,wen2020deep,zhen2020deep,Feng2020,feng2020machine,chen2021solar,xiang2021sky,zhu2021cloud,Feng2022,zuo2022ten,zhang2022solar,dolatabadi2022deep,chen20223d} \\ \cline{2-3}
& SIRTA \cite{SIRTA2005} & \cite{palettaTemporallyConsistentImagebased2020,palettaBenchmarkingDeepLearning2021,Paletta2021eclipse,al2021improvement,insaf2021global,palettaSPINSimplifyingPolar2021,paletta2022omnivision} \\ \cline{2-3}
& ARM-STORMVEX \cite{ARM_STORMVEX2010}  & \cite{matsui2012evaluation} \\ \cline{2-3}
& ARM-COPS \cite{ARM_COPS2011}  & \cite{hanschmann2012evaluation}\\ \cline{2-3}
& ARM-HFE \cite{ARM_EAST_AIRC2011}  & \cite{liu2013cloud}\\ \cline{2-3}
& ARM-BAECC \cite{BAECCA2016} & \cite{ylivinkka2020clouds} \\ \cline{2-3}
& ARM-MAGIC \cite{ARM_MAGIC2016} & \cite{yang2019cloud} \\ \cline{2-3}
& ARM-GoAmazon \cite{ARM_GoAmazon2016}  & \cite{pennypacker2021exploring} \\ \cline{2-3}
& ARM-SGP \cite{ARM_SGP2016} & \cite{lim2019long} \\ \cline{2-3}
& Les dataset \cite{caldas2019very} & \cite{caldas2019very} \\ \cline{2-3}
& UCSD-Folsom \cite{UCSD2019}  & \cite{chu2014smart,chu2015real,pedro2015nearest,li2016quantitative,chu2017short,pedro2018assessment,yang2019ultra,wen2020deep,yang2020probabilistic,yang20213d,wang2021hybrid} \\ \cline{2-3}
& ARM-AWARE \cite{ARM_AWARE2020}  & \cite{liu2022novel}\\ \cline{2-3}
& BASS \cite{c2021feasibility} & \cite{c2021feasibility} \\ \cline{2-3}
& Girasol \cite{Girasol2021}  & \cite{terren_serranoDataProcessingShortTerm, ajith2021deep} \\ \cline{2-3}
& SIPM \cite{SIPM2021} & \cite{SIPM2021} \\ \cline{2-3}
& SIPPMIF \cite{UOW2021} & \cite{UOW2021} \\ \cline{2-3}
& Waggle \cite{park2021prediction} & \cite{park2021prediction} \\ \cline{2-3}
& SKIPP'D \cite{nie2022skippd} & \cite{Sun2018,sun2018convolutional,Sun2019,Sun2019dissertation,Venugopal2019,Venugopal2019thesis,Nie2020,Nie2021}\\ \hline

\multirow{16}{*}{CS} & SURFRAD \cite{SURFRAD2000} & \cite{long2006estimation,ten2016aerosol,calbo2017thin} \\ \cline{2-3}
& HYTA (Binary) \cite{li2011hybrid}  & \cite{li2011hybrid,dev2014systematic,dev2015multi,dev2016rough,rajini2018classification,andrade2019formation,dev2019multi,dianne2019deep,song2020efficient,shete2020tasselgan,pahurkarcloud,park2021prediction} \\ \cline{2-3}
& HYTA (Ternary) \cite{dev2015multi}  & \cite{dev2015multi,dev2019multi,dianne2019deep} \\ \cline{2-3}
& NCU \cite{cheng2017cloud} & \cite{cheng2017cloud} \\ \cline{2-3}
& SWIMSEG \cite{SWIMSEG2017}  & \cite{SWIMSEG2017,tulpan2017detection,dianne2019deep,dev2019cloudsegnet,funk2019passive,shi2020cloudu,song2020efficient,shete2020tasselgan,de2020unsupervised,shi2021cloudu,park2021prediction,roy2021towards,shirazi2021cloud,shi2022cloudraednet}\\ \cline{2-3}
& SWINSEG \cite{dev2017nighttime}  & \cite{dev2017nighttime,dev2019cloudsegnet,leveque2019subjective,shi2020cloudu,jain2021using,shi2021cloudu,shi2022cloudraednet}\\ \cline{2-3}
& SHWIMSEG \cite{SHWIMSEG2018}  & \cite{SHWIMSEG2018,rudrappa2020cloud} \\ \cline{2-3}
& NAO-CAS \cite{shi2019diurnal} & \cite{shi2019diurnal} \\ \cline{2-3}
& FGCDR \cite{ye2019supervised} & \cite{ye2019supervised,ye2022ccad} \\ \cline{2-3}
& SWINySEG \cite{dev2019cloudsegnet} & \cite{dev2019cloudsegnet,shi2020cloudu,shi2021cloudu,shi2022cloudraednet,makwana2022aclnet} \\ \cline{2-3}
& WSISEG \cite{xie2020segcloud} & \cite{xie2020segcloud} \\ \cline{2-3}
& WMD \cite{krauz2020assessing} & \cite{krauz2020assessing} \\ \cline{2-3}
& Girasol \cite{Girasol2021}  & \cite{terren2021comparative,terren2021explicit,terren2021segmentation,terren2021detection, terren_serranoDataProcessingShortTerm} \\ \cline{2-3}
& TCDD \cite{zhang2021ground} & \cite{zhang2021ground} \\ \cline{2-3}
& TLCDD \cite{zhang2022ground} & \cite{zhang2022ground} \\ \cline{2-3}
& PSA Fabel \cite{fabel2022applying} & \cite{fabel2022applying} \\ \cline{2-3}
& ACS WSI \cite{ye2022self} & \cite{ye2022self} \\ \hline

\multirow{13}{*}{CC} & \multirow{2}{*}{SWIMCAT \cite{SWIMCAT2015}}  & \cite{SWIMCAT2015,shi2017deep,wang2017measure,rajini2018classification,phung2018deep,wang2018ground,xu2018unsupervised,wang2018selection,luo2018ground,andrade2019formation,phung2019high,liu2019ground,zhang2020ensemble,liu2020multi,liu2020multimodal,wang2020clouda,hoang2020adaptive,hong2020comparative,manzo2021voting,tang2021improving} \\ \cline{2-3}
& TCIS \cite{li2016pixels}  & \cite{li2016pixels,gan2017cloud,yang2018analyzing,oikonomou2019local,wang2020clouda} \\ \cline{2-3}
& Zenithal \cite{luo2018ground} & \cite{luo2018ground} \\ \cline{2-3}
& CCSN \cite{CCSN2018} & \cite{CCSN2018,song2020efficient,zhang2020ensemble,manzo2021voting,roy2021towards,ambildhuke2021transfer,li2021novel,wang2021hacloudnet,zhang2022machine,tougaccar2022classification,sethy2022cloud} \\ \cline{2-3}
& FGCDR \cite{ye2019supervised} & \cite{ye2019supervised,ye2022ccad} \\ \cline{2-3}
& MGCD \cite{liu2020multi}  & \cite{liu2020multi,liu2020multimodal,manzo2021voting,zhu2022classification} \\ \cline{2-3}
& GRSCD \cite{liu2020ground} & \cite{liu2020ground} \\ \cline{2-3}
& WMD \cite{krauz2020assessing} & \cite{krauz2020assessing} \\ \cline{2-3}
& GCD \cite{liu2021ground} & \cite{liu2021ground} \\ \cline{2-3}
& NIMS-KMA \cite{kim2021twenty} & \cite{kim2021twenty} \\ \cline{2-3}
& Girasol \cite{Girasol2021}  & \cite{ajith2021deep,terren2021comparative,terren2021explicit,terren2021segmentation, terren_serranoDataProcessingShortTerm} \\ \cline{2-3}
& PSA Fabel \cite{fabel2022applying} & \cite{fabel2022applying} \\ \cline{2-3}
& NAO-CAS XJ \cite{li2022all} & \cite{li2022all} \\ \hline
\end{longtable}

\bibliographystyle{unsrtnat}
%\bibliographystyle{cas-model2-names}

\bibliography{refs}

\begin{thebibliography}{255}
\providecommand{\natexlab}[1]{#1}
\providecommand{\url}[1]{\texttt{#1}}
\expandafter\ifx\csname urlstyle\endcsname\relax
  \providecommand{\doi}[1]{doi: #1}\else
  \providecommand{\doi}{doi: \begingroup \urlstyle{rm}\Url}\fi

\bibitem[Gielen et~al.(2019)Gielen, Boshell, Saygin, Bazilian, Wagner, and
  Gorini]{gielen2019role}
Dolf Gielen, Francisco Boshell, Deger Saygin, Morgan~D Bazilian, Nicholas
  Wagner, and Ricardo Gorini.
\newblock The role of renewable energy in the global energy transformation.
\newblock \emph{Energy Strategy Reviews}, 24:\penalty0 38--50, 2019.

\bibitem[Kabir et~al.(2018)Kabir, Kumar, Kumar, Adelodun, and
  Kim]{kabir2018solar}
Ehsanul Kabir, Pawan Kumar, Sandeep Kumar, Adedeji~A Adelodun, and Ki-Hyun Kim.
\newblock Solar energy: Potential and future prospects.
\newblock \emph{Renewable and Sustainable Energy Reviews}, 82:\penalty0
  894--900, 2018.

\bibitem[Barbieri et~al.(2017)Barbieri, Rajakaruna, and
  Ghosh]{barbieri2017very}
Florian Barbieri, Sumedha Rajakaruna, and Arindam Ghosh.
\newblock Very short-term photovoltaic power forecasting with cloud modeling: A
  review.
\newblock \emph{Renewable and Sustainable Energy Reviews}, 75:\penalty0
  242--263, 2017.

\bibitem[Sun et~al.(2019)Sun, Venugopal, and Brandt]{Sun2019}
Yuchi Sun, Vignesh Venugopal, and Adam~R Brandt.
\newblock {Short-term solar power forecast with deep learning: Exploring
  optimal input and output configuration}.
\newblock \emph{Solar Energy}, 188:\penalty0 730--741, aug 2019.
\newblock \doi{10.1016/j.solener.2019.06.041}.
\newblock URL
  \url{https://linkinghub.elsevier.com/retrieve/pii/S0038092X19306164}.

\bibitem[Sun(2019)]{Sun2019dissertation}
Yuchi Sun.
\newblock \emph{{Short-term Solar Forecast Using Convolutional Neural Networks
  with Sky Images}}.
\newblock Phd, Stanford University, 2019.
\newblock URL \url{http://purl.stanford.edu/fm704js1179}.

\bibitem[Ahmed et~al.(2020)Ahmed, Sreeram, Mishra, and Arif]{ahmed2020review}
Razin Ahmed, V~Sreeram, Y~Mishra, and MD~Arif.
\newblock A review and evaluation of the state-of-the-art in pv solar power
  forecasting: Techniques and optimization.
\newblock \emph{Renewable and Sustainable Energy Reviews}, 124:\penalty0
  109792, 2020.

\bibitem[Ren et~al.(2015)Ren, Suganthan, and Srikanth]{ren2015ensemble}
Ye~Ren, PN~Suganthan, and N~Srikanth.
\newblock Ensemble methods for wind and solar power forecasting—a
  state-of-the-art review.
\newblock \emph{Renewable and Sustainable Energy Reviews}, 50:\penalty0 82--91,
  2015.

\bibitem[Das et~al.(2018)Das, Tey, Seyedmahmoudian, Mekhilef, Idris,
  Van~Deventer, Horan, and Stojcevski]{das2018forecasting}
Utpal~Kumar Das, Kok~Soon Tey, Mehdi Seyedmahmoudian, Saad Mekhilef, Moh
  Yamani~Idna Idris, Willem Van~Deventer, Bend Horan, and Alex Stojcevski.
\newblock Forecasting of photovoltaic power generation and model optimization:
  A review.
\newblock \emph{Renewable and Sustainable Energy Reviews}, 81:\penalty0
  912--928, 2018.

\bibitem[Behera et~al.(2018)Behera, Majumder, and Nayak]{behera2018solar}
Manoja~Kumar Behera, Irani Majumder, and Niranjan Nayak.
\newblock Solar photovoltaic power forecasting using optimized modified extreme
  learning machine technique.
\newblock \emph{Engineering Science and Technology, an International Journal},
  21\penalty0 (3):\penalty0 428--438, 2018.

\bibitem[Van~der Meer et~al.(2018)Van~der Meer, Wid{\'e}n, and
  Munkhammar]{van2018review}
Dennis~W Van~der Meer, Joakim Wid{\'e}n, and Joakim Munkhammar.
\newblock Review on probabilistic forecasting of photovoltaic power production
  and electricity consumption.
\newblock \emph{Renewable and Sustainable Energy Reviews}, 81:\penalty0
  1484--1512, 2018.

\bibitem[Chow et~al.(2011)Chow, Urquhart, Lave, Dominguez, Kleissl, Shields,
  and Washom]{chow2011intra}
Chi~Wai Chow, Bryan Urquhart, Matthew Lave, Anthony Dominguez, Jan Kleissl,
  Janet Shields, and Byron Washom.
\newblock Intra-hour forecasting with a total sky imager at the uc san diego
  solar energy testbed.
\newblock \emph{Solar Energy}, 85\penalty0 (11):\penalty0 2881--2893, 2011.

\bibitem[Marquez and Coimbra(2013)]{marquez2013intra}
Ricardo Marquez and Carlos~FM Coimbra.
\newblock Intra-hour dni forecasting based on cloud tracking image analysis.
\newblock \emph{Solar Energy}, 91:\penalty0 327--336, 2013.

\bibitem[{Quesada-Ruiz} et~al.(2014){Quesada-Ruiz}, Chu, {Tovar-Pescador},
  Pedro, and Coimbra]{quesada-ruizCloudtrackingMethodologyIntrahour2014a}
S.~{Quesada-Ruiz}, Y.~Chu, J.~{Tovar-Pescador}, H.~T.~C. Pedro, and C.~F.~M.
  Coimbra.
\newblock Cloud-tracking methodology for intra-hour {{DNI}} forecasting.
\newblock \emph{Solar Energy}, 102:\penalty0 267--275, April 2014.
\newblock ISSN 0038-092X.
\newblock \doi{10.1016/j.solener.2014.01.030}.

\bibitem[Fu and Cheng(2013)]{fu2013predicting}
Chia-Lin Fu and Hsu-Yung Cheng.
\newblock Predicting solar irradiance with all-sky image features via
  regression.
\newblock \emph{Solar Energy}, 97:\penalty0 537--550, 2013.

\bibitem[Chu et~al.(2013)Chu, Pedro, and Coimbra]{chuHybridIntrahourDNI2013a}
Yinghao Chu, Hugo T.~C. Pedro, and Carlos F.~M. Coimbra.
\newblock Hybrid intra-hour {{DNI}} forecasts with sky image processing
  enhanced by stochastic learning.
\newblock \emph{Solar Energy}, 98:\penalty0 592--603, December 2013.
\newblock ISSN 0038-092X.
\newblock \doi{10.1016/j.solener.2013.10.020}.

\bibitem[Chu et~al.(2015{\natexlab{a}})Chu, Li, Pedro, and
  Coimbra]{Chu2015realtime}
Yinghao Chu, Mengying Li, Hugo T~C Pedro, and Carlos F~M Coimbra.
\newblock {Real-time prediction intervals for intra-hour DNI forecasts}.
\newblock \emph{Renewable Energy}, 83:\penalty0 234--244, 2015{\natexlab{a}}.
\newblock ISSN 18790682.
\newblock \doi{10.1016/j.renene.2015.04.022}.
\newblock URL \url{http://dx.doi.org/10.1016/j.renene.2015.04.022}.

\bibitem[Chu et~al.(2015{\natexlab{b}})Chu, Urquhart, Gohari, Pedro, Kleissl,
  and Coimbra]{Chu2015reforcast}
Yinghao Chu, Bryan Urquhart, Seyyed~M.I. Gohari, Hugo~T.C. Pedro, Jan Kleissl,
  and Carlos~F.M. Coimbra.
\newblock {Short-term reforecasting of power output from a 48 MWe solar PV
  plant}.
\newblock \emph{Solar Energy}, 112:\penalty0 68--77, feb 2015{\natexlab{b}}.
\newblock ISSN 0038-092X.
\newblock \doi{10.1016/J.SOLENER.2014.11.017}.
\newblock URL
  \url{https://www.sciencedirect.com/science/article/pii/S0038092X14005611}.

\bibitem[Pedro et~al.(2019{\natexlab{a}})Pedro, Coimbra, and
  Lauret]{pedroAdaptiveImageFeatures2019}
Hugo T.~C. Pedro, Carlos F.~M. Coimbra, and Philippe Lauret.
\newblock Adaptive image features for intra-hour solar forecasts.
\newblock \emph{Journal of Renewable and Sustainable Energy}, 11\penalty0
  (3):\penalty0 036101, May 2019{\natexlab{a}}.
\newblock \doi{10.1063/1.5091952}.

\bibitem[Peng et~al.(2015)Peng, Yu, Huang, Heiser, Yoo, and
  Kalb]{peng3DCloudDetection2015}
Zhenzhou Peng, Dantong Yu, Dong Huang, John Heiser, Shinjae Yoo, and Paul Kalb.
\newblock {{3D}} cloud detection and tracking system for solar forecast using
  multiple sky imagers.
\newblock \emph{Solar Energy}, 118:\penalty0 496--519, August 2015.
\newblock ISSN 0038092X.
\newblock \doi{10.1016/j.solener.2015.05.037}.

\bibitem[Blanc et~al.(2017)Blanc, Massip, Kazantzidis, Tzoumanikas, Kuhn,
  Wilbert, Sch{\"u}ler, and Prahl]{blancShorttermForecastingHigh2017a}
Philippe Blanc, Pierre Massip, Andreas Kazantzidis, Panagiotis Tzoumanikas,
  Pascal Kuhn, Stefan Wilbert, David Sch{\"u}ler, and Christoph Prahl.
\newblock Short-term forecasting of high resolution local {{DNI}} maps with
  multiple fish-eye cameras in stereoscopic mode.
\newblock \emph{AIP Conference Proceedings}, 1850\penalty0 (1):\penalty0
  140004, June 2017.
\newblock ISSN 0094-243X.
\newblock \doi{10.1063/1.4984512}.

\bibitem[Kuhn et~al.(2018)Kuhn, Nouri, Wilbert, Prahl, Kozonek, Schmidt,
  Yasser, Ramirez, Zarzalejo, Meyer, Vuilleumier, Heinemann, Blanc, and
  Pitz-Paal]{kuhnValidationAllskyImager2018a}
Pascal Kuhn, Bijan Nouri, Stefan Wilbert, Christoph Prahl, Nora Kozonek, Thomas
  Schmidt, Zeyad Yasser, Lourdes Ramirez, Luis Zarzalejo, Angela Meyer, Laurent
  Vuilleumier, Detlev Heinemann, Philippe Blanc, and Robert Pitz-Paal.
\newblock Validation of an all-sky imager\textendash based nowcasting system
  for industrial {{PV}} plants.
\newblock \emph{Progress in Photovoltaics: Research and Applications},
  26\penalty0 (8):\penalty0 608--621, 2018.
\newblock ISSN 1099-159X.
\newblock \doi{10.1002/pip.2968}.

\bibitem[Kuhn et~al.(2019)Kuhn, Nouri, Wilbert, Hanrieder, Prahl, Ramirez,
  Zarzalejo, Schmidt, Schmidt, Yasser, Heinemann, Tzoumanikas, Kazantzidis,
  Kleissl, Blanc, and Pitz-Paal]{Kuhn_2019}
P.~Kuhn, B.~Nouri, S.~Wilbert, N.~Hanrieder, C.~Prahl, L.~Ramirez,
  L.~Zarzalejo, T.~Schmidt, T.~Schmidt, Z.~Yasser, D.~Heinemann,
  P.~Tzoumanikas, A.~Kazantzidis, J.~Kleissl, P.~Blanc, and R.~Pitz-Paal.
\newblock Determination of the optimal camera distance for cloud height
  measurements with two all-sky imagers.
\newblock \emph{Solar Energy}, 179:\penalty0 74--88, feb 2019.
\newblock \doi{10.1016/j.solener.2018.12.038}.
\newblock URL \url{https://doi.org/10.1016\%2Fj.solener.2018.12.038}.

\bibitem[Blum et~al.(2021)Blum, Nouri, Wilbert, Schmidt, L\"{u}nsdorf,
  St\"{u}hrenberg, Heinemann, Kazantzidis, and Pitz-Paal]{Blum_2021}
Niklas~Benedikt Blum, Bijan Nouri, Stefan Wilbert, Thomas Schmidt, Ontje
  L\"{u}nsdorf, Jonas St\"{u}hrenberg, Detlev Heinemann, Andreas Kazantzidis,
  and Robert Pitz-Paal.
\newblock Cloud height measurement by a network of all-sky imagers.
\newblock \emph{Atmospheric Measurement Techniques}, 14\penalty0 (7):\penalty0
  5199--5224, jul 2021.
\newblock \doi{10.5194/amt-14-5199-2021}.
\newblock URL \url{https://doi.org/10.5194\%2Famt-14-5199-2021}.

\bibitem[Paletta et~al.(2021)Paletta, Arbod, and
  Lasenby]{palettaBenchmarkingDeepLearning2021}
Quentin Paletta, Guillaume Arbod, and Joan Lasenby.
\newblock Benchmarking of deep learning irradiance forecasting models from sky
  images \textendash{} {{An}} in-depth analysis.
\newblock \emph{Solar Energy}, 224:\penalty0 855--867, August 2021.
\newblock ISSN 0038-092X.
\newblock \doi{10.1016/j.solener.2021.05.056}.
\newblock URL
  \url{https://linkinghub.elsevier.com/retrieve/pii/S0038092X21004266}.

\bibitem[Venugopal et~al.(2019)Venugopal, Sun, and Brandt]{Venugopal2019}
Vignesh Venugopal, Yuchi Sun, and Adam~R. Brandt.
\newblock {Short-term solar PV forecasting using computer vision: The search
  for optimal CNN architectures for incorporating sky images and PV generation
  history}.
\newblock \emph{Journal of Renewable and Sustainable Energy}, 11\penalty0
  (6):\penalty0 066102, nov 2019.
\newblock ISSN 1941-7012.
\newblock \doi{10.1063/1.5122796}.
\newblock URL \url{http://aip.scitation.org/doi/10.1063/1.5122796}.

\bibitem[Feng and Zhang(2020)]{Feng2020}
Cong Feng and Jie Zhang.
\newblock {SolarNet: A sky image-based deep convolutional neural network for
  intra-hour solar forecasting}.
\newblock \emph{Solar Energy}, 204\penalty0 (April):\penalty0 71--78, 2020.
\newblock \doi{10.1016/j.solener.2020.03.083}.
\newblock URL \url{https://doi.org/10.1016/j.solener.2020.03.083}.

\bibitem[Paletta and
  Lasenby(2020{\natexlab{a}})]{palettaConvolutionalNeuralNetworks2020}
Quentin Paletta and Joan Lasenby.
\newblock Convolutional {{Neural Networks Applied}} to {{Sky Images}} for
  {{Short}}-{{Term Solar Irradiance Forecasting}}.
\newblock In \emph{{{EU PVSEC}}}, pages 1834 -- 1837, 2020{\natexlab{a}}.
\newblock ISBN 3-936338-73-6.
\newblock \doi{10.4229/EUPVSEC20202020-6BV.5.15}.
\newblock URL
  \url{https://www.eupvsec-proceedings.com/proceedings?paper=49346}.

\bibitem[Feng et~al.(2022)Feng, Zhang, Zhang, and Hodge]{Feng2022}
Cong Feng, Jie Zhang, Wenqi Zhang, and Bri~Mathias Hodge.
\newblock {Convolutional neural networks for intra-hour solar forecasting based
  on sky image sequences}.
\newblock \emph{Applied Energy}, 310:\penalty0 118438, mar 2022.
\newblock \doi{10.1016/J.APENERGY.2021.118438}.

\bibitem[Zhang et~al.(2018{\natexlab{a}})Zhang, Verschae, Nobuhara, and
  Lalonde]{Zhang2018}
Jinsong Zhang, Rodrigo Verschae, Shohei Nobuhara, and Jean~Fran{\c{c}}ois
  Lalonde.
\newblock {Deep photovoltaic nowcasting}.
\newblock \emph{Solar Energy}, 176\penalty0 (September):\penalty0 267--276,
  2018{\natexlab{a}}.
\newblock \doi{10.1016/j.solener.2018.10.024}.
\newblock URL \url{https://doi.org/10.1016/j.solener.2018.10.024}.

\bibitem[Paletta et~al.(2022{\natexlab{a}})Paletta, Hu, Arbod, and
  Lasenby]{Paletta2021eclipse}
Quentin Paletta, Anthony Hu, Guillaume Arbod, and Joan Lasenby.
\newblock {{ECLIPSE}}: {{Envisioning CLoud Induced Perturbations}} in {{Solar
  Energy}}.
\newblock \emph{Applied Energy}, 326:\penalty0 119924, November
  2022{\natexlab{a}}.
\newblock ISSN 0306-2619.
\newblock \doi{10.1016/j.apenergy.2022.119924}.

\bibitem[Dev et~al.(2017{\natexlab{a}})Dev, Lee, and Winkler]{SWIMSEG2017}
Soumyabrata Dev, Yee~Hui Lee, and Stefan Winkler.
\newblock {Color-Based Segmentation of Sky/Cloud Images From Ground-Based
  Cameras}.
\newblock \emph{IEEE Journal of Selected Topics in Applied Earth Observations
  and Remote Sensing}, 10\penalty0 (1):\penalty0 231--242, 2017{\natexlab{a}}.
\newblock \doi{10.1109/JSTARS.2016.2558474}.

\bibitem[Heinle et~al.(2010)Heinle, Macke, and Srivastav]{heinle2010automatic}
Anna Heinle, Andreas Macke, and Anand Srivastav.
\newblock Automatic cloud classification of whole sky images.
\newblock \emph{Atmospheric Measurement Techniques}, 3\penalty0 (3):\penalty0
  557--567, 2010.

\bibitem[Liu et~al.(2014)Liu, Zhang, Zhang, Wang, and Xiao]{liu2014automatic}
Shuang Liu, Linbo Zhang, Zhong Zhang, Chunheng Wang, and Baihua Xiao.
\newblock Automatic cloud detection for all-sky images using superpixel
  segmentation.
\newblock \emph{IEEE Geoscience and Remote Sensing Letters}, 12\penalty0
  (2):\penalty0 354--358, 2014.

\bibitem[Long et~al.(2006{\natexlab{a}})Long, Sabburg, Calb{\'o}, and
  Pag{\`e}s]{long2006retrieving}
Charles~N Long, Jeff~M Sabburg, Josep Calb{\'o}, and David Pag{\`e}s.
\newblock Retrieving cloud characteristics from ground-based daytime color
  all-sky images.
\newblock \emph{Journal of Atmospheric and Oceanic Technology}, 23\penalty0
  (5):\penalty0 633--652, 2006{\natexlab{a}}.

\bibitem[Ghonima et~al.(2012)Ghonima, Urquhart, Chow, Shields, Cazorla, and
  Kleissl]{ghonima2012method}
MS~Ghonima, B~Urquhart, CW~Chow, JE~Shields, Alberto Cazorla, and Jan Kleissl.
\newblock A method for cloud detection and opacity classification based on
  ground based sky imagery.
\newblock \emph{Atmospheric Measurement Techniques}, 5\penalty0 (11):\penalty0
  2881--2892, 2012.

\bibitem[Li et~al.(2011{\natexlab{a}})Li, Lu, and Yang]{li2011hybrid}
Qingyong Li, Weitao Lu, and Jun Yang.
\newblock {A hybrid thresholding algorithm for cloud detection on ground-based
  color images}.
\newblock \emph{Journal of atmospheric and oceanic technology}, 28\penalty0
  (10):\penalty0 1286--1296, 2011{\natexlab{a}}.

\bibitem[Chauvin et~al.(2015)Chauvin, Nou, Thil, Traore, and
  Grieu]{chauvin2015cloud}
R{\'e}mi Chauvin, Julien Nou, St{\'e}phane Thil, Adama Traore, and St{\'e}phane
  Grieu.
\newblock Cloud detection methodology based on a sky-imaging system.
\newblock \emph{Energy Procedia}, 69:\penalty0 1970--1980, 2015.

\bibitem[Nie et~al.(2020)Nie, Sun, Chen, Orsini, and Brandt]{Nie2020}
Yuhao Nie, Yuchi Sun, Yuanlei Chen, Rachel Orsini, and Adam Brandt.
\newblock {PV power output prediction from sky images using convolutional
  neural network: The comparison of sky-condition-specific sub-models and an
  end-to-end model}.
\newblock \emph{Journal of Renewable and Sustainable Energy}, 12\penalty0
  (4):\penalty0 046101, jul 2020.
\newblock \doi{10.1063/5.0014016}.
\newblock URL \url{http://aip.scitation.org/doi/10.1063/5.0014016}.

\bibitem[Souza-Echer et~al.(2006)Souza-Echer, Pereira, Bins, and
  Andrade]{souza2006simple}
Mariza~Pereira Souza-Echer, Enio~Bueno Pereira, LS~Bins, and MAR Andrade.
\newblock A simple method for the assessment of the cloud cover state in
  high-latitude regions by a ground-based digital camera.
\newblock \emph{Journal of Atmospheric and Oceanic Technology}, 23\penalty0
  (3):\penalty0 437--447, 2006.

\bibitem[Neto et~al.(2010)Neto, von Wangenheim, Pereira, and
  Comunello]{neto2010use}
Sylvio Luiz~Mantelli Neto, Aldo von Wangenheim, Enio~Bueno Pereira, and Eros
  Comunello.
\newblock The use of euclidean geometric distance on rgb color space for the
  classification of sky and cloud patterns.
\newblock \emph{Journal of Atmospheric and Oceanic Technology}, 27\penalty0
  (9):\penalty0 1504--1517, 2010.

\bibitem[Dev et~al.(2019{\natexlab{a}})Dev, Nautiyal, Lee, and
  Winkler]{dev2019cloudsegnet}
Soumyabrata Dev, Atul Nautiyal, Yee~Hui Lee, and Stefan Winkler.
\newblock Cloudsegnet: A deep network for nychthemeron cloud image
  segmentation.
\newblock \emph{IEEE Geoscience and Remote Sensing Letters}, 16\penalty0
  (12):\penalty0 1814--1818, 2019{\natexlab{a}}.

\bibitem[Xie et~al.(2020)Xie, Liu, Yang, Chen, Wang, Wang, Xia, Liu, Wang, and
  Zhang]{xie2020segcloud}
Wanyi Xie, Dong Liu, Ming Yang, Shaoqing Chen, Benge Wang, Zhenzhu Wang,
  Yingwei Xia, Yong Liu, Yiren Wang, and Chaofang Zhang.
\newblock {SegCloud: a novel cloud image segmentation model using a deep
  convolutional neural network for ground-based all-sky-view camera
  observation}.
\newblock \emph{Atmospheric Measurement Techniques}, 13\penalty0 (4):\penalty0
  1953--1961, 2020.

\bibitem[Zhang et~al.(2021)Zhang, Yang, Liu, Xiao, and Cao]{zhang2021ground}
Zhong Zhang, Shuzhen Yang, Shuang Liu, Baihua Xiao, and Xiaozhong Cao.
\newblock Ground-based cloud detection using multiscale attention convolutional
  neural network.
\newblock \emph{IEEE Geoscience and Remote Sensing Letters}, 19:\penalty0 1--5,
  2021.

\bibitem[Fabel et~al.(2022)Fabel, Nouri, Wilbert, Blum, Triebel, Hasenbalg,
  Kuhn, Zarzalejo, and Pitz-Paal]{fabel2022applying}
Yann Fabel, Bijan Nouri, Stefan Wilbert, Niklas Blum, Rudolph Triebel, Marcel
  Hasenbalg, Pascal Kuhn, Luis~F Zarzalejo, and Robert Pitz-Paal.
\newblock Applying self-supervised learning for semantic cloud segmentation of
  all-sky images.
\newblock \emph{Atmospheric Measurement Techniques}, 15\penalty0 (3):\penalty0
  797--809, 2022.

\bibitem[Hasenbalg et~al.(2020)Hasenbalg, Kuhn, Wilbert, Nouri, and
  Kazantzidis]{hasenbalg2020benchmarking}
Marcel Hasenbalg, P~Kuhn, Stefan Wilbert, Bijan Nouri, and A~Kazantzidis.
\newblock Benchmarking of six cloud segmentation algorithms for ground-based
  all-sky imagers.
\newblock \emph{Solar Energy}, 201:\penalty0 596--614, 2020.

\bibitem[Zhuo et~al.(2014)Zhuo, Cao, and Xiao]{zhuo2014cloud}
Wen Zhuo, Zhiguo Cao, and Yang Xiao.
\newblock Cloud classification of ground-based images using texture--structure
  features.
\newblock \emph{Journal of Atmospheric and Oceanic Technology}, 31\penalty0
  (1):\penalty0 79--92, 2014.

\bibitem[Zhang et~al.(2018{\natexlab{b}})Zhang, Liu, Zhang, and Song]{CCSN2018}
Jinglin Zhang, Pu~Liu, Feng Zhang, and Qianqian Song.
\newblock {CloudNet: Ground-Based Cloud Classification With Deep Convolutional
  Neural Network}.
\newblock \emph{Geophysical Research Letters}, 45\penalty0 (16):\penalty0
  8665--8672, 2018{\natexlab{b}}.
\newblock \doi{https://doi.org/10.1029/2018GL077787}.
\newblock URL
  \url{https://agupubs.onlinelibrary.wiley.com/doi/abs/10.1029/2018GL077787}.

\bibitem[Ye et~al.(2019)Ye, Cao, Xiao, and Yang]{ye2019supervised}
Liang Ye, Zhiguo Cao, Yang Xiao, and Zhibiao Yang.
\newblock Supervised fine-grained cloud detection and recognition in whole-sky
  images.
\newblock \emph{IEEE Transactions on Geoscience and Remote Sensing},
  57\penalty0 (10):\penalty0 7972--7985, 2019.

\bibitem[Dev et~al.(2015{\natexlab{a}})Dev, Lee, and Winkler]{SWIMCAT2015}
Soumyabrata Dev, Yee~Hui Lee, and Stefan Winkler.
\newblock Categorization of cloud image patches using an improved texton-based
  approach.
\newblock In \emph{2015 IEEE International Conference on Image Processing
  (ICIP)}, pages 422--426, 2015{\natexlab{a}}.
\newblock \doi{10.1109/ICIP.2015.7350833}.

\bibitem[Li et~al.(2016{\natexlab{a}})Li, Zhang, Lu, Yang, Ma, and
  Yao]{li2016pixels}
Qingyong Li, Zhen Zhang, Weitao Lu, Jun Yang, Ying Ma, and Wen Yao.
\newblock From pixels to patches: a cloud classification method based on a bag
  of micro-structures.
\newblock \emph{Atmospheric Measurement Techniques}, 9\penalty0 (2):\penalty0
  753--764, 2016{\natexlab{a}}.

\bibitem[Luo et~al.(2018)Luo, Zhou, Meng, Li, and Li]{luo2018ground}
Qixiang Luo, Zeming Zhou, Yong Meng, Qian Li, and Miaoying Li.
\newblock Ground-based cloud-type recognition using manifold kernel sparse
  coding and dictionary learning.
\newblock \emph{Advances in Meteorology}, 2018, 2018.

\bibitem[Ye et~al.(2017)Ye, Cao, and Xiao]{ye2017deepcloud}
Liang Ye, Zhiguo Cao, and Yang Xiao.
\newblock Deepcloud: Ground-based cloud image categorization using deep
  convolutional features.
\newblock \emph{IEEE Transactions on Geoscience and Remote Sensing},
  55\penalty0 (10):\penalty0 5729--5740, 2017.

\bibitem[Paletta et~al.(2022{\natexlab{b}})Paletta, Arbod, and
  Lasenby]{paletta2022omnivision}
Quentin Paletta, Guillaume Arbod, and Joan Lasenby.
\newblock Omnivision forecasting: combining satellite observations with sky
  images for improved intra-hour solar energy predictions.
\newblock \emph{arXiv preprint arXiv:2206.03207}, 2022{\natexlab{b}}.

\bibitem[Willert and Gharib(1991)]{willert1991digital}
Christian~E Willert and Morteza Gharib.
\newblock Digital particle image velocimetry.
\newblock \emph{Experiments in fluids}, 10\penalty0 (4):\penalty0 181--193,
  1991.

\bibitem[Beauchemin and Barron(1995)]{beauchemin1995computation}
Steven~S. Beauchemin and John~L. Barron.
\newblock The computation of optical flow.
\newblock \emph{ACM computing surveys (CSUR)}, 27\penalty0 (3):\penalty0
  433--466, 1995.

\bibitem[Huang et~al.(2013)Huang, Xu, Peng, Yoo, Yu, Huang, and Qin]{Huang2013}
Hao Huang, Jin Xu, Zhenzhou Peng, Shinjae Yoo, Dantong Yu, Dong Huang, and Hong
  Qin.
\newblock Cloud motion estimation for short term solar irradiation prediction.
\newblock In \emph{2013 IEEE International Conference on Smart Grid
  Communications (SmartGridComm)}, pages 696--701, 2013.
\newblock \doi{10.1109/SmartGridComm.2013.6688040}.

\bibitem[Dev et~al.(2016{\natexlab{a}})Dev, Savoy, Lee, and
  Winkler]{dev2016short}
Soumyabrata Dev, Florian~M Savoy, Yee~Hui Lee, and Stefan Winkler.
\newblock Short-term prediction of localized cloud motion using ground-based
  sky imagers.
\newblock In \emph{2016 IEEE Region 10 Conference (TENCON)}, pages 2563--2566.
  IEEE, 2016{\natexlab{a}}.

\bibitem[Andrianakos et~al.(2019)Andrianakos, Tsourounis, Oikonomou,
  Kastaniotis, Economou, and Kazantzidis]{andrianakos2019sky}
George Andrianakos, Dimitrios Tsourounis, Spiros Oikonomou, Dimitris
  Kastaniotis, George Economou, and Andreas Kazantzidis.
\newblock Sky image forecasting with generative adversarial networks for cloud
  coverage prediction.
\newblock In \emph{2019 10th International Conference on Information,
  Intelligence, Systems and Applications (IISA)}, pages 1--7. IEEE, 2019.

\bibitem[{Le Guen} and Thome(2020)]{LeGuen2020_solar}
Vincent {Le Guen} and Nicolas Thome.
\newblock {A deep physical model for solar irradiance forecasting with fisheye
  images}.
\newblock \emph{IEEE Computer Society Conference on Computer Vision and Pattern
  Recognition Workshops}, 2020-June:\penalty0 2685--2688, 2020.
\newblock ISSN 21607516.
\newblock \doi{10.1109/CVPRW50498.2020.00323}.

\bibitem[Inman et~al.(2013)Inman, Pedro, and Coimbra]{inman2013solar}
Rich~H Inman, Hugo~TC Pedro, and Carlos~FM Coimbra.
\newblock Solar forecasting methods for renewable energy integration.
\newblock \emph{Progress in energy and combustion science}, 39\penalty0
  (6):\penalty0 535--576, 2013.

\bibitem[Diagne et~al.(2013)Diagne, David, Lauret, Boland, and
  Schmutz]{diagne2013review}
Maimouna Diagne, Mathieu David, Philippe Lauret, John Boland, and Nicolas
  Schmutz.
\newblock Review of solar irradiance forecasting methods and a proposition for
  small-scale insular grids.
\newblock \emph{Renewable and Sustainable Energy Reviews}, 27:\penalty0 65--76,
  2013.

\bibitem[Antonanzas et~al.(2016)Antonanzas, Osorio, Escobar, Urraca,
  Martinez-de Pison, and Antonanzas-Torres]{antonanzas2016review}
Javier Antonanzas, Natalia Osorio, Rodrigo Escobar, Ruben Urraca, Francisco~J
  Martinez-de Pison, and Fernando Antonanzas-Torres.
\newblock Review of photovoltaic power forecasting.
\newblock \emph{Solar energy}, 136:\penalty0 78--111, 2016.

\bibitem[Voyant et~al.(2017)Voyant, Notton, Kalogirou, Nivet, Paoli, Motte, and
  Fouilloy]{voyant2017machine}
Cyril Voyant, Gilles Notton, Soteris Kalogirou, Marie-Laure Nivet, Christophe
  Paoli, Fabrice Motte, and Alexis Fouilloy.
\newblock Machine learning methods for solar radiation forecasting: A review.
\newblock \emph{Renewable Energy}, 105:\penalty0 569--582, 2017.

\bibitem[Sobri et~al.(2018)Sobri, Koohi-Kamali, and Rahim]{sobri2018solar}
Sobrina Sobri, Sam Koohi-Kamali, and Nasrudin~Abd Rahim.
\newblock Solar photovoltaic generation forecasting methods: A review.
\newblock \emph{Energy conversion and management}, 156:\penalty0 459--497,
  2018.

\bibitem[Yang et~al.(2018{\natexlab{a}})Yang, Kleissl, Gueymard, Pedro, and
  Coimbra]{yang2018history}
Dazhi Yang, Jan Kleissl, Christian~A Gueymard, Hugo~TC Pedro, and Carlos~FM
  Coimbra.
\newblock History and trends in solar irradiance and pv power forecasting: A
  preliminary assessment and review using text mining.
\newblock \emph{Solar Energy}, 168:\penalty0 60--101, 2018{\natexlab{a}}.

\bibitem[Kumar et~al.(2020)Kumar, Yagli, Kashyap, and
  Srinivasan]{kumar2020solar}
Dhivya~Sampath Kumar, Gokhan~Mert Yagli, Monika Kashyap, and Dipti Srinivasan.
\newblock Solar irradiance resource and forecasting: a comprehensive review.
\newblock \emph{IET Renewable Power Generation}, 14\penalty0 (10):\penalty0
  1641--1656, 2020.

\bibitem[Li and Zhang(2020)]{li2020review}
Binghui Li and Jie Zhang.
\newblock A review on the integration of probabilistic solar forecasting in
  power systems.
\newblock \emph{Solar Energy}, 210:\penalty0 68--86, 2020.

\bibitem[Guermoui et~al.(2020)Guermoui, Melgani, Gairaa, and
  Mekhalfi]{guermoui2020comprehensive}
Mawloud Guermoui, Farid Melgani, Kacem Gairaa, and Mohamed~Lamine Mekhalfi.
\newblock A comprehensive review of hybrid models for solar radiation
  forecasting.
\newblock \emph{Journal of Cleaner Production}, 258:\penalty0 120357, 2020.

\bibitem[Wang et~al.(2020{\natexlab{a}})Wang, Liu, Zhou, Li, Cao, Voropai, and
  Barakhtenko]{wang2020taxonomy}
Huaizhi Wang, Yangyang Liu, Bin Zhou, Canbing Li, Guangzhong Cao, Nikolai
  Voropai, and Evgeny Barakhtenko.
\newblock Taxonomy research of artificial intelligence for deterministic solar
  power forecasting.
\newblock \emph{Energy Conversion and Management}, 214:\penalty0 112909,
  2020{\natexlab{a}}.

\bibitem[Alkhayat and Mehmood(2021)]{alkhayat2021review}
Ghadah Alkhayat and Rashid Mehmood.
\newblock A review and taxonomy of wind and solar energy forecasting methods
  based on deep learning.
\newblock \emph{Energy and AI}, 4:\penalty0 100060, 2021.

\bibitem[Sharma and Elmenreich(2021)]{sharma2021review}
Ekanki Sharma and Wilfried Elmenreich.
\newblock A review on physical and data-driven based nowcasting methods using
  sky images.
\newblock In \emph{Future of Information and Communication Conference}, pages
  352--370. Springer, 2021.

\bibitem[Kumari and Toshniwal(2021)]{kumari2021deep}
Pratima Kumari and Durga Toshniwal.
\newblock Deep learning models for solar irradiance forecasting: A
  comprehensive review.
\newblock \emph{Journal of Cleaner Production}, 318:\penalty0 128566, 2021.

\bibitem[Chu et~al.(2021)Chu, Li, Coimbra, Feng, and Wang]{chu2021intra}
Yinghao Chu, Mengying Li, Carlos~FM Coimbra, Daquan Feng, and Huaizhi Wang.
\newblock Intra-hour irradiance forecasting techniques for solar power
  integration: A review.
\newblock \emph{Iscience}, 24\penalty0 (10):\penalty0 103136, 2021.

\bibitem[Lin et~al.(2022)Lin, Zhang, and Wang]{lin2022recent}
Fan Lin, Yao Zhang, and Jianxue Wang.
\newblock Recent advances in intra-hour solar forecasting: A review of
  ground-based sky image methods.
\newblock \emph{International Journal of Forecasting}, 2022.

\bibitem[Harzing(2013)]{harzing2013preliminary}
Anne-Wil Harzing.
\newblock A preliminary test of google scholar as a source for citation data: a
  longitudinal study of nobel prize winners.
\newblock \emph{Scientometrics}, 94\penalty0 (3):\penalty0 1057--1075, 2013.

\bibitem[Yang(2018)]{yang2018solardata}
Dazhi Yang.
\newblock Solardata: An {R} package for easy access of publicly available solar
  datasets.
\newblock \emph{Solar Energy}, 171:\penalty0 A3--A12, 2018.

\bibitem[Feng et~al.(2019{\natexlab{a}})Feng, Yang, Hodge, and
  Zhang]{Feng2019Opensolar}
Cong Feng, Dazhi Yang, Bri~Mathias Hodge, and Jie Zhang.
\newblock {OpenSolar: Promoting the openness and accessibility of diverse
  public solar datasets}.
\newblock \emph{Solar Energy}, 188:\penalty0 1369--1379, aug
  2019{\natexlab{a}}.
\newblock \doi{10.1016/j.solener.2019.07.016}.

\bibitem[Driemel et~al.(2018)Driemel, Augustine, Behrens, Colle, Cox,
  Cuevas-Agull{\'{o}}, Denn, Duprat, Fukuda, Grobe, Haeffelin, Hodges, Hyett,
  Ijima, Kallis, Knap, Kustov, Long, Longenecker, Lupi, Maturilli, Mimouni,
  Ntsangwane, Ogihara, Olano, Olefs, Omori, Passamani, Pereira,
  Schmith\"{u}sen, Schumacher, Sieger, Tamlyn, Vogt, Vuilleumier, Xia, Ohmura,
  and K\"{o}nig-Langlo]{Driemel_2018}
Amelie Driemel, John Augustine, Klaus Behrens, Sergio Colle, Christopher Cox,
  Emilio Cuevas-Agull{\'{o}}, Fred~M. Denn, Thierry Duprat, Masato Fukuda,
  Hannes Grobe, Martial Haeffelin, Gary Hodges, Nicole Hyett, Osamu Ijima, Ain
  Kallis, Wouter Knap, Vasilii Kustov, Charles~N. Long, David Longenecker,
  Angelo Lupi, Marion Maturilli, Mohamed Mimouni, Lucky Ntsangwane, Hiroyuki
  Ogihara, Xabier Olano, Marc Olefs, Masao Omori, Lance Passamani, Enio~Bueno
  Pereira, Holger Schmith\"{u}sen, Stefanie Schumacher, Rainer Sieger, Jonathan
  Tamlyn, Roland Vogt, Laurent Vuilleumier, Xiangao Xia, Atsumu Ohmura, and
  Gert K\"{o}nig-Langlo.
\newblock Baseline surface radiation network ({BSRN}): structure and data
  description (1992{\textendash}2017).
\newblock \emph{Earth System Science Data}, 10\penalty0 (3):\penalty0
  1491--1501, aug 2018.
\newblock \doi{10.5194/essd-10-1491-2018}.
\newblock URL \url{https://doi.org/10.5194\%2Fessd-10-1491-2018}.

\bibitem[Bloom et~al.(2016)Bloom, Townsend, Palchak, Novacheck, King, Barrows,
  Ibanez, O'Connell, Jordan, Roberts, Draxl, and Gruchalla]{osti2016}
Aaron Bloom, Aaron Townsend, David Palchak, Joshua Novacheck, Jack King,
  Clayton Barrows, Eduardo Ibanez, Matthew O'Connell, Gary Jordan, Billy
  Roberts, Caroline Draxl, and Kenny Gruchalla.
\newblock Eastern renewable generation integration study.
\newblock 8 2016.
\newblock \doi{10.2172/1318192}.
\newblock URL \url{https://www.osti.gov/biblio/1318192}.

\bibitem[Boyd et~al.(2017)Boyd, Chen, and Dougherty]{boyd2017nist}
M~Boyd, T~Chen, and B~Dougherty.
\newblock Nist campus photovoltaic (pv) arrays and weather station data sets.
\newblock \emph{National Institute of Standards and Technology [Data Set]},
  2017.

\bibitem[Sengupta et~al.(2018)Sengupta, Xie, Lopez, Habte, Maclaurin, and
  Shelby]{sengupta2018national}
Manajit Sengupta, Yu~Xie, Anthony Lopez, Aron Habte, Galen Maclaurin, and James
  Shelby.
\newblock The national solar radiation data base (nsrdb).
\newblock \emph{Renewable and sustainable energy reviews}, 89:\penalty0 51--60,
  2018.

\bibitem[Pombo(2022)]{VazquezPombo2022}
Daniel~Vazquez Pombo.
\newblock {The SOLETE dataset}.
\newblock 2 2022.
\newblock \doi{10.11583/DTU.17040767.v2}.
\newblock URL
  \url{https://data.dtu.dk/articles/dataset/The_SOLETE_dataset/17040767}.

\bibitem[Paletta et~al.(2022{\natexlab{c}})Paletta, Hu, Arbod, Blanc, and
  Lasenby]{palettaSPINSimplifyingPolar2021}
Quentin Paletta, Anthony Hu, Guillaume Arbod, Philippe Blanc, and Joan Lasenby.
\newblock {{SPIN}}: {{Simplifying Polar Invariance}} for {{Neural}} networks
  {{Application}} to vision-based irradiance forecasting.
\newblock In \emph{Proceedings of the {{IEEE}}/{{CVF Conference}} on {{Computer
  Vision}} and {{Pattern Recognition Workshops}}}, pages 5182--5191,
  2022{\natexlab{c}}.
\newblock URL
  \url{https://openaccess.thecvf.com/content/CVPR2022W/OmniCV/html/Paletta_SPIN_Simplifying_Polar_Invariance_for_Neural_Networks_Application_to_Vision-Based_CVPRW_2022_paper.html}.

\bibitem[Terrén-Serrano et~al.(2021)Terrén-Serrano, Bashir, Estrada, and
  Martínez-Ramón]{Girasol2021}
Guillermo Terrén-Serrano, Adnan Bashir, Trilce Estrada, and Manel
  Martínez-Ramón.
\newblock Girasol, a sky imaging and global solar irradiance dataset.
\newblock \emph{Data in Brief}, 35:\penalty0 106914, 2021.
\newblock ISSN 2352-3409.
\newblock \doi{https://doi.org/10.1016/j.dib.2021.106914}.
\newblock URL
  \url{https://www.sciencedirect.com/science/article/pii/S2352340921001980}.

\bibitem[Terr{\'e}n-Serrano and
  Mart{\'\i}nez-Ram{\'o}n(2021{\natexlab{a}})]{terren2021segmentation}
Guillermo Terr{\'e}n-Serrano and Manel Mart{\'\i}nez-Ram{\'o}n.
\newblock Segmentation algorithms for ground-based infrared cloud images.
\newblock \emph{arXiv preprint arXiv:2102.10151}, 2021{\natexlab{a}}.

\bibitem[Terr{\'e}n-Serrano and
  Mart{\'\i}nez-Ram{\'o}n(2021{\natexlab{b}})]{terren2021comparative}
Guillermo Terr{\'e}n-Serrano and Manel Mart{\'\i}nez-Ram{\'o}n.
\newblock Comparative analysis of methods for cloud segmentation in
  ground-based infrared images.
\newblock \emph{Renewable Energy}, 175:\penalty0 1025--1040,
  2021{\natexlab{b}}.

\bibitem[Stoffel and Andreas(1981{\natexlab{a}})]{SRRL1981}
T.~Stoffel and A.~Andreas.
\newblock {NREL Solar Radiation Research Laboratory (SRRL): Baseline
  Measurement System (BMS); Golden, Colorado (Data)}.
\newblock \penalty0 (NREL/DA-5500-56488), 7 1981{\natexlab{a}}.
\newblock \doi{10.7799/1052221}.
\newblock URL \url{https://www.osti.gov/biblio/1052221}.

\bibitem[{John A. Augustine and John J. DeLuisi and Charles N.
  Long}(2000)]{SURFRAD2000}
{John A. Augustine and John J. DeLuisi and Charles N. Long}.
\newblock {SURFRAD–A National Surface Radiation Budget Network for
  Atmospheric Research}.
\newblock \emph{Bulletin of the American Meteorological Society}, 81\penalty0
  (10):\penalty0 2341 -- 2358, 2000.
\newblock \doi{10.1175/1520-0477(2000)081<2341:SANSRB>2.3.CO;2}.
\newblock URL
  \url{https://journals.ametsoc.org/view/journals/bams/81/10/1520-0477_2000_081_2341_sansrb_2_3_co_2.xml}.

\bibitem[Haeffelin et~al.(2005)Haeffelin, Barth\`es, Bock, Boitel, Bony,
  Bouniol, Chepfer, Chiriaco, Cuesta, Delano\"e, Drobinski, Dufresne, Flamant,
  Grall, Hodzic, Hourdin, Lapouge, Lema\^{\i}tre, Mathieu, Morille, Naud,
  No\"el, O'Hirok, Pelon, Pietras, Protat, Romand, Scialom, and
  Vautard]{SIRTA2005}
M.~Haeffelin, L.~Barth\`es, O.~Bock, C.~Boitel, S.~Bony, D.~Bouniol,
  H.~Chepfer, M.~Chiriaco, J.~Cuesta, J.~Delano\"e, P.~Drobinski, J.-L.
  Dufresne, C.~Flamant, M.~Grall, A.~Hodzic, F.~Hourdin, F.~Lapouge,
  Y.~Lema\^{\i}tre, A.~Mathieu, Y.~Morille, C.~Naud, V.~No\"el, W.~O'Hirok,
  J.~Pelon, C.~Pietras, A.~Protat, B.~Romand, G.~Scialom, and R.~Vautard.
\newblock Sirta, a ground-based atmospheric observatory for cloud and aerosol
  research.
\newblock \emph{Annales Geophysicae}, 23\penalty0 (2):\penalty0 253--275, 2005.
\newblock \doi{10.5194/angeo-23-253-2005}.
\newblock URL \url{https://angeo.copernicus.org/articles/23/253/2005/}.

\bibitem[Miller et~al.(2005)Miller, Bucholtz, Albrecht, and
  Kollias]{ARM_MASRAD2005}
MA~Miller, A~Bucholtz, B~Albrecht, and P~Kollias.
\newblock Marine stratus radiation, aerosol, and drizzle (masrad) science plan,
  2005.

\bibitem[Slingo et~al.(2008)Slingo, Bharmal, Robinson, Settle, Allan, White,
  Lamb, L{\'e}l{\'e}, Turner, McFarlane, et~al.]{ARM_RADAGAST2008}
A~Slingo, NA~Bharmal, GJ~Robinson, JJ~Settle, RP~Allan, HE~White, Peter~J Lamb,
  M~Issa L{\'e}l{\'e}, David~D Turner, S~McFarlane, et~al.
\newblock Overview of observations from the radagast experiment in niamey,
  niger: Meteorology and thermodynamic variables.
\newblock \emph{Journal of Geophysical Research: Atmospheres}, 113\penalty0
  (D13), 2008.

\bibitem[Mace et~al.(2010)Mace, Matrosov, Shupe, Lawson, Hallar, Mc-Cubbin,
  Marchand, Orr, Coulter, Sedlacek, et~al.]{ARM_STORMVEX2010}
J~Mace, SY~Matrosov, MD~Shupe, P~Lawson, G~Hallar, I~Mc-Cubbin, R~Marchand,
  B~Orr, RL~Coulter, A~Sedlacek, et~al.
\newblock Stormvex: The storm peak lab cloud property validation experiment
  science and operations plan.
\newblock \emph{ARM Tech. Rep. DOE/SC-ARM-10-021}, 2010.

\bibitem[Long et~al.(2011)Long, Del~Genio, Deng, Fu, Gustafson, Houze, Jakob,
  Jensen, Johnson, Liu, Luke, May, McFarlane, Minnis, Schumacher, Vogelmann,
  Wang, Webster, Xie, and Zhang]{ARM_AMIE_GAN2011}
CL~Long, A~Del~Genio, M~Deng, X~Fu, W~Gustafson, R~Houze, C~Jakob, M~Jensen,
  R~Johnson, X~Liu, E~Luke, P~May, S~McFarlane, P~Minnis, C~Schumacher,
  A~Vogelmann, Y~Wang, P~Webster, S~Xie, and C~Zhang.
\newblock Arm mjo investigation experiment on gan island (amie-gan) science
  plan.
\newblock 4 2011.
\newblock URL \url{https://www.osti.gov/biblio/1010958}.

\bibitem[Wulfmeyer et~al.(2011)Wulfmeyer, Behrendt, Kottmeier, Corsmeier,
  Barthlott, Craig, Hagen, Althausen, Aoshima, Arpagaus, et~al.]{ARM_COPS2011}
Volker Wulfmeyer, Andreas Behrendt, Christoph Kottmeier, Ulrich Corsmeier,
  Christian Barthlott, George~C Craig, Martin Hagen, Dietrich Althausen, Fumiko
  Aoshima, Marco Arpagaus, et~al.
\newblock The convective and orographically-induced precipitation study (cops):
  the scientific strategy, the field phase, and research highlights.
\newblock \emph{Quarterly Journal of the Royal Meteorological Society},
  137\penalty0 (S1):\penalty0 3--30, 2011.

\bibitem[Li et~al.(2011{\natexlab{b}})Li, Li, Chen, Tsay, Holben, Huang, Li,
  Maring, Qian, Shi, et~al.]{ARM_EAST_AIRC2011}
Zhanqing Li, C~Li, H~Chen, S-C Tsay, B~Holben, J~Huang, B~Li, H~Maring, Yun
  Qian, Guangyu Shi, et~al.
\newblock East asian studies of tropospheric aerosols and their impact on
  regional climate (east-airc): An overview.
\newblock \emph{Journal of Geophysical Research: Atmospheres}, 116\penalty0
  (D7), 2011{\natexlab{b}}.

\bibitem[Kotamarthi(2013)]{ARM_GVAX2013}
VR~Kotamarthi.
\newblock Ganges valley aerosol experiment (gvax) final campaign report.
\newblock Technical report, DOE Office of Science Atmospheric Radiation
  Measurement (ARM) Program~…, 2013.

\bibitem[Dev et~al.(2015{\natexlab{b}})Dev, Lee, and Winkler]{dev2015multi}
Soumyabrata Dev, Yee~Hui Lee, and Stefan Winkler.
\newblock {Multi-level semantic labeling of sky/cloud images}.
\newblock In \emph{2015 IEEE International Conference on Image Processing
  (ICIP)}, pages 636--640. IEEE, 2015{\natexlab{b}}.

\bibitem[Wood et~al.(2015)Wood, Wyant, Bretherton, R{\'e}millard, Kollias,
  Fletcher, Stemmler, De~Szoeke, Yuter, Miller, et~al.]{ARM_CAP_MBL2015}
Robert Wood, Matthew Wyant, Christopher~S Bretherton, Jasmine R{\'e}millard,
  Pavlos Kollias, Jennifer Fletcher, Jayson Stemmler, Simone De~Szoeke, Sandra
  Yuter, Matthew Miller, et~al.
\newblock Clouds, aerosols, and precipitation in the marine boundary layer: An
  arm mobile facility deployment.
\newblock \emph{Bulletin of the American Meteorological Society}, 96\penalty0
  (3):\penalty0 419--440, 2015.

\bibitem[Berg et~al.(2016)Berg, Fast, Barnard, Burton, Cairns, Chand, Comstock,
  Dunagan, Ferrare, Flynn, et~al.]{ARM_TCAP2015}
Larry~K Berg, Jerome~D Fast, James~C Barnard, Sharon~P Burton, Brian Cairns,
  Duli Chand, Jennifer~M Comstock, Stephen Dunagan, Richard~A Ferrare, Connor~J
  Flynn, et~al.
\newblock The two-column aerosol project: Phase i—overview and impact of
  elevated aerosol layers on aerosol optical depth.
\newblock \emph{Journal of Geophysical Research: Atmospheres}, 121\penalty0
  (1):\penalty0 336--361, 2016.

\bibitem[Cheng and Lin(2017)]{cheng2017cloud}
Hsu-Yung Cheng and Chih-Lung Lin.
\newblock Cloud detection in all-sky images via multi-scale neighborhood
  features and multiple supervised learning techniques.
\newblock \emph{Atmospheric Measurement Techniques}, 10\penalty0 (1):\penalty0
  199--208, 2017.

\bibitem[Petäjä et~al.(2016)Petäjä, O’Connor, Moisseev, Sinclair,
  Manninen, Väänänen, von Lerber, Thornton, Nicoll, Petersen, Chandrasekar,
  Smith, Winkler, Krüger, Hakola, Timonen, Brus, Laurila, Asmi, Riekkola,
  Mona, Massoli, Engelmann, Komppula, Wang, Kuang, Bäck, Virtanen, Levula,
  Ritsche, and Hickmon]{BAECCA2016}
Tuukka Petäjä, Ewan~J. O’Connor, Dmitri Moisseev, Victoria~A. Sinclair,
  Antti~J. Manninen, Riikka Väänänen, Annakaisa von Lerber, Joel~A.
  Thornton, Keri Nicoll, Walt Petersen, V.~Chandrasekar, James~N. Smith,
  Paul~M. Winkler, Olaf Krüger, Hannele Hakola, Hilkka Timonen, David Brus,
  Tuomas Laurila, Eija Asmi, Marja-Liisa Riekkola, Lucia Mona, Paola Massoli,
  Ronny Engelmann, Mika Komppula, Jian Wang, Chongai Kuang, Jaana Bäck, Annele
  Virtanen, Janne Levula, Michael Ritsche, and Nicki Hickmon.
\newblock Baecc: A field campaign to elucidate the impact of biogenic aerosols
  on clouds and climate.
\newblock \emph{Bulletin of the American Meteorological Society}, 97\penalty0
  (10):\penalty0 1909 -- 1928, 2016.
\newblock \doi{10.1175/BAMS-D-14-00199.1}.
\newblock URL
  \url{https://journals.ametsoc.org/view/journals/bams/97/10/bams-d-14-00199.1.xml}.

\bibitem[Leung(2016)]{ARM_ACAPEX2016}
L~Ruby Leung.
\newblock Arm cloud-aerosol-precipitation experiment (acapex) field campaign
  report.
\newblock 3 2016.
\newblock URL \url{https://www.osti.gov/biblio/1251152}.

\bibitem[Lewis(2016)]{ARM_MAGIC2016}
Ernie~R Lewis.
\newblock Marine arm gpci investigation of clouds (magic) field campaign
  report.
\newblock Technical report, DOE Office of Science Atmospheric Radiation
  Measurement (ARM) Program~…, 2016.

\bibitem[Martin et~al.(2016)Martin, Artaxo, Machado, Manzi, Souza, Schumacher,
  Wang, Andreae, Barbosa, Fan, et~al.]{ARM_GoAmazon2016}
Scot~T Martin, Paulo Artaxo, Luiz Augusto~Toledo Machado, Ant{\^o}nio~Ocimar
  Manzi, Rodrigo Augusto Ferreira~de Souza, C~Schumacher, Jian Wang, Meinrat~O
  Andreae, HMJ Barbosa, J~Fan, et~al.
\newblock Introduction: observations and modeling of the green ocean amazon
  (goamazon2014/5).
\newblock \emph{Atmospheric Chemistry and Physics}, 16\penalty0 (8):\penalty0
  4785--4797, 2016.

\bibitem[Verlinde et~al.(2016)Verlinde, Zak, Shupe, Ivey, and
  Stamnes]{ARM_NSA2016}
J~Verlinde, BD~Zak, MD~Shupe, MD~Ivey, and K~Stamnes.
\newblock The arm north slope of alaska (nsa) sites.
\newblock \emph{Meteorological Monographs}, 57:\penalty0 8--1, 2016.

\bibitem[Sisterson et~al.(2016)Sisterson, Peppler, Cress, Lamb, and
  Turner]{ARM_SGP2016}
DL~Sisterson, RA~Peppler, TS~Cress, PJ~Lamb, and DD~Turner.
\newblock The arm southern great plains (sgp) site.
\newblock \emph{Meteorological Monographs}, 57:\penalty0 6--1, 2016.

\bibitem[Long et~al.(2016)Long, Mather, and Ackerman]{ARM_TWP2016}
CN~Long, JH~Mather, and TP~Ackerman.
\newblock The arm tropical western pacific (twp) sites.
\newblock \emph{Meteorological Monographs}, 57:\penalty0 7--1, 2016.

\bibitem[Dev et~al.(2017{\natexlab{b}})Dev, Savoy, Lee, and
  Winkler]{dev2017nighttime}
Soumyabrata Dev, Florian~M Savoy, Yee~Hui Lee, and Stefan Winkler.
\newblock Nighttime sky/cloud image segmentation.
\newblock In \emph{2017 IEEE International Conference on Image Processing
  (ICIP)}, pages 345--349. IEEE, 2017{\natexlab{b}}.

\bibitem[Dev et~al.(2018)Dev, Savoy, Lee, and Winkler]{SHWIMSEG2018}
S.~Dev, F.~M. Savoy, Y.~H. Lee, and S.~Winkler.
\newblock High-dynamic-range imaging for cloud segmentation.
\newblock \emph{Atmospheric Measurement Techniques}, 11\penalty0 (4):\penalty0
  2041--2049, 2018.
\newblock \doi{10.5194/amt-11-2041-2018}.
\newblock URL \url{https://amt.copernicus.org/articles/11/2041/2018/}.

\bibitem[Zuidema et~al.(2018)Zuidema, Alvarado, Chiu, de~Szoeke, Fairall,
  Feingold, Freedman, Ghan, Haywood, Kollias, Lewis, McFarquhar, McComiskey,
  Mechem, Onasch, Redemann, Romps, Turner, Wang, Wood, Yuter, and
  Zhu]{ARM_LASIC2018}
Paquita Zuidema, Matthew Alvarado, Christine Chiu, Simon de~Szoeke, Chris
  Fairall, Graham Feingold, Andrew Freedman, Steve Ghan, James Haywood, Pavlos
  Kollias, Ernie Lewis, Greg McFarquhar, Allison McComiskey, David Mechem, Tim
  Onasch, Jens Redemann, David Romps, David Turner, Hailong Wang, Robert Wood,
  Sandra Yuter, and Ping Zhu.
\newblock Layered atlantic smoke interactions with clouds (lasic) field
  campaign report.
\newblock 5 2018.
\newblock URL \url{https://www.osti.gov/biblio/1437446}.

\bibitem[Shi et~al.(2019)Shi, Zhou, Qiu, He, Ding, and Wei]{shi2019diurnal}
Chaojun Shi, Yatong Zhou, Bo~Qiu, Jingfei He, Mu~Ding, and Shiya Wei.
\newblock Diurnal and nocturnal cloud segmentation of all-sky imager (asi)
  images using enhancement fully convolutional networks.
\newblock \emph{Atmospheric Measurement Techniques}, 12\penalty0 (9):\penalty0
  4713--4724, 2019.

\bibitem[Caldas and Alonso-Su{\'a}rez(2019)]{caldas2019very}
M~Caldas and R~Alonso-Su{\'a}rez.
\newblock Very short-term solar irradiance forecast using all-sky imaging and
  real-time irradiance measurements.
\newblock \emph{Renewable energy}, 143:\penalty0 1643--1658, 2019.

\bibitem[Pedro et~al.(2019{\natexlab{b}})Pedro, Larson, and Coimbra]{UCSD2019}
Hugo T.~C. Pedro, David~P. Larson, and Carlos F.~M. Coimbra.
\newblock A comprehensive dataset for the accelerated development and
  benchmarking of solar forecasting methods.
\newblock \emph{Journal of Renewable and Sustainable Energy}, 11\penalty0
  (3):\penalty0 036102, 2019{\natexlab{b}}.
\newblock \doi{10.1063/1.5094494}.
\newblock URL \url{https://doi.org/10.1063/1.5094494}.

\bibitem[Dandini et~al.(2019)Dandini, Ulanowski, Campbell, and
  Kaye]{dandiniHaloRatioGroundbased2019}
Paolo Dandini, Zbigniew Ulanowski, David Campbell, and Richard Kaye.
\newblock Halo ratio from ground-based all-sky imaging.
\newblock \emph{Atmospheric Measurement Techniques}, 12\penalty0 (2):\penalty0
  1295--1309, February 2019.
\newblock ISSN 1867-8548.
\newblock \doi{10.5194/amt-12-1295-2019}.

\bibitem[Lothon et~al.(2019)Lothon, Barn{\'e}oud, Gabella, Lohou, Derrien,
  Rondi, Chiriaco, Bastin, Dupont, Haeffelin, et~al.]{lothon2019elifan}
Marie Lothon, Paul Barn{\'e}oud, Omar Gabella, Fabienne Lohou, Sol{\`e}ne
  Derrien, Sylvain Rondi, Marjolaine Chiriaco, Sophie Bastin, Jean-Charles
  Dupont, Martial Haeffelin, et~al.
\newblock Elifan, an algorithm for the estimation of cloud cover from sky
  imagers.
\newblock \emph{Atmospheric Measurement Techniques}, 12\penalty0 (10):\penalty0
  5519--5534, 2019.

\bibitem[OHP()]{OHP}
Observatoire de haute provence.
\newblock URL \url{http://www.obs-hp.fr/ohp.shtml}.

\bibitem[Baray et~al.(2008)Baray, Delmas, Courcoux, Metzger, Ferr{\'e},
  Gabarrot, Keckhut, and Porteneuve]{OPAR}
Jean-Luc Baray, Robert Delmas, Yann Courcoux, Jean-Marc Metzger, H.~Ferr{\'e},
  Franck Gabarrot, Philippe Keckhut, and Jacques Porteneuve.
\newblock {L'OPAR (Observatoire de Physique de l'Atmosph{\`e}re de la
  R{\'e}union), un site privil{\'e}gi{\'e} pour l'{\'e}tude de l'atmosph{\`e}re
  tropicale : parc instrumental, r{\'e}sultats scientifiques et projets}.
\newblock In \emph{{Atelier Instrumentation et Exp{\'e}rimentation}}, Toulouse,
  France, May 2008.
\newblock URL \url{https://hal.archives-ouvertes.fr/hal-00968050}.

\bibitem[Varble et~al.(2019)Varble, Nesbitt, Salio, Avila, Borque, DeMott,
  McFarquhar, van~den Heever, Zipser, Gochis, Houze, Jensen, Kollias,
  Kreidenweis, Leung, Rasmussen, Romps, and Williams]{ARM_CACTI2019}
Adam Varble, Steve Nesbitt, P~Salio, E~Avila, P~Borque, Paul DeMott, Greg
  McFarquhar, Susan van~den Heever, Edward Zipser, D~Gochis, Robert Houze,
  Michael Jensen, Pavlos Kollias, S~Kreidenweis, Ruby Leung, K~Rasmussen, David
  Romps, and Chris Williams.
\newblock Cloud, aerosol, and complex terrain interactions (cacti) field
  campaign report.
\newblock 11 2019.
\newblock \doi{10.2172/1574024}.
\newblock URL \url{https://www.osti.gov/biblio/1574024}.

\bibitem[Jensen(2019)]{ARM_HOU2019}
Michael~P Jensen.
\newblock Tracking aerosol convection interactions experiment (tracer) science
  plan.
\newblock Technical report, Brookhaven National Lab.(BNL), Upton, NY (United
  States), 2019.

\bibitem[McFarquhar et~al.(2019)McFarquhar, Maarchand, Bretherton, Alexander,
  Protat, Siems, Wood, and DeMott]{ARM_MARCUS2019}
Greg McFarquhar, Roger Maarchand, Chris Bretherton, Simon Alexander, Alain
  Protat, Steven Siems, Robert Wood, and Paul DeMott.
\newblock Measurements of aerosols, radiation, and clouds over the southern
  ocean (marcus) field campaign report.
\newblock Technical report, DOE Office of Science Atmospheric Radiation
  Measurement (ARM) Program~…, 2019.

\bibitem[Trigo-Rodr{\'\i}guez et~al.(2005)Trigo-Rodr{\'\i}guez, Castro-Tirado,
  Llorca, Fabregat, Mart{\'\i}nez, Reglero, Jel{\'\i}nek, Kub{\'a}nek, Mateo,
  and Ugarte~Postigo]{trigo2005development}
JM~Trigo-Rodr{\'\i}guez, AJ~Castro-Tirado, J~Llorca, J~Fabregat,
  VJ~Mart{\'\i}nez, V~Reglero, M~Jel{\'\i}nek, P~Kub{\'a}nek, T~Mateo, and A~de
  Ugarte~Postigo.
\newblock The development of the spanish fireball network using a new all-sky
  ccd system.
\newblock In \emph{Modern Meteor Science An Interdisciplinary View}, pages
  553--567. Springer, 2005.

\bibitem[C{\'o}rdoba-Jabonero et~al.(2020)C{\'o}rdoba-Jabonero,
  G{\'o}mez-Mart{\'\i}n, del {\'A}guila, Vilaplana, L{\'o}pez-Cayuela, and
  Zorzano]{El_Arenosillo}
Carmen C{\'o}rdoba-Jabonero, Laura G{\'o}mez-Mart{\'\i}n, Ana del {\'A}guila,
  Jos{\'e}~Manuel Vilaplana, Mar{\'\i}a-{\'A}ngeles L{\'o}pez-Cayuela, and
  Mar{\'\i}a-Paz Zorzano.
\newblock Cirrus-induced shortwave radiative effects depending on their optical
  and physical properties: Case studies using simulations and measurements.
\newblock \emph{Atmospheric research}, 246:\penalty0 105095, 2020.

\bibitem[Liu et~al.(2020{\natexlab{a}})Liu, Li, Zhang, Xiao, and
  Durrani]{liu2020multi}
Shuang Liu, Mei Li, Zhong Zhang, Baihua Xiao, and Tariq~S Durrani.
\newblock Multi-evidence and multi-modal fusion network for ground-based cloud
  recognition.
\newblock \emph{Remote Sensing}, 12\penalty0 (3):\penalty0 464,
  2020{\natexlab{a}}.

\bibitem[Liu et~al.(2020{\natexlab{b}})Liu, Li, Zhang, Cao, and
  Durrani]{liu2020ground}
Shuang Liu, Mei Li, Zhong Zhang, Xiaozhong Cao, and Tariq~S Durrani.
\newblock Ground-based cloud classification using task-based graph
  convolutional network.
\newblock \emph{Geophysical Research Letters}, 47\penalty0 (5):\penalty0
  e2020GL087338, 2020{\natexlab{b}}.

\bibitem[Krauz et~al.(2020)Krauz, Janout, Bla{\v{z}}ek, and
  P{\'a}ta]{krauz2020assessing}
Luk{\'a}{\v{s}} Krauz, Petr Janout, Martin Bla{\v{z}}ek, and Petr P{\'a}ta.
\newblock Assessing cloud segmentation in the chromacity diagram of all-sky
  images.
\newblock \emph{Remote Sensing}, 12\penalty0 (11):\penalty0 1902, 2020.

\bibitem[Baray et~al.(2020)Baray, Deguillaume, Colomb, Sellegri, Freney, Rose,
  Van~Baelen, Pichon, Picard, Fr{\'e}ville, et~al.]{baray2020cezeaux}
Jean-Luc Baray, Laurent Deguillaume, Aur{\'e}lie Colomb, Karine Sellegri,
  Evelyn Freney, Cl{\'e}mence Rose, Jo{\"e}l Van~Baelen, Jean-Marc Pichon,
  David Picard, Patrick Fr{\'e}ville, et~al.
\newblock C{\'e}zeaux-aulnat-opme-puy de d{\^o}me: a multi-site for the
  long-term survey of the tropospheric composition and climate change.
\newblock \emph{Atmospheric Measurement Techniques}, 13\penalty0 (6):\penalty0
  3413--3445, 2020.

\bibitem[Lubin et~al.(2020)Lubin, Zhang, Silber, Scott, Kalogeras, Battaglia,
  Bromwich, Cadeddu, Eloranta, Fridlind, et~al.]{ARM_AWARE2020}
Dan Lubin, Damao Zhang, Israel Silber, Ryan~C Scott, Petros Kalogeras,
  Alessandro Battaglia, David~H Bromwich, Maria Cadeddu, Edwin Eloranta, Ann
  Fridlind, et~al.
\newblock Aware: The atmospheric radiation measurement (arm) west antarctic
  radiation experiment.
\newblock \emph{Bulletin of the American Meteorological Society}, 101\penalty0
  (7):\penalty0 E1069--E1091, 2020.

\bibitem[Liu et~al.(2021)Liu, Duan, Zhang, Cao, and Durrani]{liu2021ground}
Shuang Liu, Linlin Duan, Zhong Zhang, Xiaozhong Cao, and Tariq~S Durrani.
\newblock Ground-based remote sensing cloud classification via context graph
  attention network.
\newblock \emph{IEEE Transactions on Geoscience and Remote Sensing},
  60:\penalty0 1--11, 2021.

\bibitem[C.~Valdelomar et~al.(2021)C.~Valdelomar, G{\'o}mez-Amo,
  Peris-Ferr{\'u}s, Scarlatti, and Utrillas]{c2021feasibility}
Pedro C.~Valdelomar, Jos{\'e}~L G{\'o}mez-Amo, Caterina Peris-Ferr{\'u}s,
  Francesco Scarlatti, and Mar{\'\i}a~Pilar Utrillas.
\newblock Feasibility of ground-based sky-camera hdr imagery to determine solar
  irradiance and sky radiance over different geometries and sky conditions.
\newblock \emph{Remote Sensing}, 13\penalty0 (24):\penalty0 5157, 2021.

\bibitem[Kim et~al.(2021)Kim, Cha, and Chang]{kim2021twenty}
Bu-Yo Kim, Joo~Wan Cha, and Ki-Ho Chang.
\newblock Twenty-four-hour cloud cover calculation using a ground-based imager
  with machine learning.
\newblock \emph{Atmospheric Measurement Techniques}, 14\penalty0 (10):\penalty0
  6695--6710, 2021.

\bibitem[Ntavelis et~al.(2021)Ntavelis, Remund, and Schmid]{SkyCam2021}
Evangelos Ntavelis, Jan Remund, and Philipp Schmid.
\newblock {SkyCam: A Dataset of Sky Images and their Irradiance values}.
\newblock 2021.
\newblock \doi{10.48550/ARXIV.2105.02922}.
\newblock URL \url{https://arxiv.org/abs/2105.02922}.

\bibitem[Bassous et~al.(2021)Bassous, Calili, and Barbosa]{SIPM2021}
Guilherme~Fonseca Bassous, Rodrigo~Flora Calili, and Carlos~Hall Barbosa.
\newblock Development of a low-cost data acquisition system for very short-term
  photovoltaic power forecasting.
\newblock \emph{Energies}, 14\penalty0 (19):\penalty0 6075, 2021.

\bibitem[Dissawa et~al.(2021)Dissawa, Godaliyadda, Ekanayake, Agalgaonkar,
  Robinson, Ekanayake, and Perera]{UOW2021}
Lasanthika~H Dissawa, Roshan~I Godaliyadda, Parakrama~B Ekanayake, Ashish~P
  Agalgaonkar, Duane Robinson, Janaka~B Ekanayake, and Sarath Perera.
\newblock Sky image-based localized, short-term solar irradiance forecasting
  for multiple pv sites via cloud motion tracking.
\newblock \emph{International Journal of Photoenergy}, 2021, 2021.

\bibitem[Park et~al.(2021)Park, Kim, Ferrier, Collis, Sankaran, and
  Beckman]{park2021prediction}
Seongha Park, Yongho Kim, Nicola~J Ferrier, Scott~M Collis, Rajesh Sankaran,
  and Pete~H Beckman.
\newblock Prediction of solar irradiance and photovoltaic solar energy product
  based on cloud coverage estimation using machine learning methods.
\newblock \emph{Atmosphere}, 12\penalty0 (3):\penalty0 395, 2021.

\bibitem[Kremser et~al.(2021)Kremser, Harvey, Kuma, Hartery, Saint-Macary,
  McGregor, Schuddeboom, Von~Hobe, Lennartz, Geddes,
  et~al.]{TAN1802_Voyage2021}
Stefanie Kremser, Mike Harvey, Peter Kuma, Sean Hartery, Alexia Saint-Macary,
  John McGregor, Alex Schuddeboom, Marc Von~Hobe, Sinikka~T Lennartz, Alex
  Geddes, et~al.
\newblock Southern ocean cloud and aerosol data: a compilation of measurements
  from the 2018 southern ocean ross sea marine ecosystems and environment
  voyage.
\newblock \emph{Earth System Science Data}, 13\penalty0 (7):\penalty0
  3115--3153, 2021.

\bibitem[Feldman et~al.(2021)Feldman, Aiken, Boos, Carroll, Chandrasekar,
  Collins, Collis, Deems, DeMott, Fan, et~al.]{ARM_GUC2021}
Daniel Feldman, Allison Aiken, William Boos, Rosemary Carroll, Venkatachalam
  Chandrasekar, William Collins, Scott Collis, Jeff Deems, Paul DeMott, Jiwen
  Fan, et~al.
\newblock Surface atmosphere integrated field laboratory (sail) science plan.
\newblock Technical report, Oak Ridge National Lab.(ORNL), Oak Ridge, TN
  (United States). Atmospheric~…, 2021.

\bibitem[Nie et~al.(2022{\natexlab{a}})Nie, Li, Scott, Sun, Venugopal, and
  Brandt]{nie2022skippd}
Yuhao Nie, Xiatong Li, Andea Scott, Yuchi Sun, Vignesh Venugopal, and Adam
  Brandt.
\newblock {SKIPP'D: a SKy Images and Photovoltaic Power Generation Dataset for
  Short-term Solar Forecasting}.
\newblock \emph{arXiv preprint arXiv:2207.00913}, 2022{\natexlab{a}}.

\bibitem[Li et~al.(2022)Li, Wang, Qiu, and Wu]{li2022all}
Xiaotong Li, Baozhu Wang, Bo~Qiu, and Chao Wu.
\newblock An all-sky camera image classification method using cloud cover
  features.
\newblock \emph{Atmospheric Measurement Techniques}, 15\penalty0 (11):\penalty0
  3629--3639, 2022.

\bibitem[Zhang et~al.(2022{\natexlab{a}})Zhang, Yang, Liu, Cao, and
  Durrani]{zhang2022ground}
Zhong Zhang, Shuzhen Yang, Shuang Liu, Xiaozhong Cao, and Tariq~S Durrani.
\newblock Ground-based remote sensing cloud detection using dual pyramid
  network and encoder--decoder constraint.
\newblock \emph{IEEE Transactions on Geoscience and Remote Sensing},
  60:\penalty0 1--10, 2022{\natexlab{a}}.

\bibitem[Ye et~al.(2022{\natexlab{a}})Ye, Wang, Cao, Yang, and Min]{ye2022self}
Liang Ye, Yufeng Wang, Zhiguo Cao, Zhibiao Yang, and Huasong Min.
\newblock A self training mechanism with scanty and incompletely annotated
  samples for learning-based cloud detection in whole sky images.
\newblock \emph{Earth and Space Science}, 9\penalty0 (6):\penalty0
  e2022EA002220, 2022{\natexlab{a}}.

\bibitem[Geerts et~al.(2022)Geerts, Giangrande, McFarquhar, Xue, Abel,
  Comstock, Crewell, DeMott, Ebell, Field, et~al.]{ARM_COMBLE2022}
Bart Geerts, Scott~E Giangrande, Greg~M McFarquhar, Lulin Xue, Steven~J Abel,
  Jennifer~M Comstock, Susanne Crewell, Paul~J DeMott, Kerstin Ebell, Paul
  Field, et~al.
\newblock The comble campaign: A study of marine boundary layer clouds in
  arctic cold-air outbreaks.
\newblock \emph{Bulletin of the American Meteorological Society}, 103\penalty0
  (5):\penalty0 E1371--E1389, 2022.

\bibitem[Shupe et~al.(2022)Shupe, Rex, Blomquist, Persson, Schmale, Uttal,
  Althausen, Angot, Archer, Bariteau, et~al.]{ARM_MOSAIC2022}
Matthew~D Shupe, Markus Rex, Byron Blomquist, P~Ola~G Persson, Julia Schmale,
  Taneil Uttal, Dietrich Althausen, H{\'e}l{\`e}ne Angot, Stephen Archer,
  Ludovic Bariteau, et~al.
\newblock Overview of the mosaic expedition: Atmosphere, 2022.

\bibitem[Wang et~al.(2022)Wang, Wood, Jensen, Chiu, Liu, Lamer, Desai,
  Giangrande, Knopf, Kollias, et~al.]{ARM_ACE_ENA2022}
Jian Wang, Rob Wood, Michael~P Jensen, J~Christine Chiu, Yangang Liu, Katia
  Lamer, Neel Desai, Scott~E Giangrande, Daniel~A Knopf, Pavlos Kollias, et~al.
\newblock Aerosol and cloud experiments in the eastern north atlantic
  (ace-ena).
\newblock \emph{Bulletin of the American Meteorological Society}, 103\penalty0
  (2):\penalty0 E619--E641, 2022.

\bibitem[Ori()]{Orion_Starshoot}
Cnr-imaa atmospheric observatory.
\newblock CIAO.
\newblock URL \url{http://www.ciao.imaa.cnr.it/}.

\bibitem[war()]{warsaw}
Radiation transfer laboratory platform (ltr).
\newblock LTR.
\newblock URL
  \url{https://www.igf.fuw.edu.pl/en/meteo-station/lab_tr_pasteura_5/}.

\bibitem[LOA()]{LOA}
Laboratoire d'optique atmosphérique.
\newblock URL \url{http://www-loa.univ-lille1.fr/}.

\bibitem[Flynn and Morris()]{ARM_OLI}
Donna Flynn and Victor Morris.
\newblock Total sky imager (tsicldmask).

\bibitem[Stoffel and Andreas(1981{\natexlab{b}})]{stoffel1981nrel}
T~Stoffel and A~Andreas.
\newblock {NREL} solar radiation research laboratory ({SRRL}): Baseline
  measurement system ({BMS}); golden, colorado (data).
\newblock Technical report, National Renewable Energy Lab.(NREL), Golden, CO
  (United States), 1981{\natexlab{b}}.

\bibitem[Augustine et~al.(2000)Augustine, DeLuisi, and
  Long]{augustine2000surfrad}
John~A Augustine, John~J DeLuisi, and Charles~N Long.
\newblock {SURFRAD}--a national surface radiation budget network for
  atmospheric research.
\newblock \emph{Bulletin of the American Meteorological Society}, 81\penalty0
  (10):\penalty0 2341--2358, 2000.

\bibitem[Jiang et~al.(2020)Jiang, Lv, and Gao]{jiang2020ultra}
Junxia Jiang, Qingquan Lv, and Xiaoqing Gao.
\newblock The ultra-short-term forecasting of global horizonal irradiance based
  on total sky images.
\newblock \emph{Remote Sensing}, 12\penalty0 (21):\penalty0 3671, 2020.

\bibitem[Wang et~al.(2020{\natexlab{b}})Wang, Xuan, Zhen, Li, Li, Zhao,
  Shafie-khah, and Catal{\~a}o]{wang2020minutely}
Fei Wang, Zhiming Xuan, Zhao Zhen, Yu~Li, Kangping Li, Liqiang Zhao, Miadreza
  Shafie-khah, and Jo{\~a}o~PS Catal{\~a}o.
\newblock A minutely solar irradiance forecasting method based on real-time sky
  image-irradiance mapping model.
\newblock \emph{Energy Conversion and Management}, 220:\penalty0 113075,
  2020{\natexlab{b}}.

\bibitem[Sun et~al.(2018{\natexlab{a}})Sun, Szűcs, and Brandt]{Sun2018}
Yuchi Sun, Gergely Szűcs, and Adam~R. Brandt.
\newblock {Solar PV output prediction from video streams using convolutional
  neural networks}.
\newblock \emph{Energy \& Environmental Science}, 11\penalty0 (7):\penalty0
  1811--1818, jul 2018{\natexlab{a}}.
\newblock \doi{10.1039/C7EE03420B}.
\newblock URL \url{http://xlink.rsc.org/?DOI=C7EE03420B}.

\bibitem[Nie et~al.(2021)Nie, Zamzam, and Brandt]{Nie2021}
Yuhao Nie, Ahmed~S Zamzam, and Adam Brandt.
\newblock {Resampling and data augmentation for short-term PV output prediction
  based on an imbalanced sky images dataset using convolutional neural
  networks}.
\newblock \emph{Solar Energy}, 224\penalty0 (May):\penalty0 341--354, 2021.
\newblock \doi{10.1016/j.solener.2021.05.095}.
\newblock URL \url{https://doi.org/10.1016/j.solener.2021.05.095}.

\bibitem[Cazorla~Cabrera et~al.(2010)]{cazorla2010development}
Alberto Cazorla~Cabrera et~al.
\newblock Development of a sky imager for cloud classification and aerosol
  characterization.
\newblock 2010.

\bibitem[Yang et~al.(2014)Yang, Kurtz, Nguyen, Urquhart, Chow, Ghonima, and
  Kleissl]{yang2014solar}
Handa Yang, Ben Kurtz, Dung Nguyen, Bryan Urquhart, Chi~Wai Chow, Mohamed
  Ghonima, and Jan Kleissl.
\newblock Solar irradiance forecasting using a ground-based sky imager
  developed at uc san diego.
\newblock \emph{Solar energy}, 103:\penalty0 502--524, 2014.

\bibitem[Richardson et~al.(2017)Richardson, Krishnaswami, Vega, and
  Cervantes]{Richardson_2017}
Walter Richardson, Hariharan Krishnaswami, Rolando Vega, and Michael Cervantes.
\newblock A low cost, edge computing, all-sky imager for cloud tracking and
  intra-hour irradiance forecasting.
\newblock \emph{Sustainability}, 9\penalty0 (4):\penalty0 482, mar 2017.
\newblock \doi{10.3390/su9040482}.
\newblock URL \url{https://doi.org/10.3390\%2Fsu9040482}.

\bibitem[Jain et~al.(2022)Jain, Sengar, Gollini, Bertolotto, McArdle, and
  Dev]{Jain_2022}
Mayank Jain, Vishal~Singh Sengar, Isabella Gollini, Michela Bertolotto, Gavin
  McArdle, and Soumyabrata Dev.
\newblock {LAMSkyCam}: A low-cost and miniature ground-based sky camera.
\newblock \emph{{HardwareX}}, 12:\penalty0 e00346, oct 2022.
\newblock \doi{10.1016/j.ohx.2022.e00346}.
\newblock URL \url{https://doi.org/10.1016\%2Fj.ohx.2022.e00346}.

\bibitem[Terr{\'e}n-Serrano and
  Mart{\'\i}nez-Ram{\'o}n(2021{\natexlab{c}})]{terren2021explicit}
Guillermo Terr{\'e}n-Serrano and Manel Mart{\'\i}nez-Ram{\'o}n.
\newblock Explicit basis function kernel methods for cloud segmentation in
  infrared sky images.
\newblock \emph{Energy Reports}, 7:\penalty0 442--450, 2021{\natexlab{c}}.

\bibitem[Ajith and Mart{\'\i}nez-Ram{\'o}n(2021)]{ajith2021deep}
Meenu Ajith and Manel Mart{\'\i}nez-Ram{\'o}n.
\newblock Deep learning based solar radiation micro forecast by fusion of
  infrared cloud images and radiation data.
\newblock \emph{Applied Energy}, 294:\penalty0 117014, 2021.

\bibitem[Lave et~al.(2015)Lave, Hayes, Pohl, and Hansen]{lave2015evaluation}
Matthew Lave, William Hayes, Andrew Pohl, and Clifford~W Hansen.
\newblock Evaluation of global horizontal irradiance to plane-of-array
  irradiance models at locations across the united states.
\newblock \emph{IEEE journal of Photovoltaics}, 5\penalty0 (2):\penalty0
  597--606, 2015.

\bibitem[Xu et~al.(2018{\natexlab{a}})Xu, Anguelov, and
  Jain]{xu2018pointfusion}
Danfei Xu, Dragomir Anguelov, and Ashesh Jain.
\newblock Pointfusion: Deep sensor fusion for 3d bounding box estimation.
\newblock In \emph{Proceedings of the IEEE conference on computer vision and
  pattern recognition}, pages 244--253, 2018{\natexlab{a}}.

\bibitem[Zhou and Hauser(2017)]{zhou2017incorporating}
Yilun Zhou and Kris Hauser.
\newblock Incorporating side-channel information into convolutional neural
  networks for robotic tasks.
\newblock In \emph{2017 IEEE International Conference on Robotics and
  Automation (ICRA)}, pages 2177--2183. IEEE, 2017.

\bibitem[Ajith and {Mart{\'i}nez-Ram{\'o}n}(2021)]{ajithDeepLearningBased2021}
Meenu Ajith and Manel {Mart{\'i}nez-Ram{\'o}n}.
\newblock Deep learning based solar radiation micro forecast by fusion of
  infrared cloud images and radiation data.
\newblock \emph{Applied Energy}, 294:\penalty0 117014, July 2021.
\newblock ISSN 0306-2619.
\newblock \doi{10.1016/j.apenergy.2021.117014}.

\bibitem[{Terr{\'e}n-Serrano} and
  {Martinez-Ramon}(2022)]{terren-serranoDeepLearningIntraHour2022}
Guillermo {Terr{\'e}n-Serrano} and Manel {Martinez-Ramon}.
\newblock Deep {{Learning}} for {{Intra-Hour Solar Forecasting}} with
  {{Fusion}} of {{Features Extracted}} from {{Infrared Sky Images}}, March
  2022.

\bibitem[Chen and Chen(2013)]{chen2013using}
Deliang Chen and Hans~Weiteng Chen.
\newblock Using the k{\"o}ppen classification to quantify climate variation and
  change: An example for 1901--2010.
\newblock \emph{Environmental Development}, 6:\penalty0 69--79, 2013.

\bibitem[Deng et~al.(2009)Deng, Dong, Socher, Li, Li, and
  {Fei-Fei}]{ImageNet2009}
Jia Deng, Wei Dong, Richard Socher, Li-Jia Li, Kai Li, and Li~{Fei-Fei}.
\newblock {{ImageNet}}: {{A}} large-scale hierarchical image database.
\newblock In \emph{2009 {{IEEE Conference}} on {{Computer Vision}} and
  {{Pattern Recognition}}}, pages 248--255, June 2009.
\newblock \doi{10.1109/CVPR.2009.5206848}.

\bibitem[Nie et~al.(2022{\natexlab{b}})Nie, Paletta, Scotta, Pomares, Arbod,
  Sgouridis, Lasenby, and Brandt]{transferlearningpaper}
Yuhao Nie, Quentin Paletta, Andea Scotta, Luis~Martin Pomares, Guillaume Arbod,
  Sgouris Sgouridis, Joan Lasenby, and Adam Brandt.
\newblock Sky-image-based solar forecasting using deep learning with
  multi-location data: training models locally, globally or via transfer
  learning?, 2022{\natexlab{b}}.
\newblock URL \url{https://arxiv.org/abs/2211.02108}.

\bibitem[Dev et~al.(2014)Dev, Lee, and Winkler]{dev2014systematic}
Soumyabrata Dev, Yee~Hui Lee, and Stefan Winkler.
\newblock Systematic study of color spaces and components for the segmentation
  of sky/cloud images.
\newblock In \emph{2014 IEEE International Conference on Image Processing
  (ICIP)}, pages 5102--5106. IEEE, 2014.

\bibitem[Dianne et~al.(2019)Dianne, Wiliem, and Lovell]{dianne2019deep}
Gemma Dianne, Arnold Wiliem, and Brian~C Lovell.
\newblock Deep-learning from mistakes: automating cloud class refinement for
  sky image segmentation.
\newblock In \emph{2019 Digital Image Computing: Techniques and Applications
  (DICTA)}, pages 1--8. IEEE, 2019.

\bibitem[Hong and Fan(2016)]{hong2016probabilistic}
Tao Hong and Shu Fan.
\newblock Probabilistic electric load forecasting: A tutorial review.
\newblock \emph{International Journal of Forecasting}, 32\penalty0
  (3):\penalty0 914--938, 2016.

\bibitem[Zhu et~al.(2016)Zhu, Wei, Zhao, Zhang, and Fang]{zhu2016method}
Tingting Zhu, Haikun Wei, Xin Zhao, Kanjian Zhang, and Shixiong Fang.
\newblock A method of cloud classification based on dni.
\newblock In \emph{2016 35th Chinese Control Conference (CCC)}, pages
  4155--4160. IEEE, 2016.

\bibitem[Feng et~al.(2019{\natexlab{b}})Feng, Yang, Hodge, and
  Zhang]{fengOpenSolarPromotingOpenness2019}
Cong Feng, Dazhi Yang, Bri-Mathias Hodge, and Jie Zhang.
\newblock {{OpenSolar}}: {{Promoting}} the openness and accessibility of
  diverse public solar datasets.
\newblock \emph{Solar Energy}, 188:\penalty0 1369--1379, August
  2019{\natexlab{b}}.
\newblock ISSN 0038-092X.
\newblock \doi{10.1016/j.solener.2019.07.016}.

\bibitem[Wen et~al.(2020)Wen, Du, Chen, Lim, Wen, Jiang, and
  Xiang]{wen2020deep}
Haoran Wen, Yang Du, Xiaoyang Chen, Enggee Lim, Huiqing Wen, Lin Jiang, and Wei
  Xiang.
\newblock Deep learning based multistep solar forecasting for pv ramp-rate
  control using sky images.
\newblock \emph{IEEE Transactions on Industrial Informatics}, 17\penalty0
  (2):\penalty0 1397--1406, 2020.

\bibitem[Zhen et~al.(2020)Zhen, Liu, Zhang, Wang, Chai, Yu, Lu, Wang, and
  Lin]{zhen2020deep}
Zhao Zhen, Jiaming Liu, Zhanyao Zhang, Fei Wang, Hua Chai, Yili Yu, Xiaoxing
  Lu, Tieqiang Wang, and Yuzhang Lin.
\newblock Deep learning based surface irradiance mapping model for solar pv
  power forecasting using sky image.
\newblock \emph{IEEE Transactions on Industry Applications}, 56\penalty0
  (4):\penalty0 3385--3396, 2020.

\bibitem[Feng(2020)]{feng2020machine}
Cong Feng.
\newblock \emph{Machine Learning-Based Renewable and Load Forecasting in Power
  and Energy Systems}.
\newblock The University of Texas at Dallas, 2020.

\bibitem[Chen et~al.(2021)Chen, Li, Du, and Lai]{chen2021solar}
Lei Chen, Yangluxi Li, Hu~Du, and Yukun Lai.
\newblock Solar radiation nowcasting through advanced cnn model integrated with
  resnet structure.
\newblock 2021.

\bibitem[Xiang et~al.(2021)Xiang, Cui, Wan, and Zhao]{xiang2021sky}
Mingjun Xiang, Wenkang Cui, Can Wan, and Changfei Zhao.
\newblock A sky image-based hybrid deep learning model for nonparametric
  probabilistic forecasting of solar irradiance.
\newblock In \emph{2021 International Conference on Power System Technology
  (POWERCON)}, pages 946--952. IEEE, 2021.

\bibitem[Zhu et~al.(2021)Zhu, Wei, and Guo]{zhu2021cloud}
Tingting Zhu, Liang Wei, and Yiren Guo.
\newblock Cloud classification of ground-based cloud images based on
  convolutional neural network.
\newblock In \emph{Journal of Physics: Conference Series}, volume 2035, page
  012020. IOP Publishing, 2021.

\bibitem[Zuo et~al.(2022)Zuo, Qiu, Jia, Wang, and Li]{zuo2022ten}
Hui-Min Zuo, Jun Qiu, Ying-Hui Jia, Qi~Wang, and Fang-Fang Li.
\newblock Ten-minute prediction of solar irradiance based on cloud detection
  and a long short-term memory (lstm) model.
\newblock \emph{Energy Reports}, 8:\penalty0 5146--5157, 2022.

\bibitem[Zhang et~al.(2022{\natexlab{b}})Zhang, Zhen, Sun, Zhang, Ren, Ma,
  Yang, and Wang]{zhang2022solar}
Xinyang Zhang, Zhao Zhen, Yiqian Sun, Yagang Zhang, Hui Ren, Hui Ma, Jian Yang,
  and Fei Wang.
\newblock Solar irradiance prediction interval estimation and deterministic
  forecasting model using ground-based sky image.
\newblock In \emph{2022 IEEE/IAS 58th Industrial and Commercial Power Systems
  Technical Conference (I\&CPS)}, pages 1--8. IEEE, 2022{\natexlab{b}}.

\bibitem[Dolatabadi et~al.(2022)Dolatabadi, Abdeltawab, and
  Mohamed]{dolatabadi2022deep}
Amirhossein Dolatabadi, Hussein~Hassan Abdeltawab, and Yasser Abdel-Rady~I
  Mohamed.
\newblock Deep reinforcement learning-based self-scheduling strategy for a
  caes-pv system using accurate sky images-based forecasting.
\newblock \emph{IEEE Transactions on Power Systems}, 2022.

\bibitem[Chen et~al.(2022)Chen, Chen, and Huang]{chen20223d}
Yuxuan Chen, Jing Chen, and Wumeng Huang.
\newblock 3d cumulus cloud scene modelling and shadow analysis method based on
  ground-based sky images.
\newblock \emph{International Journal of Applied Earth Observation and
  Geoinformation}, 109:\penalty0 102765, 2022.

\bibitem[Paletta and
  Lasenby(2020{\natexlab{b}})]{palettaTemporallyConsistentImagebased2020}
Quentin Paletta and Joan Lasenby.
\newblock A temporally consistent image-based sun tracking algorithm for solar
  energy forecasting applications.
\newblock In \emph{NeurIPS 2020 Workshop on Tackling Climate Change with
  Machine Learning}, page~10, 2020{\natexlab{b}}.
\newblock URL \url{https://www.climatechange.ai/papers/neurips2020/8}.

\bibitem[Al~Asmar et~al.(2021)Al~Asmar, Musson-Genon, Dupont, Dupont, and
  Sartelet]{al2021improvement}
L{\'e}a Al~Asmar, Luc Musson-Genon, Eric Dupont, Jean-Charles Dupont, and
  Karine Sartelet.
\newblock Improvement of solar irradiance modelling during cloudy-sky days
  using measurements.
\newblock \emph{Solar Energy}, 230:\penalty0 1175--1188, 2021.

\bibitem[Insaf et~al.(2021)Insaf, Wickramathilaka, Upendra, Godaliyadda,
  Ekanayake, Herath, Dissawa, and Ekanayake]{insaf2021global}
IM~Insaf, HMKD Wickramathilaka, MAN Upendra, GMRI Godaliyadda, MPB Ekanayake,
  HMVR Herath, DMLH Dissawa, and JB~Ekanayake.
\newblock Global horizontal irradiance modeling from sky images using resnet
  architectures.
\newblock In \emph{2021 IEEE 16th International Conference on Industrial and
  Information Systems (ICIIS)}, pages 239--244. IEEE, 2021.

\bibitem[Matsui et~al.(2012)Matsui, Long, Augustine, Halliwell, Uttal,
  Longenecker, Niebergall, Wendell, and Albee]{matsui2012evaluation}
N~Matsui, Charles~N Long, J~Augustine, D~Halliwell, Taneil Uttal,
  D~Longenecker, O~Niebergall, J~Wendell, and R~Albee.
\newblock Evaluation of arctic broadband surface radiation measurements.
\newblock \emph{Atmospheric Measurement Techniques}, 5\penalty0 (2):\penalty0
  429--438, 2012.

\bibitem[Hanschmann et~al.(2012)Hanschmann, Deneke, Roebeling, and
  Macke]{hanschmann2012evaluation}
T~Hanschmann, H~Deneke, R~Roebeling, and A~Macke.
\newblock Evaluation of the shortwave cloud radiative effect over the ocean by
  use of ship and satellite observations.
\newblock \emph{Atmospheric Chemistry and Physics}, 12\penalty0 (24):\penalty0
  12243--12253, 2012.

\bibitem[Liu et~al.(2013)Liu, Li, Zheng, Chiu, Zhao, Cadeddu, Weng, and
  Cribb]{liu2013cloud}
Jianjun Liu, Zhanqing Li, Youfei Zheng, J~Christine Chiu, Fengsheng Zhao, Maria
  Cadeddu, Fuzhong Weng, and Maureen Cribb.
\newblock Cloud optical and microphysical properties derived from ground-based
  and satellite sensors over a site in the yangtze delta region.
\newblock \emph{Journal of Geophysical Research: Atmospheres}, 118\penalty0
  (16):\penalty0 9141--9152, 2013.

\bibitem[Ylivinkka et~al.(2020)Ylivinkka, Kaupinm{\"a}ki, Virman, Peltola,
  Taipale, Pet{\"a}j{\"a}, Kerminen, Kulmala, and Ezhova]{ylivinkka2020clouds}
Ilona Ylivinkka, Santeri Kaupinm{\"a}ki, Meri Virman, Maija Peltola, Ditte
  Taipale, Tuukka Pet{\"a}j{\"a}, Veli-Matti Kerminen, Markku Kulmala, and
  Ekaterina Ezhova.
\newblock Clouds over hyyti{\"a}l{\"a}, finland: an algorithm to classify
  clouds based on solar radiation and cloud base height measurements.
\newblock \emph{Atmospheric Measurement Techniques}, 13\penalty0 (10):\penalty0
  5595--5619, 2020.

\bibitem[Yang et~al.(2019)Yang, Marshak, and Wen]{yang2019cloud}
Weidong Yang, Alexander Marshak, and Guoyong Wen.
\newblock Cloud edge properties measured by the arm shortwave spectrometer over
  ocean and land.
\newblock \emph{Journal of Geophysical Research: Atmospheres}, 124\penalty0
  (15):\penalty0 8707--8721, 2019.

\bibitem[Pennypacker(2021)]{pennypacker2021exploring}
Samuel Pennypacker.
\newblock \emph{Exploring the Interface of Land-Atmosphere Interactions and
  Boundary Layer Cloud Physics}.
\newblock PhD thesis, University of Washington, 2021.

\bibitem[Lim et~al.(2019)Lim, Riihimaki, Shi, Flynn, Kleiss, Berg, Gustafson,
  Zhang, and Johnson]{lim2019long}
Kyo-Sun~Sunny Lim, Laura~D Riihimaki, Yan Shi, Donna Flynn, Jessica~M Kleiss,
  Larry~K Berg, William~I Gustafson, Yunyan Zhang, and Karen~L Johnson.
\newblock Long-term retrievals of cloud type and fair-weather shallow cumulus
  events at the arm sgp site.
\newblock \emph{Journal of Atmospheric and Oceanic Technology}, 36\penalty0
  (10):\penalty0 2031--2043, 2019.

\bibitem[Chu et~al.(2014)Chu, Pedro, Nonnenmacher, Inman, Liao, and
  Coimbra]{chu2014smart}
Yinghao Chu, Hugo~TC Pedro, Lukas Nonnenmacher, Rich~H Inman, Zhouyi Liao, and
  Carlos~FM Coimbra.
\newblock A smart image-based cloud detection system for intrahour solar
  irradiance forecasts.
\newblock \emph{Journal of Atmospheric and Oceanic Technology}, 31\penalty0
  (9):\penalty0 1995--2007, 2014.

\bibitem[Chu et~al.(2015{\natexlab{c}})Chu, Pedro, Li, and
  Coimbra]{chu2015real}
Yinghao Chu, Hugo~TC Pedro, Mengying Li, and Carlos~FM Coimbra.
\newblock Real-time forecasting of solar irradiance ramps with smart image
  processing.
\newblock \emph{Solar Energy}, 114:\penalty0 91--104, 2015{\natexlab{c}}.

\bibitem[Pedro and Coimbra(2015)]{pedro2015nearest}
Hugo~TC Pedro and Carlos~FM Coimbra.
\newblock Nearest-neighbor methodology for prediction of intra-hour global
  horizontal and direct normal irradiances.
\newblock \emph{Renewable Energy}, 80:\penalty0 770--782, 2015.

\bibitem[Li et~al.(2016{\natexlab{b}})Li, Chu, Pedro, and
  Coimbra]{li2016quantitative}
Mengying Li, Yinghao Chu, Hugo~TC Pedro, and Carlos~FM Coimbra.
\newblock Quantitative evaluation of the impact of cloud transmittance and
  cloud velocity on the accuracy of short-term dni forecasts.
\newblock \emph{Renewable Energy}, 86:\penalty0 1362--1371, 2016{\natexlab{b}}.

\bibitem[Chu and Coimbra(2017)]{chu2017short}
Yinghao Chu and Carlos~FM Coimbra.
\newblock Short-term probabilistic forecasts for direct normal irradiance.
\newblock \emph{Renewable Energy}, 101:\penalty0 526--536, 2017.

\bibitem[Pedro et~al.(2018)Pedro, Coimbra, David, and
  Lauret]{pedro2018assessment}
Hugo~TC Pedro, Carlos~FM Coimbra, Mathieu David, and Philippe Lauret.
\newblock Assessment of machine learning techniques for deterministic and
  probabilistic intra-hour solar forecasts.
\newblock \emph{Renewable Energy}, 123:\penalty0 191--203, 2018.

\bibitem[Yang(2019)]{yang2019ultra}
Dazhi Yang.
\newblock Ultra-fast analog ensemble using kd-tree.
\newblock \emph{Journal of Renewable and Sustainable Energy}, 11\penalty0
  (5):\penalty0 053703, 2019.

\bibitem[Yang et~al.(2020)Yang, van~der Meer, and
  Munkhammar]{yang2020probabilistic}
Dazhi Yang, Dennis van~der Meer, and Joakim Munkhammar.
\newblock Probabilistic solar forecasting benchmarks on a standardized dataset
  at folsom, california.
\newblock \emph{Solar Energy}, 206:\penalty0 628--639, 2020.

\bibitem[Yang et~al.(2021)Yang, Wang, Huang, and Luo]{yang20213d}
Hao Yang, Long Wang, Chao Huang, and Xiong Luo.
\newblock 3d-cnn-based sky image feature extraction for short-term global
  horizontal irradiance forecasting.
\newblock \emph{Water}, 13\penalty0 (13):\penalty0 1773, 2021.

\bibitem[Wang et~al.(2021{\natexlab{a}})Wang, Wang, Huang, and
  Luo]{wang2021hybrid}
Zhongju Wang, Long Wang, Chao Huang, and Xiong Luo.
\newblock A hybrid ensemble learning model for short-term solar irradiance
  forecasting using historical observations and sky images.
\newblock In \emph{2021 IEEE/IAS Industrial and Commercial Power System Asia
  (I\&CPS Asia)}, pages 1404--1408. IEEE, 2021{\natexlab{a}}.

\bibitem[Liu et~al.(2022)Liu, Ye, Li, Hu, and Wang]{liu2022novel}
Lei Liu, Jin Ye, Shulei Li, Shuai Hu, and Qi~Wang.
\newblock A novel machine learning algorithm for cloud detection using aeri
  measurement data.
\newblock \emph{Remote Sensing}, 14\penalty0 (11):\penalty0 2589, 2022.

\bibitem[{Terr{\'e}n-Serrano} and
  {Mart{\'i}nez-Ram{\'o}n}(2021)]{terren_serranoDataProcessingShortTerm}
Guillermo {Terr{\'e}n-Serrano} and Manel {Mart{\'i}nez-Ram{\'o}n}.
\newblock Data {{Processing}} for {{Short-Term Solar Irradiance Forecasting}}
  using {{Ground-Based Infrared Images}}.
\newblock page~31, 2021.

\bibitem[Sun et~al.(2018{\natexlab{b}})Sun, Venugopal, and
  Brandt]{sun2018convolutional}
Yuchi Sun, Vignesh Venugopal, and Adam~R Brandt.
\newblock Convolutional neural network for short-term solar panel output
  prediction.
\newblock In \emph{2018 IEEE 7th World Conference on Photovoltaic Energy
  Conversion (WCPEC)(A Joint Conference of 45th IEEE PVSC, 28th PVSEC \& 34th
  EU PVSEC)}, pages 2357--2361. IEEE, 2018{\natexlab{b}}.

\bibitem[Venugopal(2019)]{Venugopal2019thesis}
Vignesh Venugopal.
\newblock \emph{{Search for Optimal CNN Architectures Incorporating
  Heterogeneous Inputs for Short- term Solar PV Forecasting}}.
\newblock Master thesis, Stanford University, 2019.

\bibitem[Long et~al.(2006{\natexlab{b}})Long, Ackerman, Gaustad, and
  Cole]{long2006estimation}
Charles~N Long, Thomas~P Ackerman, Krista~L Gaustad, and JNS Cole.
\newblock Estimation of fractional sky cover from broadband shortwave
  radiometer measurements.
\newblock \emph{Journal of Geophysical Research: Atmospheres}, 111\penalty0
  (D11), 2006{\natexlab{b}}.

\bibitem[Ten~Hoeve and Augustine(2016)]{ten2016aerosol}
John~E Ten~Hoeve and John~A Augustine.
\newblock Aerosol effects on cloud cover as evidenced by ground-based and
  space-based observations at five rural sites in the united states.
\newblock \emph{Geophysical Research Letters}, 43\penalty0 (2):\penalty0
  793--801, 2016.

\bibitem[Calb{\'o} et~al.(2017)Calb{\'o}, Long, Gonz{\'a}lez, Augustine, and
  McComiskey]{calbo2017thin}
Josep Calb{\'o}, Charles~N Long, Josep-Abel Gonz{\'a}lez, John Augustine, and
  Allison McComiskey.
\newblock The thin border between cloud and aerosol: Sensitivity of several
  ground based observation techniques.
\newblock \emph{Atmospheric Research}, 196:\penalty0 248--260, 2017.

\bibitem[Dev et~al.(2016{\natexlab{b}})Dev, Savoy, Lee, and
  Winkler]{dev2016rough}
Soumyabrata Dev, Florian~M Savoy, Yee~Hui Lee, and Stefan Winkler.
\newblock Rough-set-based color channel selection.
\newblock \emph{IEEE Geoscience and remote sensing letters}, 14\penalty0
  (1):\penalty0 52--56, 2016{\natexlab{b}}.

\bibitem[Rajini and Tamilpavai(2018)]{rajini2018classification}
S~Akila Rajini and G~Tamilpavai.
\newblock Classification of cloud/sky images based on knn and modified genetic
  algorithm.
\newblock In \emph{2018 International Conference on Intelligent Computing and
  Communication for Smart World (I2C2SW)}, pages 1--8. IEEE, 2018.

\bibitem[Andrade et~al.(2019)Andrade, Katoch, Turaga, Spanias, Tepedelenlioglu,
  and Jaskie]{andrade2019formation}
Juan Andrade, Sameeksha Katoch, Pavan Turaga, Andreas Spanias, Cihan
  Tepedelenlioglu, and Kristen Jaskie.
\newblock Formation-aware cloud segmentation of ground-based images with
  applications to pv systems.
\newblock In \emph{2019 10th International Conference on Information,
  Intelligence, Systems and Applications (IISA)}, pages 1--7. IEEE, 2019.

\bibitem[Dev et~al.(2019{\natexlab{b}})Dev, Manandhar, Lee, and
  Winkler]{dev2019multi}
Soumyabrata Dev, Shilpa Manandhar, Yee~Hui Lee, and Stefan Winkler.
\newblock Multi-label cloud segmentation using a deep network.
\newblock In \emph{2019 USNC-URSI Radio Science Meeting (Joint with AP-S
  Symposium)}, pages 113--114. IEEE, 2019{\natexlab{b}}.

\bibitem[Song et~al.(2020)Song, Cui, and Liu]{song2020efficient}
Qianqian Song, Zhihui Cui, and Pu~Liu.
\newblock An efficient solution for semantic segmentation of three ground-based
  cloud datasets.
\newblock \emph{Earth and Space Science}, 7\penalty0 (4):\penalty0
  e2019EA001040, 2020.

\bibitem[Shete et~al.(2020)Shete, Srinivasan, and
  Gonsalves]{shete2020tasselgan}
Snehal Shete, Srikant Srinivasan, and Timothy~A Gonsalves.
\newblock Tasselgan: An application of the generative adversarial model for
  creating field-based maize tassel data.
\newblock \emph{Plant Phenomics}, 2020, 2020.

\bibitem[Pahurkar and Losarwar()]{pahurkarcloud}
Shipra~P Pahurkar and VA~Losarwar.
\newblock Cloud detection using hygta dataset using principal component
  analysis.

\bibitem[Tulpan et~al.(2017)Tulpan, Bouchard, Ellis, and
  Minwalla]{tulpan2017detection}
Dan Tulpan, Cajetan Bouchard, Kristopher Ellis, and Cyrus Minwalla.
\newblock Detection of clouds in sky/cloud and aerial images using moment based
  texture segmentation.
\newblock In \emph{2017 International Conference on Unmanned Aircraft Systems
  (ICUAS)}, pages 1124--1133. IEEE, 2017.

\bibitem[Funk and Stuetz(2019)]{funk2019passive}
Franziska Funk and Peter Stuetz.
\newblock A passive cloud detection system for uav: System functions and
  validation.
\newblock In \emph{AIAA Scitech 2019 Forum}, page 2076, 2019.

\bibitem[Shi et~al.(2020)Shi, Zhou, Qiu, Guo, and Li]{shi2020cloudu}
Chaojun Shi, Yatong Zhou, Bo~Qiu, Dongjiao Guo, and Mengci Li.
\newblock Cloudu-net: A deep convolutional neural network architecture for
  daytime and nighttime cloud images’ segmentation.
\newblock \emph{IEEE Geoscience and Remote Sensing Letters}, 18\penalty0
  (10):\penalty0 1688--1692, 2020.

\bibitem[de~Mello~Koch et~al.(2020)de~Mello~Koch, de~Mello~Koch, and
  de~Mello~Koch]{de2020unsupervised}
Anita de~Mello~Koch, Ellen de~Mello~Koch, and Robert de~Mello~Koch.
\newblock Why unsupervised deep networks generalize.
\newblock \emph{arXiv e-prints}, pages arXiv--2012, 2020.

\bibitem[Shi et~al.(2021)Shi, Zhou, and Qiu]{shi2021cloudu}
Chaojun Shi, Yatong Zhou, and Bo~Qiu.
\newblock Cloudu-netv2: A cloud segmentation method for ground-based cloud
  images based on deep learning.
\newblock \emph{Neural Processing Letters}, 53\penalty0 (4):\penalty0
  2715--2728, 2021.

\bibitem[Roy et~al.(2021)Roy, Ahan, Soni, and Chittora]{roy2021towards}
Roshan Roy, MR~Ahan, Vaibhav Soni, and Ashish Chittora.
\newblock Towards automatic transformer-based cloud classification and
  segmentation.
\newblock In \emph{NeurIPS 2021 Workshop on Tackling Climate Change with
  Machine Learning}, volume 2021, page~60, 2021.

\bibitem[Shirazi et~al.(2021)Shirazi, Lemmens, Chowdhury, Tuomiranta, Catthoor,
  Voroshazi, and Gordon]{shirazi2021cloud}
Elham Shirazi, Joris Lemmens, Mohammed~Gofran Chowdhury, Arttu Tuomiranta,
  Francky Catthoor, Eszter Voroshazi, and Ivan Gordon.
\newblock Cloud detection for pv power forecast based on colour components of
  sky images.
\newblock In \emph{2021 IEEE 48th Photovoltaic Specialists Conference (PVSC)},
  pages 2389--2391. IEEE, 2021.

\bibitem[Shi et~al.(2022)Shi, Zhou, and Qiu]{shi2022cloudraednet}
Chaojun Shi, Yatong Zhou, and Bo~Qiu.
\newblock Cloudraednet: residual attention-based encoder--decoder network for
  ground-based cloud images segmentation in nychthemeron.
\newblock \emph{International Journal of Remote Sensing}, 43\penalty0
  (6):\penalty0 2059--2075, 2022.

\bibitem[L{\'e}v{\^e}que et~al.(2019)L{\'e}v{\^e}que, Dev, Hossari, Lee, and
  Winkler]{leveque2019subjective}
Lucie L{\'e}v{\^e}que, Soumyabrata Dev, Murhaf Hossari, Yee~Hui Lee, and Stefan
  Winkler.
\newblock Subjective quality assessment of ground-based camera images.
\newblock In \emph{2019 Photonics \& Electromagnetics Research Symposium-Fall
  (PIERS-Fall)}, pages 3168--3174. IEEE, 2019.

\bibitem[Jain et~al.(2021)Jain, Meegan, and Dev]{jain2021using}
Mayank Jain, Conor Meegan, and Soumyabrata Dev.
\newblock Using gans to augment data for cloud image segmentation task.
\newblock In \emph{2021 IEEE International Geoscience and Remote Sensing
  Symposium IGARSS}, pages 3452--3455. IEEE, 2021.

\bibitem[Rudrappa and Vijapur(2020)]{rudrappa2020cloud}
Gujanatti Rudrappa and Nataraj Vijapur.
\newblock Cloud classification using k-means clustering and content based image
  retrieval technique.
\newblock In \emph{2020 International Conference on Communication and Signal
  Processing (ICCSP)}, pages 0700--0704. IEEE, 2020.

\bibitem[Ye et~al.(2022{\natexlab{b}})Ye, Cao, Yang, and Min]{ye2022ccad}
Liang Ye, Zhiguo Cao, Zhibiao Yang, and Huasong Min.
\newblock Ccad-net: A cascade cloud attribute discrimination network for cloud
  genera segmentation in whole-sky images.
\newblock \emph{IEEE Geoscience and Remote Sensing Letters}, 19:\penalty0 1--5,
  2022{\natexlab{b}}.

\bibitem[Makwana et~al.(2022)Makwana, Nag, Susladkar, Deshmukh, Teja~R, Mittal,
  and Mohan]{makwana2022aclnet}
Dhruv Makwana, Subhrajit Nag, Onkar Susladkar, Gayatri Deshmukh, Sai~Chandra
  Teja~R, Sparsh Mittal, and C~Krishna Mohan.
\newblock Aclnet: an attention and clustering-based cloud segmentation network.
\newblock \emph{Remote Sensing Letters}, 13\penalty0 (9):\penalty0 865--875,
  2022.

\bibitem[Terr{\'e}n-Serrano and
  Mart{\'\i}nez-Ram{\'o}n(2021)]{terren2021detection}
Guillermo Terr{\'e}n-Serrano and Manel Mart{\'\i}nez-Ram{\'o}n.
\newblock Detection of clouds in multiple wind velocity fields using
  ground-based infrared sky images.
\newblock \emph{arXiv preprint arXiv:2105.03535}, 2021.

\bibitem[Shi et~al.(2017)Shi, Wang, Wang, and Xiao]{shi2017deep}
Cunzhao Shi, Chunheng Wang, Yu~Wang, and Baihua Xiao.
\newblock Deep convolutional activations-based features for ground-based cloud
  classification.
\newblock \emph{IEEE Geoscience and Remote Sensing Letters}, 14\penalty0
  (6):\penalty0 816--820, 2017.

\bibitem[Wang et~al.(2017)Wang, Shi, Wang, and Xiao]{wang2017measure}
Yu~Wang, Cunzhao Shi, Chunheng Wang, and Baihua Xiao.
\newblock Measure for the difference between lbp features extracted from
  original and resized cloud images with varying resolutions.
\newblock \emph{IEEE Geoscience and Remote Sensing Letters}, 14\penalty0
  (7):\penalty0 1106--1110, 2017.

\bibitem[Phung and Rhee(2018)]{phung2018deep}
Van~Hiep Phung and Eun~Joo Rhee.
\newblock A deep learning approach for classification of cloud image patches on
  small datasets.
\newblock \emph{Journal of information and communication convergence
  engineering}, 16\penalty0 (3):\penalty0 173--178, 2018.

\bibitem[Wang et~al.(2018{\natexlab{a}})Wang, Shi, Wang, and
  Xiao]{wang2018ground}
Yu~Wang, Cunzhao Shi, Chunheng Wang, and Baihua Xiao.
\newblock Ground-based cloud classification by learning stable local binary
  patterns.
\newblock \emph{Atmospheric Research}, 207:\penalty0 74--89,
  2018{\natexlab{a}}.

\bibitem[Xu et~al.(2018{\natexlab{b}})Xu, Wang, Qi, Shi, and
  Xiao]{xu2018unsupervised}
Jian Xu, Chunheng Wang, Chengzuo Qi, Cunzhao Shi, and Baihua Xiao.
\newblock Unsupervised semantic-based aggregation of deep convolutional
  features.
\newblock \emph{IEEE Transactions on Image Processing}, 28\penalty0
  (2):\penalty0 601--611, 2018{\natexlab{b}}.

\bibitem[Wang et~al.(2018{\natexlab{b}})Wang, Wang, Shi, and
  Xiao]{wang2018selection}
Yu~Wang, Chunheng Wang, Cunzhao Shi, and Baihua Xiao.
\newblock A selection criterion for the optimal resolution of ground-based
  remote sensing cloud images for cloud classification.
\newblock \emph{IEEE Transactions on Geoscience and Remote Sensing},
  57\penalty0 (3):\penalty0 1358--1367, 2018{\natexlab{b}}.

\bibitem[Phung and Rhee(2019)]{phung2019high}
Van~Hiep Phung and Eun~Joo Rhee.
\newblock A high-accuracy model average ensemble of convolutional neural
  networks for classification of cloud image patches on small datasets.
\newblock \emph{Applied Sciences}, 9\penalty0 (21):\penalty0 4500, 2019.

\bibitem[Liu et~al.(2019)Liu, Zhou, Wang, Peng, and Shen]{liu2019ground}
Zhanhua Liu, Shudao Zhou, Min Wang, Shuling Peng, and Ao~Shen.
\newblock Ground-based visible-light cloud image classification based on a
  convolutional neural network.
\newblock In \emph{2019 4th International Conference on Information Systems and
  Computer Networks (ISCON)}, pages 108--112. IEEE, 2019.

\bibitem[Zhang et~al.(2020)Zhang, Liu, Zhang, Iwabuchi, e~Ayres,
  De~Albuquerque, et~al.]{zhang2020ensemble}
Jinglin Zhang, Pu~Liu, Feng Zhang, Hironobu Iwabuchi, Antonio Artur de~H
  e~Ayres, Victor Hugo~C De~Albuquerque, et~al.
\newblock Ensemble meteorological cloud classification meets internet of
  dependable and controllable things.
\newblock \emph{IEEE Internet of Things Journal}, 8\penalty0 (5):\penalty0
  3323--3330, 2020.

\bibitem[Liu et~al.(2020{\natexlab{c}})Liu, Duan, Zhang, Cao, and
  Durrani]{liu2020multimodal}
Shuang Liu, Linlin Duan, Zhong Zhang, Xiaozhong Cao, and Tariq~S Durrani.
\newblock Multimodal ground-based remote sensing cloud classification via
  learning heterogeneous deep features.
\newblock \emph{IEEE Transactions on Geoscience and Remote Sensing},
  58\penalty0 (11):\penalty0 7790--7800, 2020{\natexlab{c}}.

\bibitem[Wang et~al.(2020{\natexlab{c}})Wang, Zhou, Yang, and
  Liu]{wang2020clouda}
Min Wang, Shudao Zhou, Zhong Yang, and Zhanhua Liu.
\newblock Clouda: A ground-based cloud classification method with a
  convolutional neural network.
\newblock \emph{Journal of Atmospheric and Oceanic Technology}, 37\penalty0
  (9):\penalty0 1661--1668, 2020{\natexlab{c}}.

\bibitem[Hoang(2020)]{hoang2020adaptive}
Vinh~Truong Hoang.
\newblock Adaptive ternary pattern based on supervised learning approach for
  ground-based cloud type classification.
\newblock In \emph{International Conference on Image Processing and Capsule
  Networks}, pages 280--286. Springer, 2020.

\bibitem[Hong and Hoang(2020)]{hong2020comparative}
Ha~Duong~Thi Hong and Vinh~Truong Hoang.
\newblock A comparative study of color spaces for cloud images recognition
  based on lbp and ltp features.
\newblock In \emph{International Conference on Integrated Science}, pages
  375--382. Springer, 2020.

\bibitem[Manzo and Pellino(2021)]{manzo2021voting}
Mario Manzo and Simone Pellino.
\newblock Voting in transfer learning system for ground-based cloud
  classification.
\newblock \emph{Machine Learning and Knowledge Extraction}, 3\penalty0
  (3):\penalty0 542--553, 2021.

\bibitem[Tang et~al.(2021)Tang, Yang, Zhou, Pan, Chen, and
  Zhao]{tang2021improving}
Yuzhu Tang, Pinglv Yang, Zeming Zhou, Delu Pan, Jianyu Chen, and Xiaofeng Zhao.
\newblock Improving cloud type classification of ground-based images using
  region covariance descriptors.
\newblock \emph{Atmospheric Measurement Techniques}, 14\penalty0 (1):\penalty0
  737--747, 2021.

\bibitem[Gan et~al.(2017)Gan, Lu, Li, Zhang, Yang, Ma, and Yao]{gan2017cloud}
Jinrui Gan, Weitao Lu, Qingyong Li, Zhen Zhang, Jun Yang, Ying Ma, and Wen Yao.
\newblock Cloud type classification of total-sky images using duplex
  norm-bounded sparse coding.
\newblock \emph{IEEE Journal of Selected Topics in Applied Earth Observations
  and Remote Sensing}, 10\penalty0 (7):\penalty0 3360--3372, 2017.

\bibitem[Yang et~al.(2018{\natexlab{b}})Yang, Lyu, Ma, Zhang, Li, Yao, and
  Lu]{yang2018analyzing}
Jun Yang, Weitao Lyu, Ying Ma, Yijun Zhang, Qingyong Li, Wen Yao, and Tianshu
  Lu.
\newblock Analyzing of cloud macroscopic characteristics in the shigatse area
  of the tibetan plateau using the total-sky images.
\newblock \emph{Journal of Applied Meteorology and Climatology}, 57\penalty0
  (9):\penalty0 1977--1987, 2018{\natexlab{b}}.

\bibitem[Oikonomou et~al.(2019)Oikonomou, Kazantzidis, Economou, and
  Fotopoulos]{oikonomou2019local}
Spiros Oikonomou, Andreas Kazantzidis, George Economou, and Spiros Fotopoulos.
\newblock A local binary pattern classification approach for cloud types
  derived from all-sky imagers.
\newblock \emph{International Journal of Remote Sensing}, 40\penalty0
  (7):\penalty0 2667--2682, 2019.

\bibitem[Ambildhuke and Banik(2021)]{ambildhuke2021transfer}
Geeta~Mahadeo Ambildhuke and Barnali~Gupta Banik.
\newblock Transfer learning approach-an efficient method to predict rainfall
  based on ground-based cloud images.
\newblock \emph{Ing{\'e}nierie des Syst{\`e}mes d'Information}, 26\penalty0
  (4), 2021.

\bibitem[Li(2021)]{li2021novel}
Guanshen Li.
\newblock A novel computer-aided cloud type classification method based on
  convolutional neural network with squeeze-and-excitation.
\newblock In \emph{Journal of Physics: Conference Series}, volume 1802, page
  032051. IOP Publishing, 2021.

\bibitem[Wang et~al.(2021{\natexlab{b}})Wang, Fu, Chu, Zhu, and
  Jing]{wang2021hacloudnet}
Min Wang, Yucheng Fu, Rong Chu, Shouxian Zhu, and Dahai Jing.
\newblock Hacloudnet: A ground-based cloud image classification network guided
  by height-driven attention.
\newblock In \emph{2021 International Conference on Networking Systems of AI
  (INSAI)}, pages 228--233. IEEE, 2021{\natexlab{b}}.

\bibitem[Zhang et~al.(2022{\natexlab{c}})Zhang, Mansfield, Li, Russell, Young,
  Adams, and Wang]{zhang2022machine}
Zhuomin Zhang, Elizabeth~C Mansfield, Jia Li, John Russell, George~S Young,
  Catherine Adams, and James~Z Wang.
\newblock A machine learning paradigm for studying pictorial realism: Are
  constable's clouds more real than his contemporaries?
\newblock \emph{arXiv preprint arXiv:2202.09348}, 2022{\natexlab{c}}.

\bibitem[To{\u{g}}a{\c{c}}ar and Ergen(2022)]{tougaccar2022classification}
Mesut To{\u{g}}a{\c{c}}ar and Burhan Ergen.
\newblock Classification of cloud images by using super resolution, semantic
  segmentation approaches and binary sailfish optimization method with deep
  learning model.
\newblock \emph{Computers and Electronics in Agriculture}, 193:\penalty0
  106724, 2022.

\bibitem[Sethy and Dash(2022)]{sethy2022cloud}
Prabira~Kumar Sethy and Sidhant~Kumar Dash.
\newblock Cloud classification-based fine knn using texture feature and
  opponent color features.
\newblock In \emph{Biologically Inspired Techniques in Many Criteria Decision
  Making}, pages 567--573. Springer, 2022.

\bibitem[Zhu et~al.(2022)Zhu, Chen, Hou, Bian, Yu, Chen, Tang, and
  Zhu]{zhu2022classification}
Wen Zhu, Tianliang Chen, Beiping Hou, Chen Bian, Aihua Yu, Lingchao Chen, Ming
  Tang, and Yuzhen Zhu.
\newblock Classification of ground-based cloud images by improved combined
  convolutional network.
\newblock \emph{Applied Sciences}, 12\penalty0 (3):\penalty0 1570, 2022.

\end{thebibliography}



\end{document}